\pdfoutput=1
\documentclass[runningheads]{llncs}
\usepackage[T1]{fontenc}
\usepackage{caption}
\usepackage{graphicx}
	\usepackage{url}

\usepackage{color}

	\newcommand{\remove}[1]{}  

\usepackage{myMac-llncs}

	\usepackage{graphics}
	\usepackage{epsfig}
\usepackage{mwe}                       

\newcommand{\wafrORarxiv}[2]{}
 \renewcommand{\wafrORarxiv}[2]{ #2}

	\newcommand{\cheeX}[1]{}
	\newcommand{\yijenX}[1]{}
	\newcommand{\zhaoqiX}[1]{}

	\newcommand{\savespace}[1]{}
	\newcommand{\nopath}{{\tt NO-PATH}}
	\renewcommand{\parallel}{\mathrel{/\mkern-5mu/}}
	\makeatletter
	\newcommand{\notparallel}{%
	  \mathrel{\mathpalette\not@parallel\relax}%
	}
	\newcommand{\not@parallel}[2]{%
	  \ooalign{\reflectbox{$\m@th#1\smallsetminus$}\cr\hfil$\m@th#1\parallel$\cr}%
	}
	\makeatother
	\newcommand{\cp}{\sc{cp}} 
	\newcommand{\Azero}{{\bf (A0)}}
	\newcommand{\Aone}{{\bf (A1)}}
	\newcommand{\Atwo}{{\bf (A2)}}
	\newcommand{\Athree}{{\bf (A3)}}
	\newcommand{\Afour}{{\bf (A4)}}

	\newcommand{\ccc}{C}
	\newcommand{\TTT}{{\cal T}}
	
	\newcommand{\Cl}{\mathop{C\ell}}	

	\newcommand{\cspace}{\ensuremath{C_{space}}}
	\newcommand{\cfree}{\ensuremath{C_{free}}}

	\newcommand{\getnext}{{\tt GetNext}}
	\newcommand{\expand}{{\tt Expand}}

	\newcommand{\Fp}{\mathop{Fp}} 
	\newcommand{\wtFp}{\mathop{\wt{Fp}}} 

	\newcommand{\free}{\ensuremath{\mathtt{FREE}}}
	
	\newcommand{\stuck}{\ensuremath{\mathtt{STUCK}}}
	\newcommand{\mixed}{\ensuremath{\mathtt{MIXED}}}

	\newcommand{\olmu}{\ol{\mu}} 

	\newcommand{\pA}{\calA}
	\newcommand{\pB}{\calB}
	\newcommand{\pO}{\calO}
	\newcommand{\ic}{\textrm{icc}}
	\newcommand{\tc}{\textrm{tc}}
	\newcommand{\TC}{\textrm{TC}}
	\newcommand{\seg}{\textrm{seg}}
	\newcommand{\ball}{\textrm{Ball}}
	\newcommand{\we}{\textrm{we}}
	
	\newcommand{\Cyl}{\textrm{Cyl}}
	
	\newcommand{\diam}{\mathop{}\textrm{diam}}
	\newcommand{\suff}{{\mathop{\mbox{\rm suff}}}}
	\newcommand{\pre}{{\mathop{\mbox{\rm pre}}}}
	\newcommand{\dder}[1]{\overset{#1}{\longrightarrow}} 

	\newcommand{\closure}[1]{\mathop{\textrm{clos}(#1)}} 
	\newcommand{\Null}{\mathrm{null}} 

	\newcommand{\switchLabel}[2]{#1}

	\newcommand{\whSO}{\wh{SO}(3)}

	\newcommand{\Proj}{\mathop{\textrm{Proj}}}
	
\title{Theory and Explicit Design of a \\
       Path Planner for an $SE(3)$ Robot\thanks{
       This work is supported in part by NSF Grant \#CCF-2008768.}
      } 
	\titlerunning{Explicit 6DOF Planner} 

\author{Zhaoqi Zhang \inst{1} \and
        Yi-Jen Chiang \inst{2} \and
        Chee Yap     \inst{1}
}
\authorrunning{Zhang, Chiang, and Yap}  
\institute{Department of Computer Science, Courant Institute,
	   New York University, 
	   New York, NY, USA.
           \email{zz1918@nyu.edu; yap@cs.nyu.edu}
           \and
	   Department of Computer Science and Engineering,
	   Tandon School of Engineering,
	   New York University,
           Brooklyn, NY, USA. 
	   \email{chiang@nyu.edu}
           }

\remove{ 



\savespace{

	\supplement{}
	
	\funding{
	Supported in part by NSF Grants \#CCF-1423228 and \#CCF-1563942.
	}
	
	\acknowledgements{} 

} 
} 

\begin{document}
\maketitle
\begin{abstract}
	We consider path planning for a rigid spatial robot
	with 6 degrees of freedom (6 DOFs),
	moving amidst polyhedral obstacles.
	A correct, complete and practical path planner for such
	a robot has never been achieved, although this is
	widely recognized as a key challenge in robotics.
	This paper provides a complete ``explicit'' design, down to
	explicit geometric primitives that are easily implementable.
\cheeX{Can we claim that we only need solving quadratic equations?}

	Our design is within an algorithmic framework for path planners,
	called \dt{Soft Subdivision Search} (SSS).
	The framework is based on the twin foundations of 
	$\vareps$-exactness and soft predicates, two
	concepts that are critical for rigorous numerical implementations.
	These concepts allow us to escape from
	``Zero Problems'' that prevent the correct or practical
	implementations of most exact algorithms
	of Computational Geometry.
	The practicality of SSS has been previously demonstrated for various
	robots including 5-DOF spatial robots.

	In this paper, we solve several significant
	technical challenges for $SE(3)$ robots: 
	(1) We first ensure the correct theory by proving a 
	general form of the Fundamental Theorem of the SSS theory.
	We prove this within an axiomatic framework,
	thus making it easy for future applications of this theory.
	(2) One component of $SE(3) =\RR^3\times SO(3)$
	is the non-Euclidean space $SO(3)$.
	We design a novel topologically correct data structure for $SO(3)$.
	Using the concept of \dt{subdivision charts and atlases} for $SO(3)$,
	we can now carry out subdivision of $SO(3)$.
	(3) The geometric problem of collision detection
	takes place in $\RR^3$, via the footprint map.
	Unlike sampling-based approaches, we must reason with the		
	notion of \dt{footprints of configuration boxes}, which is much
	harder to characterize.
	Exploiting the theory of \dt{soft predicates}, we design
	suitable approximate footprints which, when combined with
	the highly effective feature-set technique,
	lead to soft predicates.
	(4) Finally, we make the underlying geometric computation
	``explicit'', i.e., avoiding a general solver of 
	polynomial systems, in order to allow a direct implementation.

%

\end{abstract}

\sectL[intro]{Introduction}
	Motion planning 
	\cite{choset-etal:bk,lavalle:planning:bk}
	is a fundamental topic in robotics because a robot,
	almost by definition,
	is capable of movement.
	  \zhaoqiX{I'm curious about if the ``otherwise'' part is necessary?}
	  \cheeX{OK, remove}
	There is growing interest in
	motion planners because of the wide availability
	of inexpensive commercial robots,
	from domestic robots for vacuuming the floor, to drones that
	deliver packages.
	We focus on \dt{path planning} which, in its 
	elemental form, asks for a collision-free path from a start to a goal
	robot position, assuming a known map of the environment.
	Path planning is based on
	robot kinematics and collision-detection only, and the variety
	of such problems are surveyed in
	\cite{hss:motionplanning:crc:17}. 
	Although we ignore the issues of
	dynamics (timing, velocity, acceleration), a path
	is often used as the basis for solving
	restricted dynamics problems.

	Exact path planning have been studied from the 1980s \cite{ss2},
	and is reducible to the existential theory
	of connectivity of semi-algebraic sets
	(e.g., \cite{eldin-schost:connectivity:17}).
	The output of an exact path planner is either a robot path,
	or a \nopath\ indicator if no path exists.
	Unfortunately, the exact path planning is largely impractical.
	Even in simpler cases, correct
	implementation are rare for two reasons:
	it requires exact algebraic number computation and
	has numerous
	degenerate conditions (even in the plane)
	that are hard to enumerate or detect
	(e.g., \cite[p.32]{emiris-karavales:apollonius:06}).
	Correct implementations are possible using 
	libraries such as LEDA or CGAL or our own Core Library
	that support
	exact algebraic number types
	(see \cite{sharma-yap:crc,halperin-fogel-wein:bk}).

	The last 30 years saw a flowering of practical path planning
	algorithms based on either the \dt{Sampling Approach}
	(e.g., PRM, EST, RRT, SRT \cite{choset-etal:bk})
	or the \dt{Subdivision Approach}
	\cite{latombe:robot-motion:bk}.
	The dominance of Sampling Approach is described in a standard
	textbook in this area:
	``{\em PRM, EST, RRT, SRT, and their variants have changed the way
	path planning is performed for high-dimensional robots. They have
	also paved the way for the development of planners for problems
	beyond basic path planning.
	}''
	\cite[p.201]{choset-etal:bk}.
	Remarkably, the single bit of information, as encoded by
	\nopath\ output, is missing in the correctness criteria of these
	approaches as noted in \cite{wang-chiang-yap:motion-planning:15}.
	The standard notions of
	\dt{resolution completeness} (for Subdivision Approach)
	or \dt{probabilistic completeness} (for Sampling Approach)
	(\cite[Chapter 7.4]{choset-etal:bk})
	do not talk about detecting no paths.
	Instead, they speak of
	{\em eventually finding a path} when
	``the resolution is small enough'' (Subdivision Approach)
	or
	\zhaoqiX{Corresponding to the last sentence, should this be ``or''?}
	\cheeX{Yes, "or" is correct}
	``when the sampling is large enough'' (Sampling Approach).
	Both are recipes for non-terminating algorithms\footnote{
		These are overcome by user-supplied 
		``hyperparameters'' that are not part of the
		original problem specification.  Typically,
		it is some quitting criteria
		based on time-out or maximum sampling size.
	}
	but these are couched as ``narrow passage issues''
	(e.g., \cite{nowakiewicz:mst:10,lazy-toggle-prm:13}).
\wafrORarxiv{
See Appendix~A in the full version of this
paper~\cite{zhang-chiang-yap:se3-arxiv:24} for the literature on this
issue.
} 
{
    See Appendix~\ref{app:nopath} for the literature on this issue.
}
%

	The Subdivision Approach goes back to the beginning
	of algorithmic robotics -- see 
	\cite{brooks-perez:subdivision:83,zhu-latombe:hierarchical:91}.
	The present paper falls under this approach, but clearly
	a new theoretical foundation is needed. 
	This foundation is ultimately based on
	interval methods \cite{moore-etal:bk}
	which is needed to provide guarantees in the presence
	of numerical approximation.  The interval idea is
	encoded in the concept of \dt{soft predicates}
	\cite{wang-chiang-yap:motion-planning:15}.
	The other foundation is the concept\footnote{
		For the reader's convenience, we reproduce the
		basic definitions such as soft predicates
		and $\veps$-exactness
\wafrORarxiv{
%
%
  in~\cite[Appendix~B]{zhang-chiang-yap:se3-arxiv:24}.
            }
{
  in~Appendix~\ref{sec:sss}.
}
}
\  	of \dt{$\veps$-exactness}
	\cite{wang-chiang-yap:motion-planning:15,sss}.
	The idea here is rooted in an issue that afflicts
	all {\em exact} geometric algorithms: such algorithms must
	ultimately decide
	the sign of various computed numerical quantities, say $x$.
	For path planning, $x$ might represent the clearance of the path,
	and we need $x$ to be positive.
	Deciding the sign of $x$ is easily reduced
	\cite{sharma-yap:crc}
	to deciding if $x=0$ (``the Zero Problem'')
	\zhaoqiX{Is this kind of a jump?
	(Deciding the sign of $x$ can be reduced to deciding if $x=0$ by ...)}.
	The Zero Problem might well be
	undecidable~\cite{chang+4:computable:05,sharma-yap:crc}.
	The concept of $\veps$-exactness
	allows us to escape the Zero Problem.
	Clearly, both of the above concepts have wide spread ramification for
	computational geometry since all exact algorithms have
	implicit Zero Problems.

	Based on this dual foundation, a general framework for
	path planning called \dt{Soft Subdivision Search} (SSS) was formulated
	\cite{wang-chiang-yap:motion-planning:15,sss}.
	A series of papers
	\cite{sss,wang-chiang-yap:motion-planning:15,luo-chiang-lien-yap:link:14,yap-luo-hsu:thicklink:16,zhou-chiang-yap:complex-robot:20,rod-ring},
	has shown that SSS planners are implementable and practical.
	They included planar fat robots 
	\cite{yap-luo-hsu:thicklink:16}
	and complex robots \cite{zhou-chiang-yap:complex-robot:20},
	as well as spatial 5-DOF robots (rod and ring \cite{rod-ring}).
	The latter represents the first rigorous and complete planner for a
	5-DOF spatial robot.
	In each case, it was experimentally shown that SSS planners
	match or surpass the performance of
	state-of-art sampling algorithms.  This is surprising,
	considering the much stronger
	theoretical guarantees of SSS, including
	its ability to decide \nopath.


\remove{
%
}

\vspace*{-2mm}
\begin{figure}[htbp]
\hspace{5pt}
  \centering
  \begin{minipage}[t]{0.3\textwidth}
    \centering
    \includegraphics[width=1\textwidth]{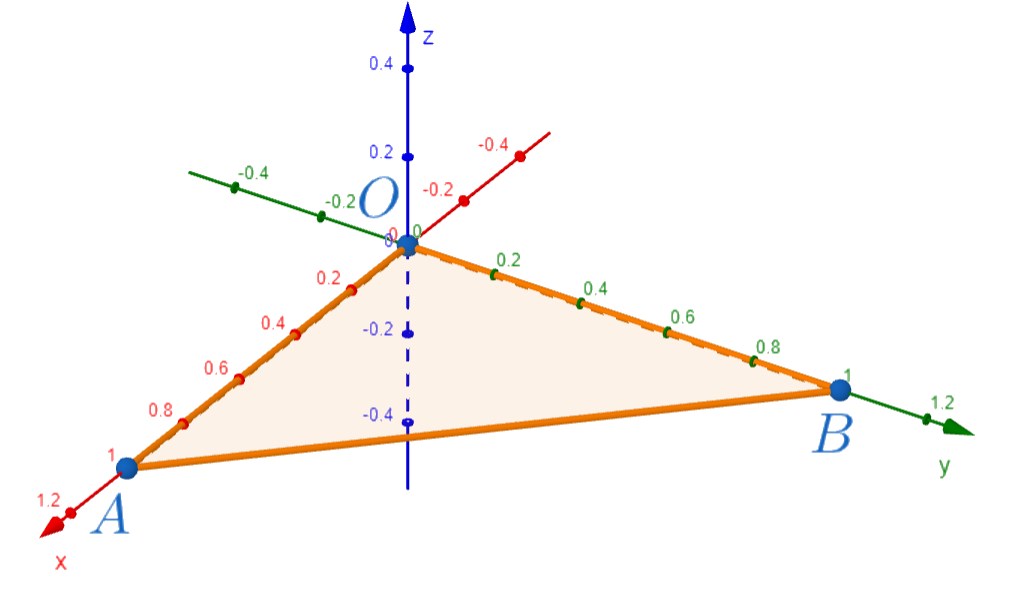} 
    \caption*{(a)}
  \end{minipage}%
  	\hspace{0.03\textwidth}
  \begin{minipage}[t]{0.3\textwidth}
    \centering
    \includegraphics[width=1\textwidth]{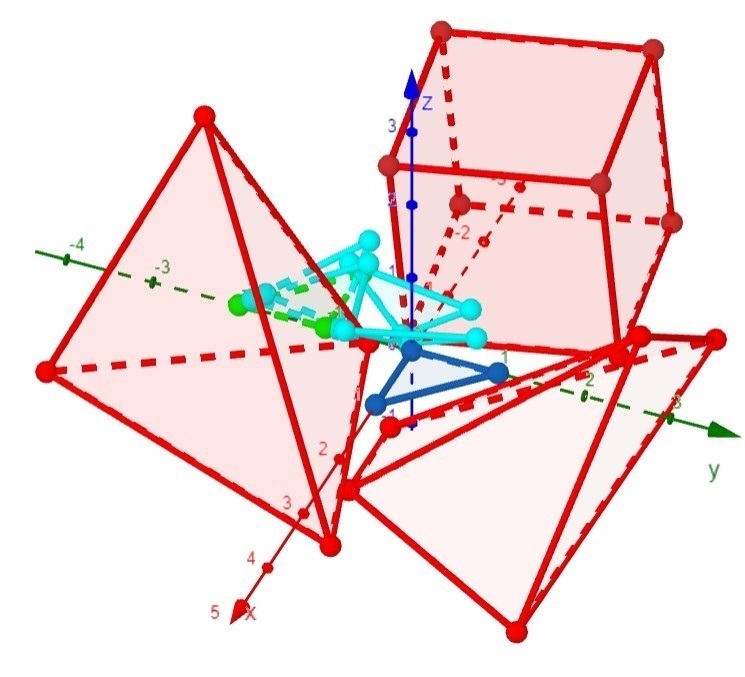} 
    \caption*{(b)}
  \end{minipage}%
  	\hspace{0.03\textwidth}
  \begin{minipage}[t]{0.3\textwidth}
    \centering
    \includegraphics[width=1\textwidth]{figs/6DOF\_Fp.png}
    \caption*{(c)}
  \end{minipage}%

	\caption{{\scriptsize  Delta Robot amidst obstacles $\Omega$:\\
	(a) Delta Robot defined by points $\pA=(1,0,0), \pO=(0,0,0), \pB=(0,1,0)$.\\
	(b) Sampled path (\cocyan{$\pA\pO\pB$}) from start (\coblue{$\pA\pO\pB$})
		to goal (\cogreen{$\pA\pO\pB$}) configurations.\\
	(c) Approximate Footprint $\wtFp(B)$ of box $B$.}}
\label{fig:Delta_Robot}
\end{figure}
\vspace*{-7mm}

	In this paper, we address a well-known
	challenge of path planning:
	\cheeX{Rewritten, check!}
	{\em to design a complete, rigorous and practical planner
	for a ``spatial 6-DOF robot''}.  It is not the 6 degrees of
	freedom per se (this is routinely achieved for robot arms),
	but the configuration space $SE(3)=\RR^3\times SO(3)$ that
	is challenging.  Like similar challenges in the past
	(rod for $SE(2)$ and in $\RR^3\times S^2$),
	we choose a simple $SE(3)$ robot
	to demonstrate the principles.  The robot is
	a planar triangle $\pA \pO \pB$ in $\RR^3$,
	a.k.a.~\dt{Delta robot}.\footnote{
		Not to be confused with
		a class of parallel manipulator robots
		called \dt{delta robots}
		E.g.,
		\myHrefx{https://en.wikipedia.org/wiki/Delta\_robot}
	}
	This is illustrated in
\switchLabel{\refFig{Delta_Robot}(a).}{Figure 1(a).}
	Its ``approximate footprint''
	at some configuration box $B\ib SE(3)$ is shown in
\switchLabel{\refFig{Delta_Robot}(c).}{Figure 1(c).}
	The path planning problem is specified as follows:

	\zhaoqiX{New paragraph? The two sentences cannot connect.}
	Given a polyhedral set $\Omega\ib\RR^3$ of obstacles,
	we want to find an $\Omega$-avoiding path from a start
	$\alpha$ to a goal $\beta$ configuration: 
	\Ldent\progb{
		\lline[-6] Path Planning for $\pA\pO\pB$-robot:
		\lline[-2] \INPUt: $(\alpha,\beta,\Omega,B_0,\veps)$
		\lline[5] where $\alpha,\beta\in SE(3)$, $B_0$ is a box
					in $SE(3)$, 
		\lline[5] $\Omega\in\RR^3$ is a polyhedral obstacle set,
				and $\veps>0$ is the resolution.
		\lline[-2] \OUTPUt: an $\Omega$-avoiding path of $\pA\pO\pB$ restricted
			to $B_0$,
		\lline[5] from $\alpha$ to $\beta$ or \nopath.
	}
	The $\veps$ parameter is used as follows:
	\begin{definition}
	A path planner is said to be \dt{resolution-exact}
	if it always terminates with an output satisfying these conditions:
	there is a constant $K>1$ independent of the input (but
	depending on the planner) such that:
		\\ (\textbf{Path})
			If the optimal clearance of a solution path is $>K\veps$,
			then the planner outputs a path.
		\\ (\textbf{NoPath}) If there is no path
			of essential clearance $<\veps/K$,
			then the planner outputs \nopath. 
	\end{definition}

	The definition of clearance and other concepts are found
\wafrORarxiv{
in~\cite[Appendix~B]{zhang-chiang-yap:se3-arxiv:24}
            }
{
in Appendix~\ref{sec:sss}
}
   and Section~\ref{ssec:sss}. 
	\zhaoqiX{Note
		that the clearance is defined in Appendix B.1 and section 2.1.}
%
	The output is indeterminate because when the optimal clearance
	lies in $[\veps/K, K\veps]$, it can output (Path) or (NoPath).
	It can be argued that $\veps$-exactness is an
	appropriate notion of ``exactness'' for real world
	applications because the physical world is inherently\footnote{
		All common constants of physics
		and chemistry have less than 8 digits of accuracy.
		Among the few exceptions is the speed of light, which is exact by
		definition.
	}
	inexact and uncertain.
\cheeX{Rewritten 16sep24, please check:}
	We believe this is the first
	completely rigorous alternative to exact path planning;
	see the Literature Review below for other attempts to 
	resolve this issue.
	\ignore{%
	The term \dt{motion planning} is very general:
	The solution to any motion planning problem is a \dt{motion}.
	Underlying the motion is a path in configuration space,
	called its \dt{(kinematic) path}.  Since we only focus on this
	path, our problem is commonly known as
	``path planning'' and the corresponding algorithms as ``planners''.
	Motion planning addresses many other
	issues beyond kinematics and collision-detection:
	call them \dt{extra-kinematic} issues.
	E.g., in trajectory planning, we need to
	find a kinematic path together with timing information along the path.
	The timing information may have to satisfy various
	constraints based on velocity, acceleration, dynamics,
	smoothness, optimality, non-holonomic constraints, etc.
	Such problems have very high
	computational complexity or have no exact solutions.
	A pragmatic approach for addressing them is to
	first solve the path planning problem (since \nopath\ implies
	no solution for the extra-kinematic planning problem).
	If a path is found, we then use it as a basis for
	constructing motions that address the extra-kinematic issues.
	It is therefore critical to have a rigorous
	foundation for path finding.
	}%
\ssectL[explicit]{What is an Explicit Algorithm in Computational
	Geometry?}
	As suggested by the title of this paper,
	our 6-DOF path planner is ``explicit''.
	This is an informal idea, attempting to characterize 
	algorithms that are recognizably in computational geometry (CG).
	Classic CG algorithms
		(see \cite{bkos:bk,goodman-orourke-toth:crc:17})
	are explicit in the sense that they construct
	well-defined combinatorial objects using explicit predicates.
	Moreover, these objects are embedded in the continuum such as
	$\RR^n$ via approximate numerical constructions, called
	semi-algebraic models in
	\cite[Sect.3.1.2, p.87]{lavalle:planning:bk}.
	E.g., Voronoi vertices are not just abstract vertices of a
	graph defined by their closest sites,
	but we typically need their approximate coordinates in $\RR^n$.
	\zhaoqiX{Then, how to compare to the ``numerical constructions''?}
	But when we address geometric problems 
	which are non-linear or in non-Euclidean spaces 
	many algorithms start to introduce highly non-trivial 
	primitives such as the following:
	\benum[(P1)]
	\item (Numerical Iteration)
	In their path planner for a spatial rod, Lee and Choset
	\cite{lee-choset:sensor-rod-planning:05}
	used a retraction approach.
	To construct edges of the generalized Voronoi diagram
	in $\RR^3\times S^2$, they invoke a numerical 
	gradient ascent method 
	\cite[p.355, column 2]{lee-choset:sensor-rod-planning:05}
	to connect Voronoi vertices.
	Such constructions are not certified or guaranteed.
	\ignore{%
		In \cite{lee-choset:sensor-rod-planning:05},
		p.355, col.2:
			"{\em In App.B, we show that once the rod achieves
		four-way equidistance, continued gradient ascent (full
		gradient ascent) brings the rod either to a 2-tangent
		edge $R_{ij/kk}$ of to a rod-GVG edge $CF_{ijklm}$.}"
		p.374, Appendix B: describes the local tangent calculation.
	}%
	\item (Optimization)
	We will need to compute the distance between
	a line and a cone in $\RR^3$ 
\wafrORarxiv{
(see~\cite[Appendix~C]{zhang-chiang-yap:se3-arxiv:24}).
            }
{
(see Appendix~\ref{app:sep}). 
}
%
        There is no known closed form expression, but
	one can reduce this to an optimization problem
	(using the Lagrangian formulation) or invoke an iterative
	procedure (e.g., \cite{zheng-chew:dist-to-cone:09}).
	\item (Purely combinatorial description)
		Nowakiewicz \cite{nowakiewicz:mst:10} described a
		sampling-and-subdivision algorithm for a 6-DOF 
		robot.  The combinatorial steps and data structures
		are explicit, but the geometric/numerical primitives
		are unspecified (presumably out sourced to various numerical
		routines).
	\item (Algebraic operations and solving systems)
	Is the intersection of two surfaces in $\RR^3$
	a geometric construction?  Depending on the surface representation,
	this may be seen as a purely algebraic construction.
	As noted above, CG needs to extract
	numerical data from algebraic representations, and this
	amounts to solving of systems of algebraic equations
	(e.g., to compute Voronoi diagram of ellipses
	\cite[Theorem 4.2]{emiris-tsig-tzou:vor-ellipses:06}). 
	\ignore{ 
		\item
		To construct the Voronoi diagram
		of a set of additively weighted points in $\RR^2$
		(a.k.a.~Apollonius diagram), Emiris and Karavales
		\cite{emiris-karavales:apollonius:06}
		}%
	\eenum

	We regard algorithms such as (P1)-(P3) as
	``non-explicit''.  But (P4) is a harder call
	because nonlinear CG is inextricably connected to algebra. 
	Some algebraic operations and analysis are
	inevitable.  Moreover, solving polynomials systems can be seen as 
	necessary geometric constructions for extracting 
	numerical data from algebra. 
	But invoking a generic polynomial solver
	inevitably gives rise to many irrelevant solutions
	(complex ones or geometrically wrong ones
	\cite{emiris-tsig-tzou:vor-ellipses:06}) 
	that must be culled.  To the extent possible,
	we seek explicit expressions for such constructions.
	Non-explicit CG algorithms are useful and sometimes unavoidable,
	but their overall correctness and complexity is hard to
	characterize.  We could largely identify ``explicit'' algorithms
	with those in semi-algebraic geometry \cite{basu-pollack-roy:bk}.

	To illustrate the ``explicitness'' achieved
	in this paper, we prove that our SSS planner for the Delta robot
	\zhaoqiX{We forgot to define the symbol $\Delta$, we only mentioned
		\textbf{Delta} robot.}
	is $\veps$-exact with resolution
	constant $K=4\sqrt{6}+6\sqrt{2}<18.3$.
	This constant is a small, manageable constant.
	It would be hard to derive such a constant if our primitives
	were not explicit.  

	\ignore{
	Emiris, Tsigaridas and Tzoumas.
		, quotable="Concerning Voronoi circles (or tritangent circle)
		for 3 ellipses:
		We solved this system numerically with PHCpack 2 which implements
		homotopy continuation, and found up to 22 real solutions. The
		total number of complex solutions was 184.
			Theorem 4.2. {\em Three ellipses admit at most 184 complex
		tritangent circles. This is tight since there are triplets at-
		taining this number.}
		Also, F.Ronga constructed up to 136 real solutions
		(but Emiris et al could not achieve this with 3 disjoint
		ellipses).
		"
	}%
	\ignore{
	The ideal CG algorithm uses explicitly
	defined algebraic predicates (like sign of determinants) and
	performs geometric constructions using
	closed-form expressions over elementary functions.
	Such algorithms are clearly very implementable.
	}%

\begin{figure}[htb]
	    	  \begin{center}
		   \scalebox{0.18}{
	    	     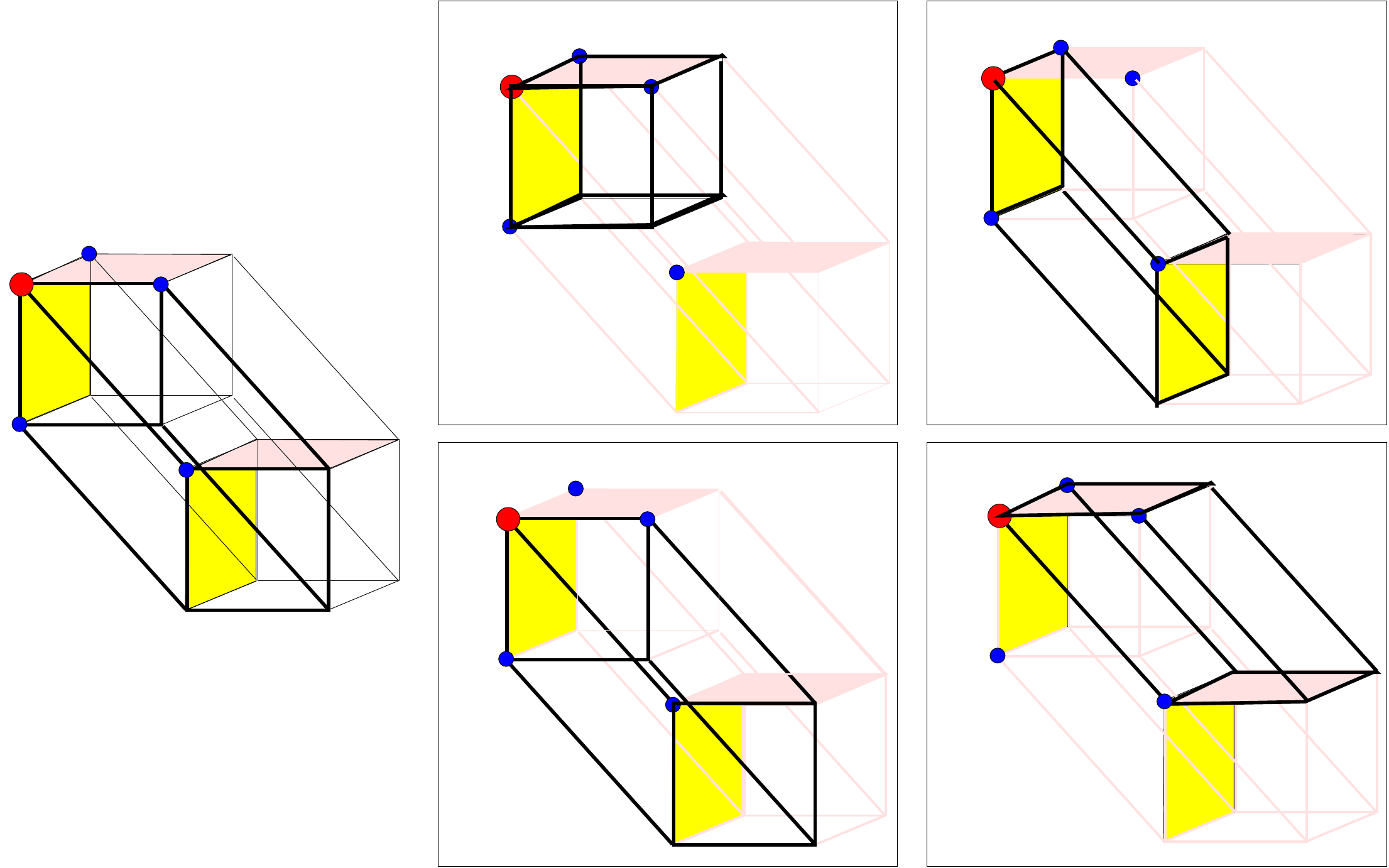}
	    	   \caption{The Cubic Model $\wh{SO}(3)$
			   			of $SO(3)$ from \cite{sss2}}
	    	   \label{fig:so3-model}
	    	  \end{center}
	    \end{figure}

	\ignore{move to conclusion?
	With suitable notions
	of soft predicates and $\veps$-exactness,
	it appears always possible to construct explicit CG algorithms that
	also automatically remove degeneracies.
	}%
\ssectL[challenge]{Challenges in $SE(3)$ Path Planning}
	Despite the successful SSS planners from previous papers
	\cite{sss,wang-chiang-yap:motion-planning:15,luo-chiang-lien-yap:link:14,yap-luo-hsu:thicklink:16,zhou-chiang-yap:complex-robot:20,rod-ring},
	there remain significant challenges in the theory and details.   
	We expand on the four issues noted in the abstract:
	\benum[(C1)]
	\item 
	By a ``fundamental theorem'' of SSS,
	we mean one that 
	says that the SSS planner is resolution exact.
	Such a theorem was proved in the original paper 
	\cite{wang-chiang-yap:motion-planning:15}, %
	albeit for a disc robot.
	Subsequent papers implicitly assumed that the fundamental
	theorem extends to other robots.
	This became less clear in subsequent development as the underlying
	techniques were generalized and configuration
	spaces became more complex.  Partly to remedy this,
	\cite{sss2} 
	gave an axiomatic account of the Fundamental Theorem.
	The power of the axiomatic approach is that,
	to verify the correctness of any future instantiations of SSS,
	one only has to check the axioms. 
	Part of axiomatization involves identifying
	the underlying mathematical spaces
	(called $X,Y,Z,W$ below). 
	There were 5 axioms, \Azero-\Afour\ in \cite{sss2}.
	These axioms introduced constants
	$C_0, D_0, L_0, \sigma$ and reveal their role
	in the implicit constant $K>1$ of the definition
	of $\veps$-exactness.
	The last axiom \Afour\ was problematic, and is remedied in this
	paper.  In \cite{sss2}, the general Fundamental
	Theorem was stated but its proof was deferred.\footnote{
		In retrospect, this deferment was appropriate in view of
		the problematic axiom (A4).
	}
	We now complete this program.
	\item 
	The configuration space\footnote{
		Some authors write
		``$SE(3)=SO(3)\ltimes \RR^3$'' where $\ltimes$
		is the semi-direct product
			\cite{selig:robotics:bk}
		on the groups $SO(3)$ and $\RR^3$.
		We forgo this algebraic detail as we
		are not interested in the group properties of $SE(3)$.
		We are only interested in $SE(3)$ as a metric space.
	}
		$SE(3)=\RR^3\times SO(3)$
	is the most general space for a rigid spatial robot,
	often simply called ``6-DOF robot''.
	A rigorous path planner for a $SE(3)$ robot
	would be a recognized milestone in robotics.
	The space $SO(3)$ is a non-Euclidean
	3-dimensional space that lives naturally in 4-dimensions
	\cite{huynh:metrics-rotation:09}.
	We will develop the algorithms and data structures
	to exploit a \dt{Cubic Model} $\wh{SO}(3)$ for $SO(3)$. 
	This model is illustrated in
\switchLabel{\refFig{so3-model},}{Figure 2,}
	and was known to Canny \cite[p.~36]{canny:thesis}.
	\ignore{
		Nowakiewicz \cite{nowakiewicz:mst:10} implied that it is
		necessary to use sampling for 6-DOF. 
		Choset-bk said that subdiv is only good for "medium" DOF,
		but sampling can reach 7-12 DOF?
		Despite achieving 5-DOF spatial robots with SSS planners,
		it is by no means clear that 6-DOF can be practically
		achieved.  
		}
		\vspace*{-2mm}
		\vspace*{-6mm}
	The design of good data structures in higher dimensions is
	generally challenging.
	For our application, our subdivisions must support the
	operation of splitting and adjacency query.  The latter is
	a nontrivial issue and raises the question of maintaining
	smooth subdivisions
	\cite{bennett-yap:smooth:14}. 
	In contrast, the sampling use of subdivision as in 
		\cite{nowakiewicz:mst:10} has no need for adjacency queries.
	\item 
	The main primitive of sampling approaches is 
	the classic \dt{collision detection problem}
	(see \cite{lin-manocha:collision:18}):
	{\em is a given configuration $\gamma$ free?}
	There are off-the-shelf solutions from well-known
	libraries \cite{lin-manocha:collision:18}.
	Our interval-based approach represents a
	nontrivial generalization: {\em is a box $B$ of configurations free
	or stuck or neither?} 
	\ignore{
		Unlike sampling-based approaches, we must reason with the
		much harder to characterize	
		notion of \dt{footprints of configuration boxes}.
		}
	Exact algorithms for this generalization
	is in general not possible (i.e., the footprint of $B$ may not
	be semi-algebraic).  
	But we can use the theory of \dt{soft predicates} to design
	practical solutions.
	\item 
	The last challenge is to make the numerical/geometric
	computations ``explicit'' as explained above.  This amounts to
	designing predicates and explicit algebraic expressions which
	allow a direct implementation. In short, we must
	avoid iterative procedures or general polynomial system solvers.
	Instead, we refine the general technique 
	of $\Sigma_2$-decomposition from \cite{rod-ring}.
	\eenum

\ignore{TOO DETAILED?  MOVE DOWN or into APPENDIX:
		The first is how to do computations over $SO(3)$, in
		particular, subdivision of $SO(3)$, viewed as
		the quotient space $S^3/\!\!\simeq$ (where
		$S^3$ is the $3$-sphere and $q\simeq -q$ for all $q\in S^3$.
		Following \cite{sss2}, consider the map
					$q\mapsto \whq = q/\|q\|_{\max}$
		where $\|q\|_{\max}=\max\set{|q_1|\dd
		|q_4|}$ is the max norm.  Then 
		$\wh{SO}(3)\as \set{\whq: q\in SO(3)}$
		is called the \dt{cubic model} for $SO(3)$.
		See 
\switchLabel{\refFig{so3-model},}{Figure 2,}
		the cubic model is the
		union of the four 3-dimensional
		cubes $C_w\cup C_x \cup C_y\cup C_z$.
		Subdivision within each 
		Since this is a non-Euclidean set, we need ...

		%
		
	\eenum
	In this paper, we shall 
	For each robot, novel algorithmic ideas are needed
	to achieve the said performance.
	In particular, a critical idea introduced to implement
	the 5-DOF spatial robots is to define approximate
	footprints of configuration boxes, and their
	$\Sigma^2$-representations.  This idea must be revisited
	in this paper.
}%

\ssect{Literature Review}
	Lavalle \cite{lavalle:planning:bk} is a comprehensive overview of
	path planning; Halperin et al \cite{hss:motionplanning:crc:17} gave a
	general survey of path planning.
	An early survey is \cite{yap:amp:87}
	where two universal approaches to exact path planning were
	described: cell-decomposition \cite{ss1} and retraction
	\cite{odun-yap:disc:85,odun-sharir-yap:retraction:83,canny:roadmap:93}.
	Since exact path planning is a semi-algebraic
	problem \cite{ss2}, it is reducible to general
	(double-exponential) cylindrical algebraic decomposition
	techniques \cite{basu-pollack-roy:bk}.  But exploiting path
	planning as a connectivity problem yields singly-exponential
	time (e.g, \cite{eldin-schost:baby-step:11}).  The case of a
	planar rod (called ``ladder'') was first studied in \cite{ss1}
	using cell-decomposition.  More efficient (quadratic time)
	methods based on the retraction method were introduced
	in \cite{odun-sharir-yap:vorI:86,odun-sharir-yap:vorII:87}.

	     Spatial rods were first treated in \cite{ss5}.
	The combinatorial complexity of its free space
	is $\Omega(n^4)$ in the worst case 
	and this can be closely matched by an
	$O(n^{4+\eps})$ time algorithm \cite{koltun:pianos-flat:05}.
	Lee and Choset
	\cite{lee-choset:sensor-rod-planning:05} gives a planner for a 3D rod
	        using a retraction approach.
	Outside of the SSS planners, perhaps the closest to this paper is
	Nowakiewicz \cite[p.~5383]{nowakiewicz:mst:10}, who uses
	subdivision of the Cubic Model.  But like many subdivision
	methods, this approach ultimately takes sample configurations
	(at the corners or centers) in subdivision boxes, and is
	actually a sampling method.
	%
	%
	\ignore{
	The method is described as \dt{MST-based} (Minimum Spanning Tree based)
	-- that would be the search strategy in our SSS terminology.
	}%
	The results were very favorable compared to pure
	sampling methods (PRM).
	For sampling-based planners, the main predicate is checking
	if a configuration is free; this is well-known 
	\dt{collision-detection problem}
	\cite{lin-manocha:collision:18}.

	The theory of soft subdivision search is the first complete
	theory of path planning that overcomes the halting issue
	in non-exact planners.  The following series of
	papers demonstrate that this theory leads to
	implementable algorithms whose efficiency beats the state-of-the-art
	sampling methods, up to 5 DOFs:
	\cite{sss,wang-chiang-yap:motion-planning:15,luo-chiang-lien-yap:link:14,yap-luo-hsu:thicklink:16,zhou-chiang-yap:complex-robot:20,rod-ring}.
	
	There is a persistent misunderstanding of the
	fundamental ``Zero Problem'' of path planning.
	Since the problem has various names
	(``disconnection proof'' \cite{basch+3:disconnection:01},
	``non-existence of path'' \cite{zhang-kim-manocha:path-non-existence:08},
	``infeasibility proof'' \cite{li-dantam:infeasibility:23}, etc),
	we will simply call it the NOPATH problem, and separately
	review this literature
\wafrORarxiv{
in~\cite[Appendix~A]{zhang-chiang-yap:se3-arxiv:24}.
            }
{
in Appendix~\ref{app:nopath}.       
}

\ssect{Overview of Paper}
	{\em Notation: We use bold font for vectors.  E.g., $\bfp\in Z$
		where $\bfp=(p_x,p_y,p_z)$. Elements in $SO(3)$ are
		viewed either as $3\times 3$ rotation
		matrices or as unit quaternions.  In the latter case, we write
		$\bfq=(q_0\dd q_3)=q_0+\ii q_1+\jj q_2+\kk q_3\in SO(3)$.
	}

	In Sect.~2, we present the axiomatic framework for SSS theory,
	and prove the Fundamental Theorem of SSS.
	In Sect.~3, we introduce the main geometric primitive
	in the design of a soft predicate for the Delta Robot.
	Various techniques for its explicit evaluation are presented.
	Sect.~4 describes the data structures for representing
	Cubic Model $\wh{SE}(3)$ of $SE(3)$. We conclude in Sect.~5.

\wafrORarxiv{ 
	Because of space limitation, additional details
	are deferred to five Appendices
        in the full version of this
        paper~\cite{zhang-chiang-yap:se3-arxiv:24}:
%
}
{ 
        We provide five Appendices for additional details:   
}
        App.~A reviews the NOPATH literature.
	App.~B reviews basic concepts of SSS.
	App.~C gives explicit ``parameterized collision
	detection predicates'' for special $\Sigma_2$-sets.
	App.~D gives details about the adjacency structures for $\wh{SE}(3)$.
	App.~E proves the Fundamental Theorem.

\sectL[fund]{The Fundamental Theorem of SSS}
	The Fundamental Theorem is about the SSS framework, which
	we review
\wafrORarxiv{
in~\cite[Appendix~B]{zhang-chiang-yap:se3-arxiv:24}.
            }
{
in~Appendix~\ref{sec:sss}.  
}
%
        This framework uses two standard data structures:
	a priority queue $Q$ and a union-find structure $U$. 
	The queue $Q$ holds boxes in $\RR^d$,
	and $U$ maintains connectivity of boxes
	through their adjacency relations ($B,B'$ are adjacent 
	if $\dim(B\cap B')= d-1$).
	SSS has 3 subroutines.
	\bitem
		\item Subroutine $B\ass Q.\getnext()$ that removes a box
			$B$ of highest priority from $Q$.  
			The search strategy of SSS amounts to defining this priority.
		\item Subroutine $\expand(B)$ that splits
			a box $B$ into its set of children (subcells). 
		\item A classifier $\wtC$ that assigns to each box $B$
			one of three values $\wtC(B)\in\set{\free,\stuck,\mixed}$.
		\eitem
	The search strategy has no effect on correctness, but
	our axioms will impose requirements on the other two subroutines.

\ssectL[sss]{The spaces of SSS Theory: $W,X,Y$ and $Z$}
	Before stating the axioms, we review some spaces
	that are central to SSS theory.

	Call $W\as \RR^d$ the \dt{computational space}
	because the SSS algorithm operates on boxes in $W$.
	Here, $d\ge 1$ is at least the degree of freedom (DOF)
	of our robot.  For $SE(3)$, we choose $d=7$ (not $d=6$)
	because we embed $SO(3)$ in $\RR^4$ to achieve the
	correct topology of $SO(3)$.
	Let $\intbox W=\intbox\RR^d$ denote\footnote{
		In \cite{sss2}, tiles were called \dt{test cells}.
		The present tiling terminology comes from the literature
		on tiling or tessellation.
	}
	the set of tiles where a \dt{tile} is
	defined to be a $d$-dimensional, compact and convex polytope
	of $\RR^d$.  Subdivision can be carried out using tiles
	(see \cite{sss2}).
	By a \dt{subdivision} of a tile $B$, we mean a
	finite set of tiles $\set{B_1\dd B_m}$
	such that $B=\bigcup_{i=1}^m B_i$
	and $\dim(B_i\cap B_j)<d$ for all $i\neq j$.  
	Suppose $\expand$ is a
	non-deterministic (i.e., multi-valued)
	function on $B\in \intbox W$
	such that $\expand(B)$ is a subdivision of $B$.
	Using $\expand$, we can grow a subdivision tree $\TTT(B)$
	rooted in $B\in\intbox W$, by repeated application of
	$\expand$ to leaves of $\TTT(B)$.  The set of leaves of $\TTT(B)$
	forms a subdivision of $B$.
	{\em General tiles are beyond the present scope;
	so we restrict them to axes-parallel boxes in this paper.}

	Next, $X\as \cspace(R_0)$ is the
	\dt{configuration space} of our robot $R_0$.
	The robot lives in some \dt{physical space} $Z \as \RR^k$
	(typically $k=2,3$), formalized via the robot's \dt{footprint map}
	$\Fp=\Fp^{R_0}: X\to 2^Z$ (power set of $Z$).
	In path planning, the input includes an
	\dt{obstacle set} $\Omega\ib Z$.
	This induces the \dt{clearance function} 
		$\Cl:X\to \RR_{\ge 0}$
	where $\Cl(\gamma)\as \Sep(\Fp(\gamma),\Omega))$
	 and $\Sep(A,B)\as \inf_{a\in A, b\in B} \|a-b\|$
	denotes the \dt{separation} between sets $A,B\ib Z$.
	We say $\gamma$ is \dt{free} iff $\Cl(\gamma)>0$. 
	Finally, $Y \as \cfree(R_0,\Omega)$ is the \dt{free space},
	comprised of all the free configurations.

	What kind\footnote{
		These spaces have many properties: this question asks
		for the minimal set of properties needed for SSS theory.
	}
	of mathematical spaces are $W,X,Y,Z$?
	Minimally, we view them as metric spaces, each 
	with its own metric: $d_W, d_X, d_Y, d_Z$.
	Since $Z, W$ are normed linear spaces, we can take 
		$d_Z(\bfa,\bfb)\as \|\bfa-\bfb\|$ ($\bfa,\bfb\in Z$),
	and similarly for $d_W$.  
	Here, $\|\cdot\|$ is the Euclidean norm, i.e., $2$-norm.
	Since $Y\ib X$, we can take $d_Y$ to be $d_X$.
	But what is $d_X$?
	The space $X$ can be\footnote{
		For instance, if $X$ is the configuration space of
		$m\ge 2$ independent, non-intersection
		discs in $\RR^2$, then $X$ is a subset of $\RR^{2m}$
		whose characterization is highly combinatorial.
	}
	very diverse in robotics.  For this paper, we assume
	$X=X^t\times X^r$ is the product of two metric
	spaces, a translational $(X^t,d_T)$ and rotational $(X^r,d_R)$ one.
	There are standard choices for $d_T$ and $d_R$ in practice.
	We can derive the metric $d_X$ from $d_T$ and $d_R$ 
	in several ways.  If $a=(a^t,a^r), b=(b^t,b^r) \in X$, we have
	three possibilities:
	{\small
		\begin{align}
			d_X(a,b) &\as \sqrt{(d_T(a^t,b^t))^2
					+\lambda\cdot (d_R(a^r,b^r))^2}
						\label{eq:dx1}\\
			d_X(a,b) &\as \max\set{d_T(a^t,b^t),
					\lambda\cdot d_R(a^r,b^r)}
						\label{eq:dx2}\\
			d_X(a,b) &\as d_T(a^t,b^t)+ \lambda\cdot d_R(a^r,b^r)
						\label{eq:dx3}
		\end{align}
		}
	where $\lambda>0$ is a fixed constant.  For definiteness,
	this paper uses the definition of \refeq{dx3} with $\lambda=1$.
	To understand the use of $\lambda$, recall that
	$X^r$ is a compact (angle) space and so $d_R$ is bounded by
	a constant.  We can take $\lambda$ to be radius of
	the ball containing\footnote{
		For a rigid robot $R_0$, we identify it with
		its footprint at $\0=(\0_t,\0_r)\in X^t\times X^r$.
		The \dt{relative center} of $R_0$ is the
		point $\bfc\in Z$ which is invariant under any 
		pure rotation $\gamma=(\0_t,\bfq)$.  Typically,
		we choose $\bfc$ to belong to $R_0\ib Z$.
	}
	$R_0$ and centered at the relative center of $R_0$.
	In this way, the pseudo-metric $d_H(a,b)$
	(see next) bounds the maximum physical displacement.

	We also need a pseudo-metric on $X$ induced by
	the footprint map: given sets $A,B\ib Z$,
	let $d_H(A,B)$ denote the standard Hausdorff distance between them
	\cite[p.86]{latombe:robot-motion:bk}.
	Given $\gamma,\gamma'\in X$, we define
		$$d_H(\gamma,\gamma')\as d_H(\Fp(\gamma),\Fp(\gamma')),$$
	called the \dt{Hausdorff pseudo-metric} on $X$.
	Although the original Hausdorff distance $d_H$ is a metric
	on closed sets, the induced $d_H$ is only a pseudo-metric in general:
	$d_H(\gamma,\gamma')=0$ may not imply $\gamma=\gamma'$.
	E.g., if $R_0$ is a rod, two configurations
	can have the same footprint.

	\dt{The case of $X=SE(3)$:}
	Here $X^t=\RR^3$ and $X^r=SO(3)$.
	As $X^t=Z$, we can choose $d_T=d_Z$ as above.
	Mathematically there is a natural choice for $d_R$ as well:
	if $M,N\in SO(3)$ are viewed as $3\times 3$ rotation matrices,
	we choose $d_R(M,N)=\|\log(MN\tr)\|$
	where $\log(MN\tr)$ is an angular measure;
	see Huynh \cite{huynh:metrics-rotation:09} who investigated 6 metrics
	$\Phi_i$ ($i=1\dd 6$) for $SO(3)$.  Our $d_R$ is the
	\dt{natural metric} denoted $\Phi_6$ in 
	\cite{huynh:metrics-rotation:09}.

	\dt{Normed linear spaces.}
	It is not enough for $Z$ and $W$ to be metric spaces.
	For example, we need to decompose sets in $Z$
	using Minkowski sum $A\oplus B$.
	For $W$, we need to scale a tile $B$ by some $\sigma>0$
	about a center $\bfm_B\in B$, denoted $\sigma B$.
	This is used in defining \dt{$\sigma$-effectivity}.
	These construction exploit the fact that $Z,W$ are
	normed linear spaces.

\ssectL[chart]{Subdivision Charts and Atlases}
	We must now connect $W$ and $X$. 
	Subdivision in Euclidean space is standard, but the
	configuration space $X$ is rarely Euclidean so that
	we cannot subdivide $X$ directly.
	To solve this, we
	use the language of charts and atlases from differential geometry.
	By a \dt{(subdivision) chart} of $X$, we mean a function
	$h: B\to X$ where $B\in\intbox W$
	and $h$ is a homeomorphism between $B$ and its image $h(B)\ib X$.
	An \dt{(subdivision) atlas} of $X$ is a set $\mu=\set{\mu_t: t\in I}$
	for some finite index set $I$ such that each $\mu_t$ ($t\in I$)
	is a chart, and if $X_t\ib X$ is the image of $\mu_t$, then
	$\dim(\mu_t\inv(X_t\cap X_s))<d$ ($t\ne s$).
	From $\mu$, we can construct a \dt{tile model} of $X$, denoted
	$X_\mu$, that is homeomorphic to $X$ via a map
	$\olmu:X\to X_\mu$
\wafrORarxiv{
(see \cite[Appendix~B.2]{zhang-chiang-yap:se3-arxiv:24}).
            }
{
(see Appendix~\ref{ssec:cubic}). 
}
%
        Note that $\olmu$ is basically the inverse of the $\mu_t$'s:
	if $x\in B_t$, then $\olmu(\mu_t(x))=x$.

\ignore{
OUR TRANSITIONS ARE TRIVIAL...
	Call the intersection $X_t\cap X_s$ ($s\neq t=1\dd m$) 
	an \dt{atlas transition} if $\dim(X_t\cap X_s)=d-1$.
	For motion planning, recall that two cells are
	adjacent if they share a face of co-dimension $1$.
	Thus atlas transitions yield adjacencies
	between cells in $\intbox X_s$ and in $\intbox X_t$.
	Thus we have two kinds of adjacencies: those that arise
	from the subdivision of tiles, and from atlas transitions.
	}%
	
	A chart $\mu:B_t\to X$ is \dt{good} if there 
	exists a \dt{chart constant} $C_0>0$ such that
	for all $q,q'\in B_t$,
		$1/C_0 \le \frac{d_X(\mu(q),\mu(q'))}{\|q-q'\|} \le C_0$.
	The subdivision atlas is \dt{good} if there is an \dt{atlas constant}
	$C_0$ that is common to its charts.
	\ignore{
		Note that good atlases can be used to produce nice sampling
	sequences: since our tiles are Euclidean
	sets, we can exploit sampling of Euclidean sets.
	}%

	\dt{The case of $X=SE(3)$:}
	First consider $X^r=SO(3)$, viewed as unit quaternions:
	the 4-cube $[-1,1]^4$ has eight $3$-dimensional cubes as faces.
	After identifying the opposite faces,
	we have four faces denoted $C_w, C_x, C_y, C_z$
	(as illustrated in
\switchLabel{\refFig{so3-model}).}{Figure 2).}
	Let $I\as \set{w,x,y,z}=\set{0,1,2,3}$ and $t\in I$.
	We view $C_t$ as a subset of $\RR^4$ where
	$\bfq=(q_0\dd q_3)\in C_t$ implies $q_t=-1$.
	Define the chart: $\mu_t: C_t\to SO(3)$ 
	by $\mu_t(\bfq)=\bfq/\|\bfq\|$ ($t\in I, \bfq\in C_t$).
	The \dt{cubic atlas} for $SO(3)$ is
	$\mu=\set{\mu_t: t\in I}$.
	The construction
\wafrORarxiv{
in~\cite[Appendix~B.2]{zhang-chiang-yap:se3-arxiv:24}
            }
{
in Appendix~\ref{ssec:cubic}  
}
%
	of the quotient space $X^r_\mu$ is called
	the \dt{cubic model} of $SO(3)$,
	also denoted $\wh{SO}(3)$.  Moreover, our special
	construction ensures that $X^r_\mu$ is embedded in $\RR^4$.
	Therefore $\wh{SE}(3)\as \RR^3\times\wh{SO}(3)$ can
	be embedded in $\RR^7$.  We define $W \as \RR^7$.

\ssect{The Axioms}
	We now state the 5 axioms in terms of the spaces $X,Y,Z,W$.
	Please refer to
\wafrORarxiv{
        \cite[Appendix~B]{zhang-chiang-yap:se3-arxiv:24}
            }
{
        Appendix~\ref{app:sss} 
}
%
        for the definitions of the terms used here.

	\benum[({A}1)]\addtocounter{enumi}{-1}
		\item {\em (Softness)}
			$\wtC$ is a soft classifier for $Y\ib X$. 
		\item {\em (Bounded dyadic expansion)}
			The expansion $\expand(B)$ is dyadic
			and there is a constant $D_0>2$ such that
			$|\expand(B)|\le D_0$, and each $B'\in\expand(B)$
			has at most $D_0$ vertices and
			has aspect ratio at most $D_0$.
		\item {\em (Pseudo-metric $d_H$ is Lipschitz)}
			There is a constant $L_0>0$ such that for all
			$\gamma,\gamma'\in Y$,
				$d_H(\gamma,\gamma') <L_0 \cdot d_X(\gamma,\gamma')$.
		\item {\em (Good Atlas)}
			The subdivision atlas $\mu$ has an atlas constant $C_0\ge 1$:
			$$ \etfrac{C_0} d_W(\olmu(\gamma),\olmu(\gamma'))
					< d_X(\gamma,\gamma')
					< C_0\cdot d_W(\olmu(\gamma),\olmu(\gamma'))$$
		\item {\em (Translational Cells)}
			Each box $B\ib \intbox W$ has the form
			$B=B^t\times B^r$ where $B^t\in \intbox Z$
			and $\Fp(B)=B^t\oplus \Fp(B^r)$.
			Such boxes\footnote{
				The original definition of translational cells
				in \cite{sss2} reads as follows:
				{\em there is a constant $K_0>0$ such that if
				$B\in\intbox X$ is free, then its inner
				center $c_0=c_0(B)$ has clearance
				$\Cl(c_0)\ge K_0\cdot r_0(B)$.}
			}
			are called \dt{translational}.
	\eenum

	\bthmT{Fundamental Theorem of SSS}{fundThm}
		Assuming Axioms \Azero-\Afour.
		If the soft classifier is $\sigma$-effective, then
		SSS Planner is resolution exact with resolution constant
			$$K=L_0C_0D_0\sigma$$
	\ethmT

	\dt{Application to our $SE(3)$ path planner:}
	For our $SE(3)$ robot design,
	$\expand(B)$ has at most $2^d$ congruent subboxes.
	Thus, we can choose $D_0=2^d$ of Axiom \Aone.
	We can easily show that the
		{\em the cubic atlases for $SO(3)$ is good}.
	However, to prove the exact bound for the distortion constant $C_0$
	for $SO(n)$, we need the tools of differential geometry
	as in \cite{zhang-yap:distortion:23}.
	The remaining issue is Axiom \Azero, that classifier
	$\wtC$ must be $\sigma$-effective for some $\sigma>1$.
	We will develop $\wtC$ in the next section and prove that it is
	$(2+\sqrt{3})$-effective.
	Hence the Fundamental Theorem implies 
	our SE(3) planner is resolution exact.
	
	Next we briefly comment on these axioms.
	Axiom \Aone\ refers to ``dyadic expansion'':
	a tile is \dt{dyadic} if its vertices are represented
	exactly by dyadic numbers (binary floats).
	Dyadic subdivision means that each tile is the
	expansion remains dyadic -- this implies that
	we can carry out subdivision without any numerical error.
	Axiom \Atwo\ shows that the Hausdorff pseudo metric
	$d_H$ is Lipschitz in the metric $d_X$.  This is
	actually a strengthening of the original axiom.
	It is strictly not necessary for the Fundamental Theorem.
\cheeX{Zhaoqi, can you please check this claim?}

	We said that the advantage of the axiomatic approach
	is that it tells us precisely which axioms are needed for
	any property of our SSS planner.  In particular,
	\cite[Theorem 2]{sss2}
	shows that Axioms \Azero\ and \Aone\ ensure the SSS planner halts.
	Very often, roboticists argue
	the correctness of their algorithms
	under the assumption of \dt{exact} predicates
	and operations.  What can we prove if the soft
	predicate of Axiom \Azero\ were exact?
	Then it can be shown \cite[Theorem 3]{sss2}
	that when the clearance is $>2C_0D_0L_0\veps$,
	the planner produces a path under Axioms \Azero-\Athree.
	But what if we want an $\veps$-exact algorithm?  
	That means that an output of \nopath\ comes
	with a guarantee the clearance is $\le K\veps$
	for some $K$.  For such a result,
	\cite[Theorem 5]{sss2} invokes the
	problematic Axiom \Afour.  We fix this issue 
\wafrORarxiv{
  in~\cite[Appendix~E]{zhang-chiang-yap:se3-arxiv:24}.
            }
{
  in Appendix~\ref{app:pf}.  
}
\sectL[appFp]{Approximate Footprint for Delta Robot:
	Computational Techniques}
	In this section, we describe the design of the approximate
	footprint of a box, and the techniques to
	compute the necessary predicates explicitly.

	Axiom \Azero\ requires an effective soft predicate
	for boxes $B\in\intbox W$.
	To compute the exact classifier function,
	 		$C(B)\in\set{\free,\stuck,\mixed}$,
	the method of features
	\cite{wang-chiang-yap:motion-planning:15} 
	says that it can be reduced to asking
			``is $\Fp(B)\cap f$ empty?''
	for features $f\in \Phi(\Omega)$.
	Since the geometry of $\Fp(B)$ is too involved,
	the paper \cite{rod-ring}
	introduced the idea of \dt{approximate footprint}
	$\wtFp(B)$ as substitute for $\Fp(B)$.
	To achieve soft predicates with effectivity $\sigma>1$, we need:
		\beql{inclusion}
			\Fp(B) \ib \wtFp(B) \ib \Fp(\sigma B).
			\eeql
	We say $\wtFp$ is $\sigma$-effective if it satisfies
	\refeq{inclusion} for all $B$.
\ssect{Design of $\wtFp(B)$ for Delta Robot}
	\FigEPS{w_box}{0.3}{Approximate rotation footprint $\wtFp(B^r)$.
		Cf.~Fig.~\protect{\ref{fig:Delta_Robot}}(c).}

	Given
		$B=B^t\times B^r$,
	we have
		$\Fp(B)=B^t\oplus \Fp(B^r)$ (by translational axiom (A4)).
	\zhaoqiX{$\Fp(B)=B^t\oplus \Fp(B^r)$ by translational axiom}
	Its \dt{approximate footprint} of $B$ is
		\beql{apprxFp}
			\wtFp(B)\as Ball(B^t)\oplus \wtFp(B^r)\eeql
	where \qquad
		$\wtFp(B^r) \as \bigcup_{i=1}^ 6 P_i
				= S_A \cup S_B
					\cup Cyl \cup Cone_A \cup Cone_B \cup Pyr.
		$

	\vspace*{3mm}
	The sets $P_1\dd P_6$ are comprised of two balls ($S_A, S_B$),
	a cylinder $Cyl$, two finite cones ($Cone_A, Cone_B$) and
	a convex polytope $Pyr$ (a pyramid with a rectangular base).
	The approximate footprint of $B^r$ is illustrated in
\switchLabel{\refFig{w_box}.}{Figure 3.}
	See
\wafrORarxiv{
 \cite[Appendix~C]{zhang-chiang-yap:se3-arxiv:24}.
            }
{
Appendix~\ref{app:sep}.  
}
	\zhaoqiX{Move the descriptions to the appendix.}
	\ignore{
	\bitem
	\item $S_A=S_A(B^r)$ is the smallest ball centered at the c.g., and
		containing the $A$-footprint of the $8$ corners of $B^r$
	\item $S_B=S_B(B^r)$ is defined similarly to $S_A$,
		but using the $B$-footprint.
	\item $Cyl=Cyl(B^r)$ is the right-cylinder whose union with $S_A$
		and $S_B$ will result in the convex hull of $S_A$ and $S_B$.
	\item $Cone_A=Cone_A(B^r)$ is the right-cone whose union with $S_A$
		will result in the convex hull of the origin $O$ and $S_A$.
	\item $Cone_B=Cone_B(B^r)$ is defined similarly to $Cone_A$
		but using $S_B$.
	\item $Pyr(B^r)$ is a pyramid with with apex at $O$
		a rectangular base such that its union with
		the $S_A, S_B, Cyl, Cone_A, Cone_B$ results in the convex hull
		of $O, S_A, S_B$.
	\eitem
	}
\cheeX{Find a way to make the theorem/lemma numbers consistent
with Appendix}
	In
\wafrORarxiv{
 \cite[Appendix~E]{zhang-chiang-yap:se3-arxiv:24}
            }
{
 Appendix~\ref{app:pf}  
}
%
        we prove the following:
	\bthm
	The approximate footprint of the Delta Robot
			is $\sigma$-effective where $\sigma=(2+\sqrt{3})<3.8$.
	\ethm
	\ignore{
	\blem
	The Delta Robot satisfies Axiom \Atwo\ with Lipschitz constant
			$L_0=1$.
	\elem
	}

	\bthmT{Correctness of Delta Robot Planner}{delta}
	Our SSS planner for the Delta Robot is
	resolution exact with constant
		$K=4\sqrt{6}+6\sqrt{2}< 18.3$.
	\ethmT
	Note that such constant is not excessive as
	it just means that we need at most five additional
	subdivision steps ($2^5>18.3$) to reach any desired resolution.
\ssectL[sep1]{Parametric Separation Query and Boundary Reduction}
	Detecting collision \cite{lin-manocha:collision:18}
	between two Euclidean sets $A,C \ib Z$
	amounts to querying if their separation is positive:
	$\Sep(A,C)>0$.  We generalize it to the query ``Is $\Sep(A,C)>s$?''
	which we call a \dt{parametric separation query}
	(with parameter $s$).  Note that we
	need not compute the $\Sep(A,C)$ to answer this Yes/No query.
	The parametric query is useful because we are often interested in
	\dt{fat objects}, i.e., sets of the form $A\oplus Ball(s)$.
	Detecting their collision with $C$
	reduces to a parametric query on $A$ as in this simple lemma:

		\bleml{Sep_Minkow}
			Let $A,C\ib\RR^n$ be closed sets.
			Then $(A\oplus \ball(s))\cap C$ is empty iff $\Sep(A,C)>s$.
		\eleml
		\savespace{
		 This lemma follows from the remark that if $\|a-c\|>r$ for
		 all $a\in A, c\in C$, then $c\nin a+B$ and
		 hence $C\cap A\oplus B=\es$.
		}
	
	In this and the next two subsections,
	we discuss techniques that are used to reduce
	the parametric separation query into ultimately explicit
	and implementable subroutines.
	Initially, the sets $A, C$ in \refLem{Sep_Minkow}
	are the approximate footprint $A=\wtFp(B)$
		(see \refeq{apprxFp}), and $C=\Omega$.  
		Since $\wtFp(B)$ is a fat version of $\wtFp(B^r)$,
		we can replace $\wtFp(B)$ by $\wtFp(B^r)$.
		Using our method of features
\wafrORarxiv{
(\cite[Appendix~B]{zhang-chiang-yap:se3-arxiv:24}),
            }
{
(Appendix~\ref{app:sss}),   
}
%
                we can replace $C$ by a feature $f$ of $\partial\Omega$.
		\dt{Remark:} this technique could be used to simplify
		similar computations in the rod robot in \cite{rod-ring}.
	
		Next, we address the problem of computing
		the separation $\Sep(A,B)=\inf\set{\|a-b\|: a\in A, b\in B}$
		between two closed semi-algebraic sets $A,B\ib\RR^3$.
		Note that $A$ is \dt{semi-algebraic} means that 
		it is the set of points that satisfy
		a set of equations and/or inequalities.
		If only equations are used, then $A$ is \dt{algebraic}.
		We say $A$ is \dt{simple} if there is a unique algebraic
		set $\olA$ such that $A\ib \olA$ and $\dim(A)=\dim(\olA)$.
		Call $\olA$ the \dt{algebraic span} of $A$.
		For instance, every feature $f\in\Phi(\Omega)$ is simple since,
		when $A$ is a point/line-segment/triangle,
		then $\olA$ is a point/line/plane ($\Omega$ is rational).
		But if $\dim(A)=3$ then $\olA=\RR^3$.
	
		\ignore{
			For instance, if $A=[\bfa,\bfb]$ is a line segment
			and $B=[\bfc,\bfd,\bfe]$
			a triangle, then $\partial A=\set{\bfa,\bfb}$
			and $\partial B=[\bfa,\bfb]\cup [\bfb,\bfc]\cup [\bfc,\bfa]$,
			$\olA$ is a line and $\olB$ is a plane.
		}%
	
		For any two closed sets $A,B$, let $\cp(A,B)$
		be the \dt{closest pair} set of
		$(\bfa,\bfb)\in A^{\circ}\times B^{\circ}$
	\zhaoqiX{we instead define $\cp(A,B)$ as the closest
	pair of $\olA\times\olB$}
		such that $(\bfa,\bfb)$ is a locally closest pair.
		Here $A^{\circ}$ is the relative interior of $A$ in $\olA$
	\zhaoqiX{relative interior}
		(e.g., if $A$ is a closed line segment,
		$A^{\circ}$ is a relatively open line segment).
		Using the algebraic spans $\olA$ and $\olB$,
		the set $\cp(A,B)$ is (generically) contained in a finite
		zero-dimensional algebraic set $S$.  Then
		$\cp(A,B) = S \cap (A^{\circ}\times B^{\circ})$.

\ignore{%
	Another example:
	say $A$ is a disc, $B$ is a triangle.

	Let us do $Q_A>s$ first:
	$\Sep(A,\partial B)>s$ i.e.,
		BigAnd of Sep(disc,segment_i)>s (i=1,2,3)
			Sep(disc,point)>s (easy)
			Sep(interior(disc), segment)>s 
				Q_0 for (int(disc), seg)
				Q_A for (disc, point)
				Q_B for (circle, segment)

	VERY CONFUSING...

	Then $\olA$ and $\olB$ are planes.
	If they coincide or are parallel, this is degenerate.
	Else, they intersect in a line.
	Then 
		$\cp(\olA,\olB)=\set{ (\bfa,\bfa): \text{where }
					\bfa\in \olA\cap \olB}$
	
}%

		Let us illustrate this idea.
		Assume the algebraic span $\olA$ is the curve defined by the
		polynomial system $f_1=f_2=0$;
		similarly $\olB$ is the curve $g_1=g_2=0$.
		Then the closest pair $(\bfp,\bfq)\in
		A^\circ\times B^\circ$ is among the solutions to the system
			{\small
			\beql{system}
			\begin{array}{llll}
				0&=f_1(\bfp) =f_2(\bfp)\\
				0&=g_1(\bfq) =g_2(\bfq)\\
				0&=\bang{(\bfp-\bfq),\nabla f_1(\bfp)\times \nabla f_2(\bfp)}\\
				0&=\bang{(\bfp-\bfq),\nabla g_1(\bfq)\times \nabla g_2(\bfq)}
			\end{array}
			\eeql
			}%
		where $\nabla f_i$ is the gradient of $f_i$,
		$\bfu \times \bfv$ and $\bang{\bfu,\bfv}$ are the
		cross-product and dot product of $\bfu,\bfv\in\RR^3$. 
		Note that \refeq{system} is a square
		system in $6$ unknown variables ($\bfp,\bfq$)
		and generically has finitely many solutions.
		We say $(A,B)$ is \dt{degenerate} if
		the system has infinitely many solutions.
		The degenerate case is easily disposed of.   
		Other examples of such computation are given in
\wafrORarxiv{
 \cite[Appendix~C]{zhang-chiang-yap:se3-arxiv:24}.
            }
{
Appendix~\ref{app:sep}.  
}
%
%
		Using $\cp(A,B)$, we now have a simple
		``reduction formula'' for $\Sep(A,B)$:	

	\blemT{Boundary Reduction Method}{reduction}
		Let $A\ib\RR^3$ be a simple closed semi-algebraic set,
		and $f$ be a feature.  Assume $A\cap \closure{f}=\es$
		where $\closure{f}$ is the closure of $f$.
		Then $\Sep(A,\closure{f})>s$ iff
				$$(Q_0>s)\land (Q_f>s)\land (Q_A>s)
				$$
		where \hspace*{4mm}
			$Q_0\as \min\set{\|\bfa-\bfb\|: (\bfa,\bfb)\in \cp(A,B)}$,
		\Ldent[14]
				$Q_f\as \Sep(\partial A, f)$,
		\Ldent[14]
				$Q_A\as \Sep(A,\partial f)$.
		\\
		By definition, $Q_0=\infty$ if $\cp(A,B)$ is empty,
		and $Q_A=\infty$ if $f$ is a corner.
		\elemT
		\savespace{
			SHOULD WRITE OUT THE PROOF:

		In proof, since $A,\closure(f)$ are disjoint closed sets,
		$\Sep(A,\closure{f})$
		has a witness $(\bfa,\bfb)\in A\times \closure{f}$
		where $\|\bfa-\bfb\|=\Sep(A,\closure{f})$.  If the witness
		does not come from $\cp(A,f)$, then either $\bfa\in\partial A$
		or $\bfb\in\partial f$.  
		}
	
		This lemma reduces the parametric query to checking
		$Q_i>s$ for all $i=0,A,f$.  
				Note that $Q_0>s$ can be reduced to solving
				a system like \refeq{system}.
		By the method of features,
			checking if $\Sep(A,\Omega)>s$
		can be reduced to
			checking if $\Sep(A,f)>s$
		for all features $f\in\Phi(\Omega)$.
		Inevitably, we check all the $(i-1)$-dimensional features
		before checking the $i$-dimensional features $(i=1,2$).
		Therefore, in application of this lemma, we would already
		know that $Q_A>s$ is true.
		Ultimately, the query reduces
		to an easy computation of $\Sep(\bfa,f)>s$ or
		$\Sep(A,\bfa)>s$ where $\bfa$ is a point.
		This technique had been exploited in our work,
		but becomes more important as the primitives becomes
		more complex. See its application in the next subsection
		and in
\wafrORarxiv{
 \cite[Appendix~C]{zhang-chiang-yap:se3-arxiv:24}.
            }
{
Appendix~\ref{app:sep}. 
}
	\ignore{ %
		\blemT{Disc Lemma}{disc}
			Let $D\ib \RR^3$ be a disc and $f$ a boundary feature.
			Then $\Sep(D,f)>r$ has the following reduction....
		\elemT
	}%
\ssectL[sigma2]{On the $\Sigma_2$ Decomposition Technique}
	The reduction technique of \refLem{reduction}
	does not work when $A$ is a complex 3-dimensional object
	like our approximate footprint.  More precisely, the reduction
	requires us to characterize the various semi-algebraic patches
	that form the boundary of $A$.  Instead, we use
	a different approach based on expressing
	$A$ as a $\Sigma_2$-set as first introduced in \cite{rod-ring}.

\yijenX{\\(The paragraph below is newly added.)\\}
		First, we say that a set $B\ib\RR^3$ is \dt{elementary} if
	$B=\set{\bfx\in \RR^3: f(\bfx)\le 0}$ for some
	polynomial 
\yijenX{integer polynomial (i.e., polynomial with integer coefficients)}
	$f(X,Y,Z)$ of total degree at most $2$, and
	the coefficients of $f$ are algebraic numbers.
	Thus elementary sets
	include half-spaces, infinite cylinders, doubly-infinite
	cones, ellipsoids, etc.
	In our Delta robot, we will show that
	the algebraic coefficients of $f$ are degree $\le 2$;
	by allowing a small increase in the effectivity constant,
	we can even assume degree $1$ (i.e., $f(X,Y,Z)$ has
	integer coefficients).
	\ignore{ 
			Why degree $2$?   
			Well, we need a ball of radius $R=r_0+r_A$
			where $r_0, r_A$ are square-roots.
			The polynomial equation is $x^2+y^2+z^3=R^2=
			r_0^2+r_A^2 + 2r_0r_A$.   But $r_0r_A$ is
			degree $2$ (not degree $4$) because both
			are square roots.
			BUT BE CAREFUL -- we have done coordinate transformations,
			so it may not be obvious that $r_0r_A$ is degree $2$.
	}%
	Next, a \dt{$\Pi_1$-set} is defined as a
	finite intersection of elementary sets, and
	a \dt{$\Sigma_2$-set} is a finite union of $\Pi_1$-sets.
        So $A$ is a $\Sigma_2$-set if it can be written as
				{\small
				$$ A= \bigcup_{i=1}^m \bigcap_{j=1}^n A_{ij}$$
				}%
    where each
        $A_{ij}$ is an elementary set, and each $A_i = \bigcap_{j=1}^n
        A_{ij}$ is a $\Pi_1$-set.
        We allow $A_{ij}=\es$ to simplify notations.  
	The simple double loop below can answer the question: ``Is
	$f\cap A$ empty?''
\cheeX{It is not hard to see that our approximate footprint $\wtFp(B)$
	is a $\Sigma_2$-set.}
\wafrORarxiv{
In~\cite[Appendix~C]{zhang-chiang-yap:se3-arxiv:24}
            }
{
In Appendix~\ref{app:sep}  
}
	we show that our $\wtFp(B)$
	is a $\Sigma_2$-set.

	{\small\progb{
		\lline[-2] $\Sigma_2$-Collision Detection($f,A$):
		\lline[0]	Input: $f$ and $A=\bigcup_{i=1}^m\bigcap_{j=1}^{n}A_{ij}$.
		\lline[0]	Output: $\success$ if $A\cap f=\es$, $\failure$ else.
		\lline[5] For $i=1$ to $m$
		\lline[10] $R \ass f$
		\lline[10] For $j=1$ to $n$
		\lline[15] $R\leftarrow R\cap A_{ij}$ \qquad (*)
		\lline[15] If $R=\es$ break \Commentx{exit current loop}
		\lline[10] If $R\ne\es$, return $\failure$
		\lline[5] Return $\success$
		}
		}

	The step (*) maintains $R$ as the intersection of $f$ with
	successive primitives.  If $f$ is a point or a line segment, this
	is trivial.  When $f$ is a triangle, this could still be solved in
	our previous paper for rod and ring robots~\cite{rod-ring}.
        But the present
	$\pA \pO \pB$ robot requires us to maintain a planar set bounded by
	degree $2$ curves; this requires a non-trivial algebraic algorithm.
    We do not consider this ``explicit''. 
	Our solution is to explicitly write
			$A=\bigcup_{i=1}^m A_i$
	where each
		$A_i=\bigcap_{j=1}^n A_{ij}$
    has a very special form,
	namely, a convex and bounded $\Pi_1$-set of the following types:
		\beql{primitive}
		\text{right cylinder, right cone,
			right frustum, convex polyhedron.}
		\eeql
	By \dt{right} cylinder, we mean that it is obtained by intersecting
	an infinite cylinder with two half-spaces whose bounding planes
	are perpendicular to the cylinder axis.
	The notion of right frustum is similar, but using
	a doubly-infinite cone instead of a cylinder.
	Thus the two ``ends'' of a right cylinder and a right frustum are 
	bounded by two discs, rather than
	\yijenX{rather than}
    general ellipses.  A right cone is a
    special case of a right frustum when one disc is just a single
    point.

	We call the sets in \refeq{primitive}
	\dt{special $\Pi_1$-sets}.  A finite union of special $\Pi_1$-sets
	is called a \dt{special $\Sigma_2$-set}.
	While the above $\Sigma_2$-collision detection does not extend to
	parametric queries, this becomes possible with 
	special $\Sigma_2$-sets:

	\bthmT{Parametric Special $\Pi_1$-set queries}{special}
		\ \\
		There are explicit methods
		for parametric separation queries of the form
					``Is $\Sep(P,f)>s$?''
		where $P$ is a special $\Pi_1$-set and $f$ is a feature.
	\ethmT

	\dt{REMARK:}
	In
\wafrORarxiv{
 \cite[Appendix~C]{zhang-chiang-yap:se3-arxiv:24},
            }
{
 Appendix~\ref{app:sep}, 
}
	we introduce a further simplification
	to show that $\wtFp(B)$ is the union of 
	``very special'' $\Sigma_2$-sets which are
	defined by polynomials of degree $2$ whose coefficients have
	algebraic degree $2$.
\ignore{
	\begin{propo}
	Given two features, the minimum distance between the two features are
	reached either at the boundary, or at the projections.
	\end{propo}
	\begin{proof}
	Suppose that the minimum distance are not reached at the boundary,
	then it is reached at the interior of the feature. In this case,
	suppose it is not reached at the projections, then the minimum
	distance is reached in a local minimum. However, for the points other
	than the projections, the gradient of the distance function is not
	$0$, hence there is no local minimum. This contradiction shows that
	it is reached at the projections, and hence the projections are in
	the interior.
	\end{proof}
	}%

	\ignore{
	VII. Given a point $v$ and a circle $c$, suppose that the center of
	$c$ is $o$, the radius is $r$, let $w$ be the projection of $v$ onto
	$c$. If $w\in c$, $\textrm{Sep}(\{v\},c)=d(v,w)$, otherwise,
	$\textrm{Sep}(\{v\},c)=\textrm{Sep}(\{v\},\partial c)$. To find this
	value, we take the plane generated by $vwo$ and intersect with
	$\partial c$ and $u$. Then $d(u,v)$ is the value. In summary,
		\[\textrm{Sep}(\{v\},c)=\left\{\begin{array}{lc}d(v,w) & w\in c\\
		\textrm{Sep}(\{v\}, \partial c) & w\notin c\end{array}\right.,\]
	where $w$ is the projection of $v$ onto $c$.

	VIII. Given a segment $g$ and a circle $c$, finding the separation
	between them is similar to the process of a segment with a triangle.
	When $g$ is not parallel to $c$, let $v$ be the intersection between
	the line of $g$ and the plane of $c$. When $v\in g$, then $\partial
	g$ is at the two different sides of $c$. Hence, if $v\in c$, then
	$\textrm{Sep}(g,c)=0$. Otherwise,
	$\textrm{Sep}(g,c)=\textrm{Sep}(g,\partial c)$. To find this value,
	we take the plane generated by $g$ and the the center of $c$, and
	intersect it with $\partial c$ at $w$. Then $\textrm{Sep}(\{w\},g)$
	is the value. When $v\notin g$, then $\partial g$ is at the same side
	of $c$. Since $g$ is not parallel to $c$, among the two points in
	$\partial g$, there must be one that is nearer than the other to the
	plane of $c$. Let the nearer end-point be $g_1$, and the projection
	from $g_1$ to $t$ is $w_1$. If $w_1\in c$, then
	$\textrm{Sep}(g,c)=d(g_1,w_1)=\textrm{Sep}(g_1,c)$. If $w_1\notin c$,
	then similar to the case when $v\in g$ and $v\notin c$,
	$\textrm{Sep}(g,c)=\textrm{Sep}(g,\partial c)$ by the computation. As
	a summary, we have
		\[\textrm{Sep}(g,c)=
			\left\{\begin{array}{lc}0 & v\in g, v\in c\\
			\textrm{Sep}(g_1,c) & v\notin g, w_1\in c\\
			\textrm{Sep}(g,\partial c) & otherwise\end{array}\right.,\]
	where $g_1$ is the nearer end-point and $w_1$ is its projection onto
	$c$. When $g$ is parallel to $c$, let $w_1,w_2$ be the projections of
	$\partial g$ onto $c$. If $w_1\in c$ or $w_2\in c$, then
	$\textrm{Sep}(g,c)=\textrm{Sep}(\partial g,c)$, otherwise,
	$\textrm{Sep}(g,c)=\textrm{Sep}(g,\partial c)$, since in this case no
	matter how we choose the minimum projection, we can always trace it
	along $g$ and reach the boundary of $c$. As a summary, we have
		\[\textrm{Sep}(g,c)=\left\{\begin{array}{lc}\textrm{Sep}(g,\partial
		c) & w_1,w_2\notin c\\ \textrm{Sep}(\partial g, c) &
		otherwise\end{array}\right.,\]
	where $w_1,w_2$ are the projections from $\partial g$ to $c$.

	IX. Given a triangle $t$ and a circle $c$, finding the separation
	between them is similar to the process of between two triangles. Let
	the center of $c$ be $o$. When $c$ is not parallel to $t$, if $c\cap
	t\neq\emptyset$, then $\textrm{Sep}(c,t)=0$, otherwise
	$\textrm{Sep}(c,t)=\min\{\textrm{Sep}(\partial
	c,t),\textrm{Sep}(c,\partial t)\}$. To find $\textrm{Sep}(\partial
	c,t)$, we consider the plane containing $t$ to be $p$, and the plane
	containing $c$ to be $q$. Let $l=p\cap q$. We define the plane
	perpendicular to $l$ and containing $o$ by $r$. Then let $g=r\cap c$,
	we have $\textrm{Sep}(\partial c,t)=\textrm{Sep}(\partial g,t)$. When
	$c$ is parallel to $t$, we consider the projection of $t$ onto the
	plane of $c$. If there is overlapping, then the separation is the
	distance between the two planes. If there is no overlapping, then the
	separation is reached on the boundary. No matter which case, the
	separation is always $\min\{\textrm{Sep}(o,t),\textrm{Sep}(c,\partial
	t)\}$. As a summary,
		\[\textrm{Sep}(c,t)=\left\{\begin{array}{lc}0 & c\cap
		t\neq\emptyset\\ \min\{\textrm{Sep}(\partial
		c,t),\textrm{Sep}(c,\partial t)\} & c\notparallel t\\
		\min\{\textrm{Sep}(o,t),\textrm{Sep}(c,\partial t)\} & c\parallel
		t\end{array}\right..\]
	}

\ignore{
\subsection{Collision Detection}
	Now we give the collision detection criteria for our finite primitives. For any feature $f$, we are always trying to find if a primitive $S$ intersects with $\textrm{Ball}(0,d)\oplus f$ or not.
	
	All collision detection are based on
	\[(f\oplus\textrm{Ball}(0,d))\cap S\neq\emptyset\Leftrightarrow\textrm{Sep}(f,S)\leq d.\]
	If $S=S_1\cup S_2$, then $\textrm{Sep}(f,S)=\min\{\textrm{Sep}(f,S_1),\textrm{Sep}(f,S_2)\}$.
}
%





\ignore{
	%
	\sect{Introduction}
	20Nov23: To effectively deal with adjacencies of boxes
	in higher dimensions, we need a calculus
	of "indices" for subdivision trees.
	
	Please read my paper \cite[Section 3.1]{bennett-yap:smooth:17}
	and also Appendix D.\P 3.
	
	However, for our application in SE(3), I develop
	the following additional notations.
	It is not all complete yet, but try to fill in any gaps!
} 

\sect{Subdivision in $\wh{SE}(3)$: Adjacency and Splitting}
	%
	We now address subdivision in the space
	$\wh{SE}(3)=\RR^3\times \wh{SO}(3)$.
	Let a box $B$ in $\wh{SE}(3)$ be decomposed
	as $B^t\times B^r$ where $B^t\in \intbox\RR^3$
	and $B^r\in \intbox\wh{SO}(3)$.
	$B^t$ is standard, but $B^r$ is slightly involved as shown next.
	Given an initial box $B_0= B_0^t\times \wh{SO}(3)$,
	the SSS algorithm will construct a subdivision tree $\TTT=\TTT(B_0)$
	that is rooted at $B_0$.  The leaves of $\TTT$ represent
	the (current) subdivision of $B_0$.  We need an efficient
	method to access the adjacent boxes in the (current) subdivision.
	The number of adjacent boxes is unbounded;
	instead, we maintain only a bounded number
	of ``principal'' neighbors from which we can access all the
	other neighbors.
	For $\RR^n$, this has been solved in \cite{aronov+3} using
	$2n$ principal neighbors.  We will show that for $\wh{SO}(3)$ boxes,
	$8$ principal neighbors suffice.  So 14=6+8 principal
	neighbors suffice for boxes in $\TTT(B_0)$.

	\ignore{%
		Conceptually, we can think of
		two \dt{template trees}, $\TTT^t=\TTT^t(B_0^t)$ and
		$\TTT^r=\TTT^r(\wh{SO}(3))$.   Each box $B\in \TTT(B_0)$ can be
		viewed as a pair $(B^t,B^r)$ where $B^t\in \TTT^t$ and
		$B^r\in\TTT^r$.
	}%

	It remains to discuss principal neighbors in $\wh{SO}(3)$.
	Following Section \ref{ssec:chart} and
\switchLabel{\refFig{so3-model},}{Figure~2,}
	$\wh{SO}(3)$ can
	be regarded as the union of four cubes,
	     $\wh{SO}(3) = \cup_{i=0}^3 C_i$
	where $C_i\as \set{(a_0\dd a_3)\in [-1,1]^4: a_i=-1}$.
	The indices in $(0,1,2,3)$ will also be identified
	with $(w,x,y,z)$: thus $C_0 =C_w$, $C_1=C_x$, etc.
	Let $d \in \{\pm e_0 \dd \pm e_3 \}$ identify one of the 8 semi-axis
	directions (here $e_i$ denotes the $i$-th standard
%
%
        basis
	vector). If two boxes $B$ and $B'$ are neighbors,
	there is a unique $d$ such that $B'$ is
	adjacent to $B$ \dt{in direction $d$}, denoted by $B \dder{d} B'$.
	In general, $B'$ is not unique for a given $B$ and $d$.
	See
        %
\wafrORarxiv{
  \cite[Appendix~D.1]{zhang-chiang-yap:se3-arxiv:24}
            }
{
  Appendix~\ref{sec:box-adj}   
}
        for details.       
	
	We now describe
	the subdivision tree rooted at $\wh{SO}(3)$:
	the first subdivision is special, and splits $\wh{SO}(3)$ into $4$
	boxes $C_i$ for $i = 0, 1, 2, 3$.  Subsequently, each box is split
	into $8$ children in an ``octree-type'' split.
	Each non-root box $B$ maintains $8$ \dt{principal neighbor pointers},
	denoted $B.d$ ($d\in \set{\pm e_0\dd \pm e_3}$). 
	However, only 6 of these pointers are non-null:
	if $B\ib C_i$ then $B.d$ is null iff $d=\pm e_i$.
		The non-null pointer $B.d$ points to the
		\dt{principal $d$-neighbor} of $B$, which is defined 
		as the box $B'$ that is a $d$-neighbor of $B$
		whose depth is {\em maximal}
		subject to the restriction that
		$depth(B') \le depth(B)$.
		Note that $B'$ is unique and has size at least that of $B$.
	The non-null pointers for $B$ are set up according to two cases.
	\dt{Case} $B=C_i$:
		Each $B.d = C_j$ if $d=\pm e_j$ and $j\ne i$
		(see Fig.~\ref{fig:so3-model}).
	\dt{Case} $B\ne C_i$:
	Three of the non-null pointers of $B$ point 
	to siblings and the other three point to non-siblings,
        and are determined
	as in
\wafrORarxiv{
  \cite[Appendix~D]{zhang-chiang-yap:se3-arxiv:24}.
            }
{
  Appendix~\ref{sec:adj}.  
}
%
 \ignore{
	}%
\ignore{ 
\subsection{Subdivision Data Structure for $\wh{SE}(3)$}
        Now
	we discuss our overall data structure for the subdivision of
	$\wh{SE}(3) = \RR^3
	\times \wh{SO}(3)$.
%
%
	Initially, we have one leaf box $B_0 = B^t_0 \times B^r_0$,
        where its translational box $B^t_0 \ib \RR^3$ is a box in
        $\RR^3$ and its rotational box $B^r_0$ is $B^r_0 = \wh{SO}(3)
        = \cup_{i=0}^3 C_i$, consisting of 4 boxes $C_i$ as
        discussed previously. Therefore at the top level, we have a
        box $B^t_0$ in the translational space and 4 boxes $C_i$ in
        the rotational space.  We then perform repeated subdivisions
        on leaves to create our subdivision data structure. When we
        split a leaf box $B = B^t \times B^r$, we either split $B^t$
        in the translations space (called {\bf T-split}), or split
        $B^r$ in the rotational space (called {\bf R-split}).  A
        T-split on $B$ splits $B^t$ into 8 child boxes, where $B^r$
        stays the same.  Before performing any R-splits, our boxes all
        have the full rotational range and we do not need to store the
        boxes of $C_i$ explicitly.  An R-split on $B$ keeps $B^t$
        the same; an initial R-split subdivides $\wh{SO}(3)$ into 4
        boxes $C_i$ for $i = 0,\cdots, 3$.  After that, an R-split
        on $B$ splits $B^r$ (a box in the octree of $C_i$ for some
        $i$) into 8 child boxes, using the scheme discussed
        previously.
	Typically we first perform a sequence of T-splits, then we interleave
	T-splits and R-splits.
\ignore{

        It would be interesting to devise a good
	strategy on how to interleave T-splits and R-splits.
}

	We remark that Nowakiewicz \cite{nowakiewicz:mst:10} also
	discusses subdividing the translational and rotational boxes
	separately, as well as subdividing the cubic model. However,
        %
        the method does not classify boxes, and does not compute or use the
	adjacency information of the boxes.
\ignore{
        the
	method eventually takes sample configurations (at the corners or
	centers) in subdivision boxes and is actually a sampling-based method.
	It does not classify the boxes and does not compute or use the
	adjacency information of the boxes.
        }
        
}

\sect{Conclusion}
\noindent\dt{Limitations.}
	We are currently in the midst
\zhaoqiX{almost the end}
	of implementing this work.
	We are not aware of any theoretical limitations, or
	implementability issues. 
	One concern is how practically efficient is the current design
	(cf. \cite[Sect.~1, Desiderata]{rod-ring}).
	It is possible that additional techniques (mostly about
	searching and/or splitting strategies) may be needed
	to achieve real-time performance.

\noindent\dt{Extensions and Open Problems.}
	Our SSS path planner for the Delta Robot is easily generalized to any
	``fat Delta robot'' defined as the Minkowski sum
	$\pA \pO \pB \oplus B(r)$ of $\pA \pO \pB$ with a ball $B(r)$.
	Here are two useful extensions that appear to be reachable:
	(1)
	Extensions would be to ``complex $SE(3)$ robots'', where
	the robot geometry is non-trivial.
	In principle, the SSS framework allows such extensions
	\cite{sss2}. 
	\savespace{
	For the case of $SE(2)$ robots, this was
	demonstrated in
	\cite{zhou-chiang-yap:complex-robot:20}.
	}
	(2)
	Spatial 7-DOF Robot Arm.   In real world
	applications, robot arms normally need more than 6 degrees
	of freedom.   In this case, the configuration space
	is a product of 2 or more rotational spaces.

\wafrORarxiv{
            }
            {
	Beyond these, extremely challenging problems are
	snakes in 2D, and humanoid robots.
	Other principles need to be invoked.  Of course, rigorous
	and implementable subdivision algorithms for
	dynamic and non-holonomic motion planning are completely open.
            }
	\bibliographystyle{abbrv} 
	\bibliography{test,st,yap,exact,geo,alge,math,com,rob,cad,algo,visual,gis,quantum,mesh,tnt,fluid} 
\newpage
\appendix

\sectL[nopath]{Appendix: Review of the NOPATH Literature} \label{app:nopath}

	We feel that the repeated misunderstanding about the
	fundamental ``Zero Problem'' \cite{sharma-yap:crc:16}
	of path planning calls for a closer overview of the literature.
	As the problem goes by different names
	(``disconnection proof'' \cite{basch+3:disconnection:01},
	``non-existence of path'' \cite{zhang-kim-manocha:path-non-existence:08},
	``infeasibility proof'' \cite{li-dantam:infeasibility:23}, etc),
	we will simply call it the NOPATH problem.
	In its starkest form, NOPATH is just a decision problem,
	with a YES/NO answer. 
	Classical path planning extends this
	problem by asking for a path (if YES), and
	a simple "NO" otherwise.  Call this the FINDPATH problem. 
	In recent years (e.g.,
	\cite{sung-stone:infeasibility:23,li-dantam:infeasibility:23}),
	we see FINDPATH extended by requiring
	an ``infeasibility proof'' in case of a "NO" answer.\footnote{
		Note that the concept of providing a proof is always
		relative to some ``proof checker'' and this could vary
		dramatically depending on how
		powerful a proof checker one has in mind.
		E.g., for a proof of NOPATH,  one often thinks
		of a cut that separates the start and goal
		configurations in the adjacency graph (so we need
		a ``cut checker'').  But one could
		also output the entire adjacency graph, and the checker
		can verify that there is no path.
	}
	Since the NOPATH problem is embedded in FINDPATH or its extension,
	we can evaluate all path planners with respect to
	their completeness for NOPATH.

	Before reviewing the literature, there are two preliminary remarks.
	First of all, the NOPATH problem is clearly solved
	by exact path planning, so we may only focus on algorithms based on
	numerical approximations and sampling.
	Second, we note that some algorithms only solve
	``promise problems'' (e.g., \cite{goldreich:promise-problems:06}).
	In such problems, the inputs in addition to being well-formed,
	must satisfy some semantic conditions
	(``promise'').  A \dt{promise algorithm} is one whose
	correctness depends on such promises.
	Below, we will see such promise algorithms.
\cheeX{YOU MUST REDEFINE "wafrORarxiv" macro to in
	lmac.tex, depending on whether you want to produce
	a version for WAFR or for ARXIV.} 
	If the promise implies the existence of paths,
	then clearly they do not solve the NOPATH problem.

	The earliest paper that appears to offer a
	solution to NOPATH is Zhu and Latombe
	\cite{zhu-latombe:hierarchical-constraint:90,zhu-latombe:hierarchical:91},
	whose ``hierarchical framework'' shares many features
	of our SSS framework. 
	They made a nice observation that if there
	is no path in the adjacency graph of cells
	that are either EMPTY or MIXED\footnote{
		Our classification of boxes as \free/\stuck/\mixed\ in SSS
		corresponds to their
		\dt{labeling} of cells as EMPTY/FULL/MIXED.
                },
        then it constitutes a proof of NOPATH.  
	But they do not offer a complete method to detect
	NOPATH;
	see also Barbehenn and Hutchinson
	\cite{barbehenn-hutchinson:single-source:95}.
	Unfortunately, the non-termination issue persists.

	Non-Halting Example:
	Consider a planar disc robot of radius 1 and the
	obstacle set $\Omega=\set{(x,0):|x|\ge 1}$ where the robot must
	move from $\alpha=(0,-1)$ to $\beta=(0,1)$. 
	Then every box that contains $(0,0)$ will be MIXED.
	This causes the halting problem for any subdivision approaches
	(and, a fortiori, for any sampling approaches).

	An early reference for infeasibility proofs is Basch et
	al.~\cite{basch+3:disconnection:01} who aimed to
	find proofs when a robot cannot move through a ``gate'' in a 3D wall
	(they do not claim completeness).
	The exact solution for a 2D gate was
		first solved in \cite{yap:chair:87}.
	Next, Zhang et al.~\cite{zhang-kim-manocha:path-non-existence:08}
	gave infeasibility proofs by giving
	a sufficient criterion for classifying a
	cell as FULL (i.e., \stuck).
	Their heuristic is clearly incomplete.
	More recently, Li and Dantam
	\cite{li-dantam:infeasibility:23,li-dantam:infeas-learn:21} 
	aimed to find infeasibility proofs by learning and other techniques.
	But they only have a promise algorithm, conditioned on the
	input having the $\veps$-blocked property
		\cite[p.2, Section III.A]{li-dantam:infeasibility:23}.
	Here $\veps>0$ is a hyperparameter
		\cite[p.7, Section VII.B]{li-dantam:infeasibility:23}
	that the user must provide.
	Note that the concept of $\veps$-blocked property is very close
	to our idea of $\veps$-exactness.  The difference is that
	they make this property a ``promise'' while our SSS algorithms
	take $\veps$ as an input.
	
	Li and Dantam \cite{li-dantam:infeasibility:23} proposed
	the idea to dovetail a PRM planner with an ``infeasibility prover''
	(see \cite[Figure 1]{li-dantam:infeasibility:23}).
	This allows sampling based methods to sometimes produce
	NOPATH outputs. 
	More generally, the concept of providing ``infeasibility proofs''
	is only needed for incomplete algorithms
	(such as sampling methods). A complete algorithm 
	does not need to provide proofs.  In particular,
	exact algorithms or our SSS algorithms
	do not need to provide such proofs.
	
	Another attempt at infeasibility proofs is
	Sung and Stone \cite{sung-stone:infeasibility:23},
	but in the limited setting of a prior roadmap.

	\dt{What is the notion of an ``infeasibility proof''?} 
	How is an ``infeasibility proof'' different from having
	a rigorous algorithm that simply outputs ``\nopath''?
	A ``proof'' that any path planning instance
			$I=(B_0,\Omega,\alpha,\beta,\veps)$
	is infeasible requires a suitable ``family of proof systems''
			$\calS \as \set{S_i: i=1,2\ldots}$
	and an (efficient) proof checker $PC$
	with the property that for any input instance $I$,
	\benum[(i)]
	\item If $I$ is infeasible, there exists some $S_i\in\calS$ such that
		$PC(I,S_i)=\true$; and
	\item if $I$ is feasible, then for all $S_i\in\calS$,
		$PC(I,S_i)=\false$.
	\eenum
	A careful description of $\calS$ and $PC$ is largely missing
	in the literature.
	The point of $PC$ is that the checker can efficiently check whether
	$I$ is infeasible with the help of some $S_i$.

	\dt{Solution of the Halting Problem via Sampling.}
	We now discuss an approach that in principle solves the
	Halting problem.
	Path planners that accept $\veps>0, \delta>0$
	as part of the input are clearly trying to do approximation.
	Thus,
	Dayan et al.~\cite{dayan-solovey-pavone-halperin:near-opt:21}
	defined a graph
	$G=G(\calX,r,x^s,x^g)$ where
			$\calX\ib \cspace$, $r>0$ and $x^s,x^g\in \cspace$.
	An edge $(u,v)$ of $G$ represents a feasible path from $u$ to $v$.
	Given $\veps,\delta$, they construct $\calX$ and $r>0$
	such that $(\calX,r)$ is $(\vareps,\delta)$-complete in this sense:
		the inequality
		\beql{optpath}
			d_G(x^s,x^g)\le (1+\veps)OPT_\delta\eeql
	holds.  Here $d_G(u,v)$
			is the length of the shortest path in $G$,
	and $OPT_\delta$ is the length of the
		shortest path among the $\delta$-clear paths.
	They did not discuss the possibility that $OPT_\delta=\infty$
	(i.e., no ($\vareps,\delta$)-path).   In this case, \refeq{optpath}
	still holds trivially, since both sides are infinite.

	In principle, this $(\vareps,\delta)$ sampling approach
	can halt when there is no $\delta$-clear path.  
	In this case, it halts after an exhaustive (or sufficiently dense)
	sampling of the configuration space.  Although of theoretical
	interest, it cannot be practical.  Moreover, the current formulation
	of Dayan et al.~\cite{dayan-solovey-pavone-halperin:near-opt:21}
        is limited to disc robots in Euclidean space.
	In contrast, our resolution-exact approach
	can solve this halting problem in an adaptive manner,
	thus escaping exponential behavior in practice.  Also our
	formulation addresses more general configuration spaces.
\sectL[sss]{Appendix: Soft Subdivision Search} \label{app:sss}
	  We review the elements of SSS Framework 
	  \cite{wang-chiang-yap:motion-planning:15,sss2}.
	  Recall that in Section 2.1, we introduced the four spaces
	  for path planning: the configuration space $X$, free space $Y$,
	  physical space $Z$, and computation space $W$ given by
				  $$
				  W=\RR^d,\quad
				  X=\cspace(R_0),\quad
				  Y=\cfree(R_0,\Omega),\quad
				  Z=\RR^k.
				  $$
	  
\ssect{Soft Predicates}
	Given $A,B\ib Z$, their \dt{separation} is defined as
	$\Sep(A,B)\as\inf_{\bfa\in A,\bfb\in B}\|\bfa-\bfb\|$.
	 A robot $R_0$ is defined by a continuous
	 \dt{footprint map} $\Fp: X\to 2^{Z}$. 
		This map may be regarded as
		a generalization\footnote{
			Typically, a fixed point $\calA$ on the robot
			is chosen and the footprint map $\Fp_\calA(\gamma)$
			is the location of $\calA$ in physical space.
			We generalize this to any set $\calS$ of
			points on the robot: if $\calS$ is the entire
			robot, we just write $\Fp(\gamma)$ instead of
			$\Fp_{\calS}(\gamma)$.
		} of the more well-known
		\dt{forward kinematic map}.    
	 Continuity of $\Fp$ comes from
	 the fact that $X$ and $2^Z$ are topological spaces where $2^Z$ has
	 the topology induced from
	 the Hausdorff pseudo-metric on subsets of $Z$.
	 \zhaoqiX{In my thesis, I'm planning to say a footprint map $\Fp$ is
	 a continuous map from a topological space $X$ to the power set of a
	 metric space $Z$ where the topology of $2^Z$ is given by its
	 Hausdorff distance. Then $X$ is called a configuration space and $Z$
	 is called a physical space. This is my new ideas and I'm trying to
	 formalize them or adapt them to examples in LaValle's book. 
	 }
	 Relative to an obstacle set $\Omega\ib Z$, its
	 \dt{clearance function} is 
	  $\Cl:X\to\RR_{\geq0}$ where $\Cl(x)\as \Sep(\Fp(x),\Omega)$.
	  We say $x\in X$
	  is \dt{free} iff $\Cl(x)>0$.
	  We define $Y$ as the set of free configurations in $X$.
	  A \dt{motion} is a continuous function $\mu:[0,1]\to X$.
	  We call $\mu$ a \dt{path}
	  if the range of $\mu$ lies in $Y$
	  (so $\mu(t)$ is free for all $t\in[0,1])$.
	  We also call $\mu(0)$ and $\mu(1)$ the start and goal
	  configurations of the path, and $\mu$ is a path from its
	  start to its goal configuration.
	  The \dt{clearance} of a path $\mu$ is $\min\set{\Cl(\mu(t)): t\in
	  [0,1]}$.  The
	  \dt{optimal clearance} between $x,y \in Y$ is the largest
	  clearance of any path from $x$ to $y$.

	  We now consider a somewhat non-intuitive concept of
	  ``essential clearance'' first introduced in
	  \cite{wang-chiang-yap:motion-planning:15}:
	  let $\mu$ be a path.  We say $\mu$ has \dt{essential clearance}
	  $C>0$ if there exists $t_0, t_1$ ($0\le t_0<t_1\le 1$)
	  such that for all $t\in [0,1]$:
	  	$$\Cl(\mu(t)) \clauses{ \ge C & \rmif\ t\in [t_0,t_1]\\
								< C & \elsE.}$$
	Note that having essential clearance $C$ means that,
	with the exception of an initial segment $[0,t_0)$ and
	a final segment $(t_1,1]$, the path has clearance at least $C$.
	If $t_0=0$ or $t_1=1$, this initial or final segment is null.
	The motivation for this definition is to make our resolution
	exact algorithms simpler, by not having to check such initial
	and final segments.  E.g, our SSS planner finds a path
	by discovering a \dt{channel}, i.e., a
	sequence of adjacent free boxes that
	connect the start and goal configurations $\alpha, \beta$.
	All the free boxes in the channel have widths $\ge \veps$.
	Then it is clear that we can find an path $\mu$ from $\alpha$ to $\beta$
	inside this channel whose essential clearance is $\ge \veps/K$
	where $K>0$ depends on the algorithm.  We cannot discount
	the possibility that the initial or final segment of $\mu$ may have
	clearance arbitrarily close to $0$.

	  \zhaoqiX{I moved the clearance function to here and
	  added the definition of optimal clearance. I also changed ``this''
	  in the definition of exact predicate into its specific term.}

	  The concept of a ``soft predicate'' is relative to some exact
	  predicate. Based on the clearance function, the \dt{exact
	  predicate} is $C:X\to\set{0,+1,-1}$ where $C(x)=0/+1$ (resp.) if
	  configuration $x$ is semi-free/free; else $C(x)=-1$.  The semi-free
	  configurations are those on the boundary of $Y$.  Call $+1$ and
	  $-1$ the \dt{definite values}, and $0$ the \dt{indefinite value}.
	  	$$C(x)=\clauses{+1 & \rmif\ x\in Y^\circ,\\
	  					0 & \eliF\ x\in \partial Y\\
	  					-1 & \elsE. }$$
	  \zhaoqiX{We should write these in the formula environment. For example, 
	  \[C(x)=\left\{\begin{array}{r l} +1 & x\in Y^\circ\\
	  0 & x\in \partial Y\\
	  -1 & x\in (X/Y)^\circ\end{array}\right.\]}
	  We can extend the definition to any set $B\ib X$:
	  for a definite value $v$, define $C(B)=v$ iff $C(x)=v$ for all $x$.
	  Otherwise, $C(B)=0$.

	  Let $\intbox W$ denote the set of $d$-dimensional boxes
	  in $W=\RR^d$.
	  In general, a box $B \in\intbox \RR^k$ is called a \dt{hypercube}
	  if $B=\prod_{i=1}^k I_i$ is the product of intervals $I_i$
	  of the same width.  Let $w(B)=\min_{i=1}^k w(I_i)$ where
	  $w([a,b])= b-a$.  The \dt{diameter} of any set $S\in\RR^n$ is the
	  		$\diam(S)\as \max_{\bfa,\bfb\in S} \|\bfa-\bfb\|_2$.
	  The \dt{aspect ratio} of $B$ is defined as
	  		$\rho(B)\as \diam(B)/w(B)$.

		In our application of $SE(3)$, $W=\RR^7$
		and each box $B\in\intbox W$ has the decomposition
				$B=B^t\times B^r$
		where $B^t\in\intbox \RR^3$, and $B^r\in\intbox \RR^4$.
		We assume that $B^t$ is a hypercube, and define the ``width''
		of $B$ to be $w(B) \as w(B^t)$.
	  %
	  %

	  For any box $B\in \intbox W$, 
	  let $C(B)$ as a short-hand for $C(\olmu(B))$ where
	  $\mu:X_\mu\to X$ is the homeomorphism from the ``square model''
	  $X_\mu$ to $X$.
	  \zhaoqiX{$X_\mu$ is $W$.}
	  A predicate $\wtC:\intbox W\to\set{0,+1,-1}$ is
	  a \dt{soft version of $C$} if
	  it is conservative and convergent.  \dt{Conservative} means that
	  if $\wtC(B)$ is a definite value, then $\wtC(B)=C(B)$.
	  \dt{Convergent} means that if for any sequence $(B_1,B_2,\ldots)$
	  of boxes, if $B_i\to p\in W$ 
	  as $i\to\infty$, then $\wtC(B_i)=C(p)$ ($=C(\olmu(p))$)
	  for $i$ large enough. 
	  \zhaoqiX{
	  	We should write as two conditions: \\
	  (Conservative)
	  \[\wtC(B)=C(B)\textrm{ if }\wtC(B)=\pm1\]
	  (Convergent)\\
	  for any sequence $(B_1,B_2,\ldots)$
	  of boxes, if $\iota(B_i)\to p\in X$ 
	  as $i\to\infty$, then $\wtC(B_i)=C(p)$ for $i$ large enough.
	  }
	  To achieve resolution-exact algorithms, we must ensure
	  $\wtC$ converges quickly in this sense:
	  say $\wtC$ is \dt{effective} if there is a constant
	  $\sigma>1$ such if $C(\sigma B)=\pm 1$ (i.e., definite)
	  then $\wtC(B)=C(B)$.
	  \zhaoqiX{
	  In our original definition, it is 
	  $\wtC(B)=C(B)\textrm{ if }C(\sigma B)=\pm1$. 
	  We should also directly write it in the formula form.
	  }
	  
\ssectL[cubic]{The Cubic Model $X_\mu$}
\ignore{
	By a \dt{(subdivision) chart} of $X$, we mean a function
	$h: B\to X$ where $B\in\intbox W$
	and $h$ is a homeomorphism between $B$ and its image $h(B)\ib X$.
	An \dt{(subdivision) atlas} of $X$ is a set $\mu=\set{\mu_t: t\in I}$
	for some finite index set $I$ such that each $\mu_t$ ($t\in I$)
	is a chart, and if $X_t\ib X$ is the image of $\mu_t$, then
	$\dim(\mu_t\inv(X_t\cap X_s))<d$ ($t\ne s$).
	From $\mu$, we can construct a \dt{tile model} of $X$, denoted
	$X_\mu$, that is homeomorphic to $X$ via a map
	$\olmu:X\to X_\mu$ (see Appendix \ref{app:sss}).
	Note that $\olmu$ is basically the inverse of the $\mu_t$'s:
	if $x\in B_t$, then $\olmu(\mu_t(x))=x$.
	}%

	For any set $S\ib \RR^m$, a \dt{subdivision} of $S$ is
	a finite set $\set{S_1\dd S_k}$ such that $S=\cup_{i=1}^k S_i$
	and $\dim(S_i\cap S_j)<\dim(S_i)$ for all $i\ne j$.
	We assume that $S_i$'s are nice sets for which the notion
	of dimension, $\dim(S_i)$ and $\dim(S_i\cap S_j)$,
	are well-defined.  In our applications,
	$S_i$'s are boxes.

	Let $X$ be a topological space and $I$ a finite index set.
	Recall the definitions of charts and atlases in
	Section~\ref{ssec:chart}.
	Let 
			$\mu=\set{\mu_t: t\in I}$
	be a \dt{subdivision atlas} of $X$.
	For each $t\in I$, let $\mu_t: B_t\to X$ 
	where $B_t\in\intbox\RR^m$ for some fixed $m$.
	We construct the space $X_\mu$ as the following
	quotient space: Let $X^+_\mu \as \uplus_{t\in I} B_t$
	the disjoint union of the $B_t$'s.
	Then $X_\mu$ is the quotient space $X^+_\mu/\sim$
	where $a\sim b$ for $a,b\in X^+_\mu$ iff $a\in B_s$
	and $b\in B_t$ implies $\mu_s(a)=\mu_t(b)$.
	Let $[b]$ denote the equivalence class of $b$.
	Finally, we can define the map $\olmu: X\to X_\mu$
	where $\olmu(x)=[b]$ iff $b\in B_t$ and $\mu_t(b)=x$.

	For $SO(3)$, we have the atlas
		$\mu=\set{\mu_w, \mu_x,\mu_y,\mu_z}$ 
	in which $\mu_t: B_t\to SO(3)$ ($t\in \set{w,x,y,z}$).
	The boxes $B_w,B_x,B_y,B_z$ are specially chosen so that
	$X_\mu$ is embedded in $\RR^4$. Obviously, $X_\mu$ has
	a non-Euclidean topology.
	  

\ssect{The SSS Framework}
	  An SSS algorithm maintains
	  a subdivision tree $\TTT=\TTT(B_0)$ rooted at
	  a box $B_0\ib \intbox W$.
	  \zhaoqiX{Not ``a'' box.}
	  \cheeX{We pretend it is a tree initially, because that is
	  	easy to understand first! We will fix this below.}
	  Each tree node is a subbox of $B_0$.
	  In Axiom \dt{(A1)}, we view $\expand(B)$ as a set of boxes
	  that represent a subdivision of $B$.
	  If $B\in\intbox\RR^m$ has dimension $p\le m$, the
	  \dt{canonical expansion}
	  $\expand_0(B)$ of $B$ is the set of
	  $2^p$ congruent subboxes of $B$ that form
	  a subdivision of $B$;
	  for simplicity, we may assume this canonical expansion; but
	  see \cite{sss2} for other expansions.
	  In the SSS framework, we also have a procedure
	  (still called) $\expand(B)$, which acts as follows:
	  given a leaf $B$ of $\TTT(B_0)$, it converts $B$ into
	  an internal node whose children form the set $\expand(B)$.
	  The expand procedure will immediately classify
	  each $B'$ in the set $\expand(B)$
	  using a soft predicate $\wtC$, and perform
	  some additional actions as outlined below.
	Thus, we see that the tree $\TTT$ is initially just
	  the root $B_0$ and it grows by repeated expansion of its leaves.
	  The set of leaves of $\TTT$ at any moment constitute a subdivision
	  of $B_0$.
	  For our SE(3) subdivision tree, the root $B_0$ is initially
	  $B_0^t \times \wh{SO}(3)$.  Our expansion of any box
	  $B=B^t\times B^r$ takes a specific form: it is either
	  	$\expand(B^t)\times B^r$ or $B^t\times\expand(B^r)$.
		This is the T/R Split idea in
		\cite[Section 3]{luo-chiang-lien-yap:link:14}.
		However, when $B^r=\wh{SO}(3)$, this expansion is
		special: we always have $\expand(B^r)=\set{C_w, C_x, C_y, C_z}$
		as illustrated in \refFig{so3-model}.

	  The SSS Algorithm maintains the subdivision tree $\TTT$
	  using three WHILE-loops -- see box below.
	  The goal is to find a path from
	  the start $\alpha$ to the goal $\beta$ where $\alpha,\beta\in X$;
	  or report \nopath, satisfying the requirements of
	  $\veps$-exactness (see Definition 1, Section 1).
	  Let $Box_{\TTT}(\alpha)$ denote the leaf box of $\TTT$
	  that contains $\alpha$.
	  The first WHILE-loop keeps
	  expanding $Box_{\TTT}(\alpha)$ until it becomes \free, or
	  returns \nopath\ when $Box_{\TTT}(\alpha)$ has
			width
	  less than $\vareps$. 
	  The second WHILE-loop does the same for
	  $Box_{\TTT}(\beta)$.

	The last WHILE-loop (Main Loop) depends on
	three data structures, $Q, G, U$:
	  \ \\(a)
	  A priority queue $Q$ contains\footnote{
		From the procedure \expand($B$),
		each box $B'$ in the tree $\TTT$ has a classification
		$\wtC(B')\in \set{\mixed,\free,\stuck}$.)
	  }
	  only \mixed\ boxes.
	  \ \\(b)
	  An \dt{adjacency graph} $G$ whose nodes are the \free\ boxes in
	  $\TTT$, and whose edges connect pairs of adjacent boxes,
	  i.e., pairs that share a $(d-1)$-face.
	  \ \\(c)
	  A Union-Find data structure $U$ to represent
	  the connected components of $G$.
	  
		The above $\expand(B)$ procedure takes these
		additional actions: for each $B'$ in the set $\expand(B)$, 
	  if $w(B')<\veps$ or $\wtC(B')=\stuck$, we discard $B'$.
	  If $\wtC(B')=\free$, we insert $B'$ into the adjacency
	  graph $G$; we also insert $B'$ into $U$ and perform
	  $U.Union(B',B'')$ with boxes $B''$ in $G$ which are adjacent.
	  If $\wtC(B')=\mixed$, it is pushed into $Q$.

	  The main WHILE loop will keep expanding
	  $Q.\getnext()$ until a path is detected or $Q$ is empty.
	  If $Q$ is empty, it returns \nopath.  Paths are detected when
	  the Union-Find data structure tells us that
	  $Box_{\TTT}(\alpha)$ and $Box_{\TTT}(\beta)$ are in the same
	  connected component.
	  It is then easy to construct a path in $G$ to
	  connect $Box_{\TTT}(\alpha)$ and $Box_{\TTT}(\beta)$.
	  Thus we get:

	    \progb{
              \lline[-1]{\sc SSS Framework:}
              \lline[-1]\zhaoqiX{radius $\to$ width}
		\lline[0] \INPUt:
				start and goal configurations $\alpha,\beta\in X$,
		\lline[10]
				obstacle set $\Omega\ib Z$,
				resolution $\vareps>0$,
				initial box $B_0\in \intbox W$.
		\lline[0] \OUTPUt: Path from $\alpha$ to $\beta$ in
					$Y\cap \mu(B_0)$ or \nopath.
		  \lline[5] Initialize a subdivision tree $\TTT$ with root
		  	$B_0$.
	          \lline[5] Initialize data structures $Q, G$ and $U$.
	          \lline[5] While ($Box_{\TTT}(\alpha) \neq \free$)
		    \lline[10] If width of $Box_{\TTT}(\alpha)$ is $<
		    	\vareps$, Return(\nopath)
	            \lline[10] Else \expand($Box_{\TTT}(\alpha)$)
	          \lline[5] While ($Box_{\TTT}(\beta) \neq \free$)
		    \lline[10] If width of $Box_{\TTT}(\beta)$ is $<
		    	\vareps$, Return(\nopath)
	            \lline[10] Else \expand($Box_{\TTT}(\beta)$)
	          \lline[5] \commenT{MAIN LOOP:}
		  \lline[5] While ($U.Find(Box_{\TTT}(\alpha))\neq
		  	U.Find(Box_{\TTT}(\beta))$)
	            \lline[10] If $Q_{\TTT}$ is empty, Return(\nopath)
	            \lline[10] $B\ass Q_{\TTT}.\getnext()$
	            \lline[10] $\expand(B)$
		  \lline[5]
		  	Using $G$, construct a channel $P=(B_1,B_2\dd B_k)$
			comprising adjacent
		  \lline[10]
			free boxes.  Construct a \dt{canonical path}
			$\olP=(\alpha, b_1, c_1, b_2\dd c_{k-1}, b_k, \beta)$
		  \lline[10]
			where $b_i=m(B_i)$ and $c_i=m(B_i\cap B_{i+1})$.
	    }


	  The correctness of SSS does not depend on 
	  search strategy (i.e., priority) of $Q$.
	  However, choosing a good search strategy
	  can have a great impact on performance. 
	  Two strategies with the best results are Greedy Best First
	  and some kind of Voronoi strategy.

\ssect{Method of Features}
	In SSS framework, the obstacle set $\Omega\ib Z =\RR^3$ is a closed
	polyhedral set.  We may assume its boundary $\partial\Omega$ is
	bounded. Then $\partial\Omega$ is partitioned into a set of boundary
	{\bf features}: {\bf corners} (points), {\bf edges} (relatively open
	line segments), or {\bf walls} (relatively open triangles). 
	\zhaoqiX{According to appendix \ref{app:sep}, our
	features only include corners, 
	closed edges and closed walls.}
	Let
	$\Phi(\Omega)$ denote the set of features of $\Omega$. The (minimal)
	set of corners and edges are uniquely defined by
	$\Omega$, but walls
	depend on a triangulation of $\partial\Omega$.
	\zhaoqiX{
	If corners and edges are uniquely defined, then walls are also 
	uniquely defined because corners and edges are boundaries of walls.}

	Our approach to soft predicates is based on the ``method of
	features'' \cite{sss,wang-chiang-yap:motion-planning:15}.
	The \dt{exact feature set} of $B$ is
		\beql{exactfeat}
			\phi(B)\as\set{f\in\Phi(\Omega): f\cap \Fp(B)\ne \es}.
			\eeql
	This is too hard to compute, and we want an approximation
	$\wtphi(B)$ with the property $\phi(B)\ib \wtphi(B)$.
	This property holds if we define
	the \dt{approximate feature set} of box $B$ as
			\beql{approximate-feature-set}
				\wtphi(B)\as\set{f\in\Phi(\Omega):f\cap
					\wtFp(B)\neq\emptyset}
			\eeql
	where $\wtFp(B)$ is the approximate footprint defined
	in Section 3.1.  Moreover the property \refeq{inclusion}
	on approximate footprint $\wtFp(B)$, i.e.,
		\beq
			\Fp(B) \ib \wtFp(B) \ib \Fp(\sigma B) \nonumber
			\eeq
	ensures the resolution exactness of our algorithm.
	Hence,
		\beq
			\phi(B) \ib \wtphi(B) \ib \phi(\sigma B).
			\eeq

	The idea is to maintain $\wtphi(B)$ for each
	box $B$ in the subdivision.
	We softly classify $B$ as $\wtC(B)=\mixed$ as long as
	$\wtphi(B)$ is non-empty; otherwise, we can decide whether
	$\wtC(B)=C(B)$ is $\free$ or $\stuck$. 
	For computational efficiency, we want the approximate
	feature sets to have {\bf inheritance property}, i.e.,
			\beql{inheritance-property}
				\wt\phi(B)\ib\wtphi(parent(B)).
			\eeql
	This can be ensured by a trick in \cite{rod-ring}.
	
	\zhaoqiX{We have not introduced yet what $\wtFp$ is.}
\sectL[sep]{Appendix: Explicit Parameterized Collision Detection
	Computation}\label{app:sep}
	Our previous paper~\cite{rod-ring}
	already provided explicit algorithms for
	the collision detection between obstacle features
	and some simple $\Pi_1$-sets, as summarized in the lemma below.

	\bleml{Sep_Ele}(\cite{rod-ring})
		Let $A$ be a point, an edge or a ball,
		and let $f$ be a feature.
		There are explicit procedures to
		answer the parametric separation query
		``Is $\Sep(A,f)>s$?''.
	\eleml
        In this paper, we provide new algorithms so that $A$ can be a
		special $\Sigma_2$-set.
		(cf.~ Sec.~\ref{ssec:sep1} on parameterized collision detection.)

	We remark that when $A$ is a cone and $f$ is a line, there
	appears to be no known explicit solution.  We only found
	iterative procedures \cite{zheng-chew:dist-to-cone:09} or
	Lagrangian minimization formulations, which are not explicit.
        
\ssect{General Geometric Notations}
	We give some basic definitions and notations used in Euclidean
	geometry $\RR^m$.
	Recall from Sec.\ref{ssec:sep1} that a set $A\ib \RR^m$ is
	\dt{simple} if there is a unique 
	algebraic set $\olA$ such that $A\subseteq\olA$ and
	$\dim(A)=\dim(\olA)$.
	In this case, we call $\olA$ an \dt{algebraic span} of $A$.

	Let $\partial A$ denote the boundary of a set $A$.
	Given two points $u,v$, let $\seg(u,v)$ denote the segment connecting
	them, and 
	$|\seg(u,v)|$ denote its length.
	Let $\ball(v,r)$ denote the ball centered at $v$ with radius $r$. 
	If $v$ is the origin $\0$, we just write $\ball(r)$.
	Let $cone(v,o,r)$ denote the right cone with apex $v$ whose base
	is a disc centered at $o$ of radius $r>0$.
	So the boundary of $cone(v,o,r)$ can be decomposed into
	a circular disc $disc(o,r)$ and a 
	conic surface which we call a \dt{traffic cone}, 
	denoted by $\tc(v,o,r)$.
	Note that the algebraic span $\ol{\tc(v,o,r)}$ 
	is an algebraic double cone surface whose equation 
	is $x^2+y^2=r^2z^2$ up to a coordinate transformation.
	
\begin{figure}
	\centering
    \includegraphics[width=0.4\textwidth]{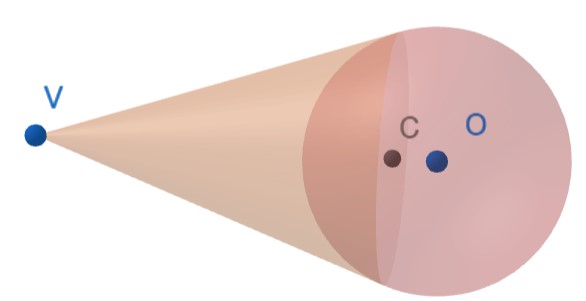} 
		\caption{An ice-cream cone $\ic(v,c,r)$ for some $r>0$.}
                \label{fig:ice-cream-cone}
	\end{figure}

	We also introduce an important shape for our analysis,
	the \dt{ice-cream cone} $\ic(v,o,r)$ which is defined
		as the union of $cone(v,o,r)$ and a ball $B$
		such that $B$ is tangential to the traffic cone $\tc(v,o,r)$.
		Note that the $disc(o,r)$ is a section of $B$ (see 
\switchLabel{Figure~\ref{fig:ice-cream-cone}).}{Figure~4).}
	The relation between ice-cream cone and its
	corresponding traffic cone is given by this lemma:
	
	\bleml{ic_tc}
	The ice-cream cone $\ic(v,o,r)$
	is the union of $cone(v,o,r)$ with the ball $\ball(c,R)$
	whose center $c$ on the axis of the cone satisfies
			$|\seg(v,c)|=\frac{h^2+r^2}{h}$
	where $h=|\seg(v,o)|$.
	The radius $R$ is given by $\frac{h^2+r^2}{r}$.
	\eleml
	
	Given any box $B\in\intbox W$, we can express it as
	$B=B^t\times B^r$,
	where $B^t\in\intbox\RR^3$ and $B^r\ib \wh{SO}(3)$.
	We now define $m(B), w(B)$ and $r(B)$ as follows:
	The \dt{center} of $B$, denoted $m_B=m(B)$,
	is the center of $B^t$.  Note that $m(B)$ is
	just the center of $B^t$, and independent of $B^r$.
	The \dt{width} of $B$, denoted by $w_B=w(B)$, is similarly just
	the width of $B^t$.
	The \dt{radius} of $B$, denoted by $r_B=r(B)$,
	is given by $r(B)=\sqrt{3}w_B/2$.
	Thus, the circumball of $B^t$ is simply $\ball(B^t)=\ball(m_B,r_B)$.
	
	\zhaoqiX{I think we can put the definitions and lemmas of ``closest
	pairs'' here.}
\ssect{Rearranging Approximate Footprint into Special $\Sigma_2$ Sets}
	In \refeqn{apprxFp}, we informally define 
	our approximate footprint $\wtFp(B)$ as
		$\ball(B^t)\oplus\wtFp(B^r)$
		where
		$$\wtFp(B^r) = \cup_{i=1}^6 P_i.$$
	Note that this means our $\wtFp(B)$ is a fat version
	of $\wtFp(B^r)$.
	In general, $A\oplus\ball(r)$ is a \dt{fat version} of $A$.

	Previously the sets $P_i$ were not fully specified there.
	We will see that each $P_i$ is, in fact,
	a special $\Pi_1$-set. Therefore, $\wtFp(B^r)$
	a special $\Sigma_2$-set.
	
\begin{figure}[htbp]
	\hspace{5pt}
	  \centering
	  \begin{minipage}[t]{0.3\textwidth}
	    \centering
	    \includegraphics[width=1\textwidth]{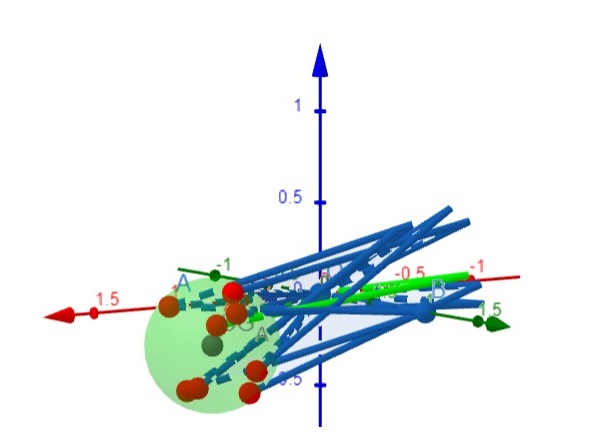} 
	    \caption{Getting $S_\pA$ from the $\pA_1\ldots\pA_8$.}\label{fig:SA}
	  \end{minipage}%
	  \hspace{0.03\textwidth}
	  \begin{minipage}[t]{0.3\textwidth}
	    \centering
	    \includegraphics[width=1\textwidth]{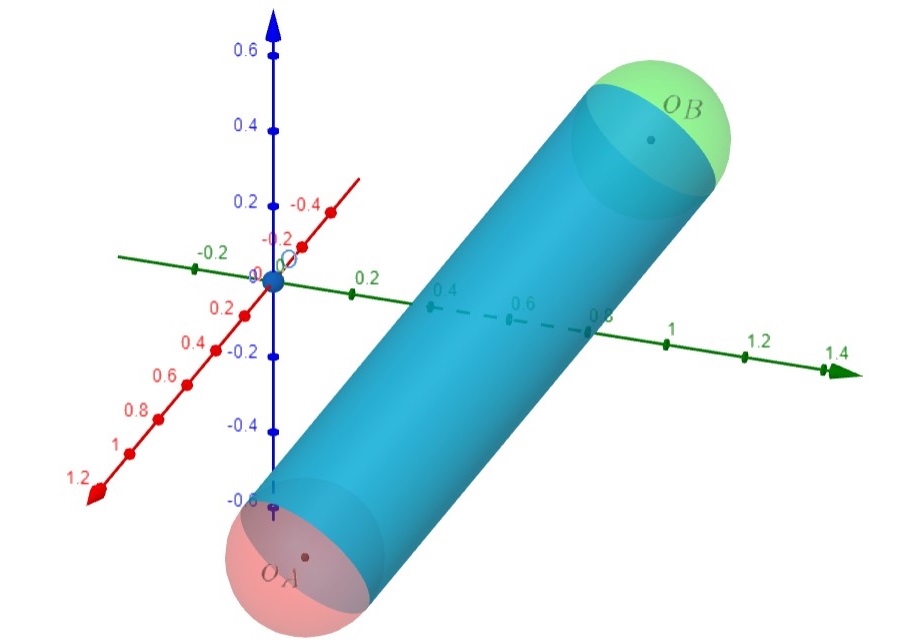}
	    \caption{The $Cyl$ makes the convex hull of $S_\pA$ and $S_\pB$}\label{fig:Cyl}
	  \end{minipage}%
	  \hspace{0.03\textwidth}
	  \begin{minipage}[t]{0.3\textwidth}
	    \centering
	    \includegraphics[width=1\textwidth]{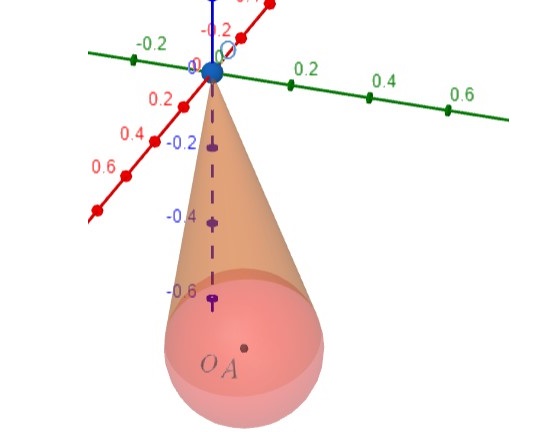}
	    \caption{The $Cone_\pA$ makes the convex hull of $\pO$ and $S_\pA$}\label{fig:ConeA}
	  \end{minipage}%
	\end{figure}

	As preliminary, consider
	the 8 corners $c_1\dd c_8$ of the box $B^r$.
		Each $c_i$ represents a rotation of the robot $\Delta\pA\pO\pB$.
		Let $\pA_i\in\RR^3$ denote the position of $\pA$ under this
		rotation.  Similarly for $\pB_i$.
		Then $o_\pA$ is defined as the center of gravity
		of these points, $o_\pA=\efrac{8}\sum_{i=1}^8 \pA_i$.
		Similarly for $o_\pB$. 
		Let $d(B)$ be the maximum distance from $o_\pA$ to any $\pA_i$
		and from $o_\pB$ to any $\pB_i$ ($i=1\dd 8$).
	See \refFig{SA}.

	\bitem
	\item
		$P_1=S_\pA$ is the $\ball(o_\pA,d(B))$.
	\item
		$P_2=S_\pB$ is the $\ball(o_\pB,d(B))$.
	\item
		$P_3=Cyl$ is the right-cylinder with $o_\pA$ and $o_\pB$ as the
		centers of its two base discs and $d(B)$ as its radius.
		Thus, $Cyl\cup S_\pA\cup S_\pB$ is just the convex
		hull of $S_\pA$ and $S_\pB$. See \refFig{Cyl}
	\item
		$P_4=Cone_\pA$ is right-cone with origin $\pO$ as its apex and a
		circular
		base that is tangent to $S_\pA$. Then
		$Cone_\pA \cup S_\pA$ is the convex hull of $\pO$ and
		$S_\pB$. See \refFig{ConeA}.
	\item
		$P_5=Cone_\pB$ is analogous to $Cone_\pA$.
	\item
		$P_6=Pry$ is a pyramid with apex at $\pO$ and a rectangular base
		that is tangential to $Cyl$.
		Thus the union $\cup_{i=1}^6 P_i$ is
		the convex hull of $\pO$, $S_\pA$ and $S_\pB$.
		See \refFig{w_box}.
		Note that each $P_i$ is a special $\Pi_1$-set.
	\eitem

	Now we apply $\wtFp(B)=\ball(B^t)\oplus\wtFp(B^r)$ to get the
	$\Sigma_2$ representation of $\wtFp(B)$. The $\wtFp(B)$ can be
	decomposed into the union of $7$ special $\Pi_1$ sets. They are the
	following:

	\begin{figure}[htbp]
	\hspace{5pt}
	  \centering
	  \begin{minipage}[t]{0.3\textwidth}
	    \centering
	    \includegraphics[width=1\textwidth]{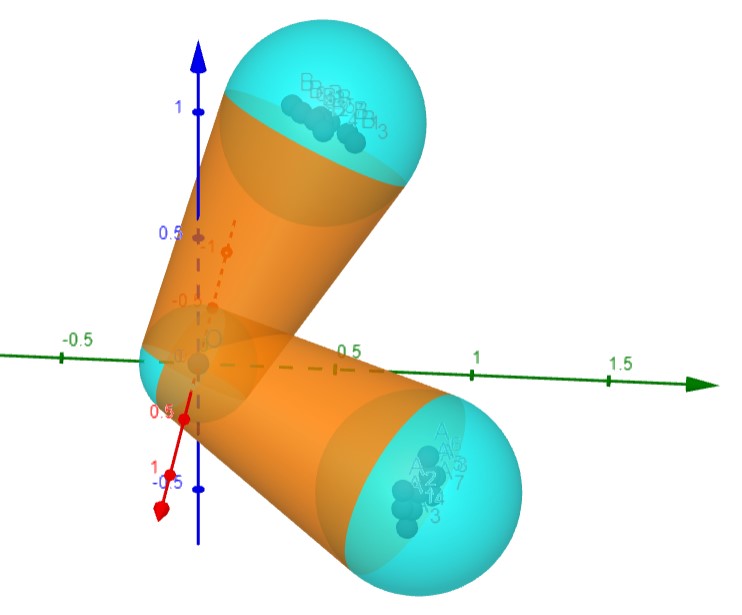}
		\caption{The frustums make the convex hull of $S_\pA$ and $S_\pO$
		  as well as the convex hull of $S_\pB$ and
		  $S_\pO$.}\label{fig:Frustum}
	  \end{minipage}%
	  \hspace{0.03\textwidth}
	  \begin{minipage}[t]{0.3\textwidth}
	    \centering
	    \includegraphics[width=1\textwidth]{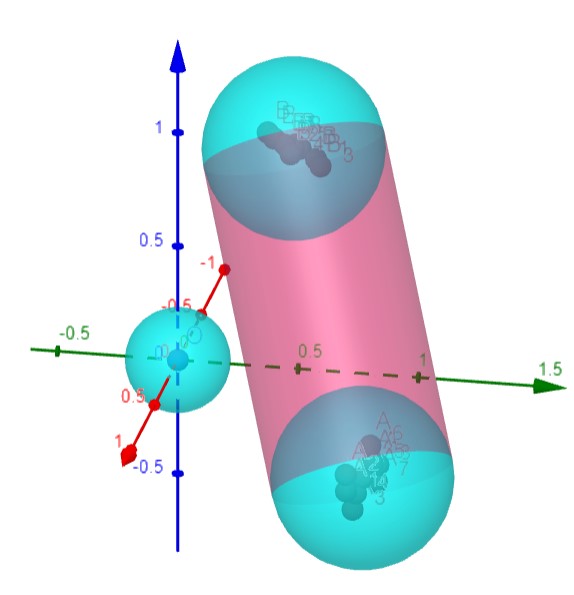} 
		\caption{The $Cyl$ makes the convex hull of $S_\pA^+$ and
		  $S_\pB^+$.}\label{fig:Cyl+}
	  \end{minipage}%
	  \hspace{0.03\textwidth}
	  \begin{minipage}[t]{0.3\textwidth}
	    \centering
	    \includegraphics[width=1\textwidth]{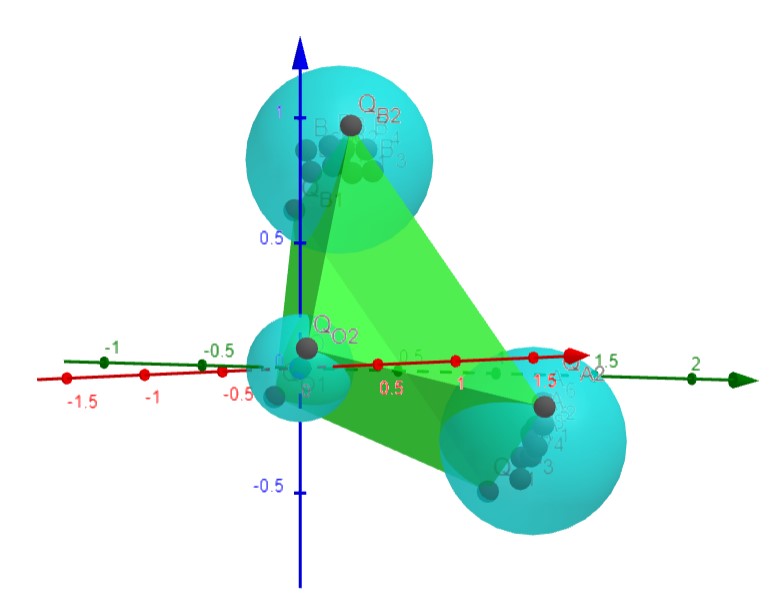}
		\caption{The $Pyr^+$ makes the convex hull of $S_\pA$, $S_\pB$
		  and $S_\pO$}\label{fig:Pyr+}
	  \end{minipage}%
	\end{figure}

	\bitem
	\item
	$S_\pO=\ball(B^t)$ (small cyan ball center at $\pO$);
	\item
	$S_\pA^+=S_\pA\oplus\ball(B^t)=\ball(o_\pA+m_B,r(B)+d(B))$ (large
	cyan ball center at $\pA$);
	\item
	$S_\pB^+=S_\pB\oplus\ball(B^t)=\ball(o_\pB+m_B,r(B)+d(B))$ (large
	cyan ball center at $\pB$);
	\item
	$Frust_A$ is the right-frustum whose union with $S_\pO$ and $S_\pA^+$
	will result in the convex hull of $S_\pO$ and $S_\pA^+$ (orange
	frustum);
	\item
	$Frust_B$ is the right-frustum whose union with $S_\pO$ and $S_\pB^+$
	will result in the convex hull of $S_\pO$ and $S_\pB^+$ (orange
	frustum), see \refFig{Frustum};
	\item
	$Cyl^+$ is the right-cylinder whose union with $S_\pA^+$ and
	$S_\pB^+$ will result in the convex hull of $S_\pA^+$ and $S_\pB^+$
	(pink cylinder), see \refFig{Cyl+};
	\item
		$Pyr^+$ is the polyhedron
		whose union with $S_\pA^+$, $S_\pB^+$, $S_\pO$, $Frust_A$,
		$Frust_B$ and $Cyl^+$ will result in the convex hull of
		$S_\pA^+$, $S_\pB^+$ and $S_\pO$
		(green polyhedron), see \refFig{Pyr+}.
	\eitem
\ssect{Exploiting a very special $\Sigma_2$-decomposition of $\wtFp(B)$}
	To implement collision detection of
	the approximate footprint $\wtFp(B)$ with a feature
	using the above special $\Sigma_2$-decomposition of $\wtFp(B)$,
	our boundary reduction technique requires us to compute
	the separation of a disc and a feature.  This requires computing
	roots of a degree 4 polynomial (see
%
          Appendix~\ref{ssec:Disc}
	and \cite[Appendix D.1]{rod-ring:arXiv}).
	The following lemma shows that we can reduce this computation to
	parametric separation query
	that amounts to checking a polynomial inequality of degree
	$2$ and whose coefficients are algebraic of degree $2$ (i.e.,
	square-roots).  
	
	\bleml{sigma2}
	If $B\in \intbox W$ has corners whose coordinates are
	rational numbers, then $\wtFp(B)$ is a $\Sigma_2$-set whose 
	defining polynomials have coefficients of degree at most $2$.
	\eleml
	\bpf
	If $q=(1,x,y,z)\in \wh{SO}(3)$ has rational coordinates,
	$q/\|q\|\in SO(3)$, then the corresponding $3\times 3$ orthogonal
	matrix $M_q$ has rational entries.
	As a result, the ball $S_\pA$ has rational center and radius
	is a square root of a rational number. 
	So the equation defining $S_\pA$ is a polynomial of degree $2$
	with rational coefficients. Moreover, $r(B)=\frac{\sqrt{3}}{2}w(B)$ 
	is a square root of rational number. It implies that $S_\pA^+=S_\pA\oplus\ball(B^t)$ 
	is a polynomial whose coefficients have
	algebraic degree $2$. The other cases are similar.
	\epf
	
	To exploit the above lemma, we need a new ``very special''
	$\Sigma_2$-decomposition of our approximate footprint,
	namely as the union of special $\Sigma_2$-sets (such as
	ice-cream cones and fat line segments).  Note that ice-cream cones
	and fat line segments avoid computations involving discs.
	
	\bleml{CD}
	Given box $B\in\intbox W$ and feature $f$, $\wtFp(B)\cap f=\emptyset$
	if and only if
		\begin{align*}
		(\Sep(\ic_\pA,f)>r(B)) & \land(\Sep(\ic_\pB,f)>r(B))\\
		& \land(\Sep(\seg(o_\pA,o_\pB),f)>d(B)+r(B))\\
		& \land(\Sep(Pyr^+,f)>0),
		\end{align*}
	where $\ic_\pA=\ic(m_B,m_B+o_\pA,d(B))$ and
	$\ic_\pB=\ic(m_B,m_B+o_\pB,d(B))$ are the ice-cream cones given by
	$Cone_\pA\cup S_\pA$ and $Cone_\pB\cup S_\pB$ with a translation of
	$m_B$.
	\eleml
	
	\bpf
	We rearrange our approximate footprint in this way:
	\[\wtFp(B)=\left(S_\pA^+\cup Frust_\pA\cup
	S_\pO\right)\cup\left(S_\pB^+\cup Frust_\pB\cup
	S_\pO\right)\cup\left(S_\pA^+\cup Cyl^+\cup S_\pB^+\right)\cup Pyr^+.\]
	We notice the following relations:
	\begin{align*}
	S_\pA^+\cup Frust_\pA\cup S_\pO & =\left(Cone_\pA\cup
		S_\pA\right)\oplus \ball(B^t)=\ic_\pA\oplus\ball(r(B))\\
	S_\pB^+\cup Frust_\pB\cup S_\pO & =\left(Cone_\pB\cup
		S_\pB\right)\oplus \ball(B^t)=\ic_\pB\oplus\ball(r(B))\\
	S_\pA^+\cup Cyl^+\cup S_\pB^+ &
		=\seg(o_\pA,o_\pB)\oplus\ball\left(d(B)+r(B)\right)
	\end{align*}
	Then by \refLem{Sep_Minkow}, $\Sep(\wtFp(B),f)>0$ if and only if the
	four terms in the conjunction form described in the lemma are all true.
	\epf
		
	\zhaoqiX{Disc lemma is ignored here.}
	\cheeX{I brought this back (at least for now, but I think
		it should be a separate section)}
	\ignore{
	\ssect{Disc}
	The collision detection for disc primitive is discussed by different features:
	
	For point feature,
	\Ldent\progb{
		\lline[0] Input: a disc $D$ and a point $v$, real number $r>0$
		\lline[0] Output: boolean ($\Sep(D,v)>r$)
		\lline[5] Find projection of $v$ on the plane of $D$, denote by $w$
		\lline[5] If ($w\in D$) return ($d(v,w)>r$)
		\lline[5] else return ($\Sep(\partial D,v)>r$)
	}
	
	For segment feature,
	\Ldent\progb{
		\lline[0] Input: a disc $D$ and a segment $g$, real number $r>0$
		\lline[0] Output: boolean ($\Sep(D,g)>r$)
		\lline[5] Let $w\as \olD\cap \ol{\ell}$,
		\lline[5] If ($w\in D\cap g$) return \false
		\lline[5] else if ($w\in D$) return ($\Sep(D,\partial g)>r$)
		\lline[5] else return ($\Sep(\partial D,g)>r$)
	}
	
	For triangle feature,
	\Ldent\progb{
		\lline[0] Input: a disc $D$ and a triangle $t$, real number $r>0$
		\lline[0] Output: boolean ($\Sep(D,t)>r$)
		\lline[5] If ($\Sep(D,\partial t)\leq r$) return \false
		\lline[5] else if ($\Sep(\partial D,t)\leq r$) return \false
		\lline[5] else return \true
	}
	}
\ssectL[Disc]{Boundary Reduction Method for Disc-Edge}
	We can apply the boundary reduction method to 
	produce a parametric separation query for discs.
	Note that discs are $\Pi_1$-sets.

	In this section, we give an example of boundary reduction method. We consider a disc $\Pi_1$ set defined by $D=\{(x,y,z)\in\RR^3|x^2+y^2\leq1,z=0\}$, and an edge feature defined by $f=\{(a_0,b_0,c_0)+t(a,b,c)|t\in[0,1]\}$. We compute the parametric collision detection ``if $\Sep(D,f)>s$'' for $s>0$.
	
	The collision detection is based on the following process:
	\Ldent\progb{
		\lline[0] Input: a disc $D$ and a segment $f$, real number $s>0$
		\lline[0] Output: boolean ($\Sep(D,f)>s$)
		\lline[5] If ($\Sep(D,\partial f)\leq s$), return \false
		\lline[5] For each $(\bfp,\bfq)\in\cp(\partial D,\olf)$,
		\lline[10] if ($d(\bfp,\bfq)\leq s$ and $\bfq\in f$),
		\lline[15] return \false
		\lline[5] For each $\bfw\in\olD\cap\olf$ (this is unique)
		\lline[10] if ($\bfw\in D$ and $\bfw\in f$)
		\lline[15] return \false
		\lline[5] return \true
	}
	
	Among the process above, we recursively call the query
	``$\Sep(D,\partial f)>s$?'', where $\partial f$ is a set of two
	points. This subquery is standard and thoroughly studied by many
	people. We focus on the computation of $\cp(\partial D,\olf)$, which
	is a good example of solving the polynomial equations listed in
	\refeq{system}.
	
	To begin with, suppose $(\bfp,\bfq)\in\cp(\partial D,\olf)$, where 
	$\partial D=\{x^2+y^2=1\}\cap\{z=0\}$ and 
	$\olf=\{(a_0,b_0,c_0)+t(a,b,c)|t\in\RR\}$. 
	We denote the normal direction of $\partial D$ to be $\bfd=(0,0,1)$ 
	and the direction of $\olf$ to be $\bff=(a,b,c)$. Given any 
	$\bfq=(x_q,y_q,z_q)\in\olf$, let's compute
	$(\bfp,\bfq)\in\cp(\partial D,\bfq)$. 
	Suppose that the solution is $\bfp=(x,y,z)$. Corresponding to symbols in 
	\refeq{system}, $f_1(x,y,z)=x^2+y^2-1$ and $f_2(x,y,z)=z$. 
	Let $\bfv=\nabla f_1(\bfp)\times \nabla f_2(\bfp)$. Noticing that
	$\bfp\perp\bfv$ 
	(since $\bfp$ is on a circle, which is equivalent to $f_1(\bfp)=0$), and 
	$(\bfq-\bfp)\perp\bfv$, $\bfp$ is the intersection between the plane
		constructed by 
	$\bfq$ and $\bfd$ with $\partial D$. This gives us
		\[\bfp=(x,y,z)=\left(\frac{x_q}{\sqrt{x_q^2+y_q^2}},
				\frac{y_q}{\sqrt{x_q^2+y_q^2}},0\right).\]
	By applying $\bang{\bfq-\bfp,\bff}=0$, we have the equation:
	\[ax_q\left(1-\sqrt{x_q^2+y_q^2}\right)
		+by_q\left(1-\sqrt{x_q^2+y_q^2}\right)
		+cz_q\sqrt{x_q^2+y_q^2}=0.\]
	This is equivalent to
			\[ax_q+by_q=(ax_q+by_q-cz_q)\sqrt{x_q^2+y_q^2},\]
	or
		\[\left(ax_q+by_q\right)^2
			=\left(ax_q+by_q-cz_q\right)^2\left(x_q^2+y_q^2\right).\]
	Here, $x_q,y_q,z_q$ are linear to the variable $t$.
	Then the equation above is a 
	polynomial of $t$ with degree $4$. 
	{\em Hence, solving the equation is the same with 
	solving a quartic equation.}
	The root of the polynomial of $t$ gives possible 
	$\bfq\in\olf$. Then if there is $\bfq\in f$ such that $d(\bfp,\bfq)\leq s$, 
	the query returns false.
\ssect{Parametric Query for Special $\Sigma_2$-sets}
	\cheeX{Zhaoqi, can you please at least sketch the argument?}
	\bthmT{Theorem 4 in main paper}{special2} \ \\
               There are explicit methods
	       for parametric separation queries of the form
			``Is $\Sep(P,f)>s$?''
	       where $P$ is a special $\Pi_1$-set and $f$ is a feature.
	\ethmT
	\bpf
		Parametric separation queries
		for convex polyhedra is standard.
		Hence, it remains to consider 
		parametric separation queries
		for other special
		$\Pi_1$-sets, viz., right cylinder, right cone and right frustum.
		
		The main idea is based on the boundary
		reduction technique.  The technique reduces
		the query to the following 3 subqueries $Q_0>s$, $Q_A>s$
		and $Q_f>s$.  We note that $Q_0>s$ is trivial:
		Given a special $\Pi_1$ set $\Pi$, its algebraic span
		is $\RR^3$. Hence the closest pair $\cp(\ol\Pi,f)$ for any
		feature $f$ takes place at the whole feature. 
	 Then $\cp(\Pi,f)\neq\es$
	 	if and only if $\Pi^\circ\cap f\neq\es$. 
		 Since $f$ is a closed set, the relation between $\Pi$ and $f$
		 can only be 
		 classified into three cases: $\partial\Pi\cap f\neq\emptyset$, 
		 $f\subset\Pi^\circ$ or $f\cap\Pi=\emptyset$. Checking the first
		 case is solving 
		 the algebraic equations which is standard. To check the second
		 case, we can 
		 pick any $\bfq\in f$ and check if $\bfq\in\Pi$ or not 
		 by checking each of algebraic inequalities that form the $\Pi_1$
		 set.  So we suppose that we are in the third 
		 case, i.e. $f\cap\Pi=\emptyset$ where  $\cp(\Pi,f)=\emptyset$. 

		It remains to consider the subqueries $Q_A>s$ and  $Q_f>s$.
		Checking if $Q_A>s$, i.e., $\Sep(\Pi,\partial f)>s$
		is solved recursively.
		Hence we only focus on deciding 
		$Q_f>s$, i.e., $\Sep(\partial \Pi,f)>s$. 
		We observe that for our special $\Pi_1$-sets,
		$\partial\Pi$ can be decomposed into two or three surfaces
		of the form:
		(i) the intersection of
		a quadric surface and a slab (which is bounded by two parallel
		planes).  
		(ii) a disc.
		The case of a disc has been addressed in
		Appendix~\ref{ssec:Disc}.
		The quadric surface has two possibilities:
		one is the non-planar surface of a right frustum,
		which is treated in Appendix~\ref{ssec:Cone}. 
		This includes the traffic cone as a special case.
		The other possibility is the non-planar surface
		of a right cylinder, denoted by $\Cyl$. 
		Noticing that if 
			$(\bfp,\bfq)\in\cp(\Cyl,f)$,
			then $\bfq$ is a projection of the axis of $\Cyl$ 
		onto $f$. Finding the projection will give us a
		potential $\bfq$ and we can find a corresponding $\bfp$.
		This finishes our proof.
	\epf
\ssectL[Ice]{Ice-cream Cone: Reducing the maximum degree from $4$ to $2$}
\ignore{
	In Subsection~\ref{ssec:sigma2}, we reduced 
	the predicate ``Is $\wtFp(B)\cap f$ empty?'' to a double-loop computation 
	based on the decomposition $\wtFp(B)=\cup_{i=1}^m \bigcap_{i=1}^n A_{ij}$ 
	where each $A_i=\bigcap_{i=1}^n A_{ij}$ is a simple $\Pi_1$-set. 
	Among these $A_i$'s are the right-cones.  The boundary of the right-cone 
	is the union of a traffic cone and circular disc $D$.  Computing 
	the separation between disc $D$ and a line amounts to solving 
	a degree 4 (see \cite[Appendix D.1]{rod-ring:arXiv}).  Its correct 
	implementation (unless we use exact computation) is numerically 
	rather unstable.  But we can avoid solving 
	quartic equations as follows: 
	$$\wtFp(B)=C_1 \cup C_2 \cup C_3 \cup C_4$$ 
	where $C_1$ and $C_2$ are fat ice-cream code, $C_3$ 
	is a fat line segment, and $C_4$ is a convex polyhedron. 
	So checking ``is $\wtFp(B)\cap f$ empty?'' is reduced to checking if
	each $C_i\cap f$ is empty as \refLem{CD} shows.  These is easy to do
	for $i=3$ and $i=4$ (see \refLem{Sep_Ele}), so we focus on the case
	where $C_i$ is a fat ice-cream cone. 
	}

	We solve the parametric separation query for the ice-cream cone,
	namely
		$\Sep(\ic(v,o,r),f)>0$
	where $f$ is a feature.

	\zhaoqiX{ THIS NEEDS TO BE FIXED to new definition of \ic.}
	
	\Ldent\progb{
		\lline[0] Input: an ice-cream cone $\ic(v,o,r)$, 
		a feature $f$ and a real number $s>0$.
		\lline[0] Output: boolean ($\Sep(\ic(v,o,r),f)>s$).
		\lline[0] Method:
		\lline[5] If $\Sep(v,f)\leq s$, return {\tt false}
		\lline[5] If $\Sep(\ball(o,r),f)\leq s$, return {\tt false}
		\lline[5] Let $h \as |\seg(v,o)|$, $k \as \sqrt{h^2-r^2}$,
			a point $c \as o+\frac{r^2}{h^2}(v-o)$
		\lline[5] If $\Sep(\tc(v,c,\frac{rk}{h}),f)\leq s$, return {\tt false}
		\lline[5] return {\tt true}
	}
	
	The check for $\Sep(v,f)\leq s$ and $\Sep(\ball(o,r),f)\leq s$ are in
	\refLem{Sep_Ele}. For the problem $\Sep(\tc(v,c,\frac{rk}{h}),f)\leq
	s$, we solve it by a reduction from each feature to its boundary
	features, according to the \refLem{reduction}.
	The case $f$ is a corner is easy, so we focus on the
	other two cases.
	\ignore{%
	Case $f$ is a point:
	\Ldent\progb{
		\lline[0] Input: a traffic cone $\tc(v,c,r)$ and a point $u$, a
			real number $s>0$.
		\lline[0] Output: boolean ($\Sep(\tc(v,c,r),u)>s$).
		\lline[0] Method:
		\lline[5] Let $p$ be the plane containing both $\seg(v,c)$ and $u$, 
		\lline[5] then $\we\as p\cap\tc(v,c,r)$ is the set of two edges
			of a triangle
		\lline[5] return ($\Sep(\we,u)>s$)
	}
	}%
	
	Case $f$ is an edge (line segment):
	\Ldent\progb{
		\lline[0] Input: a traffic cone $\tc(v,c,r)$ and a line segment
		$g$, a real number $s>0$.
		\lline[0] Output: boolean ($\Sep(\tc(v,c,r),g)>s$).
		\lline[0] Method:
		\lline[5] If $\Sep(\tc(v,c,r),\partial g)\leq s$, return \false
		\lline[5] If $\cp(\ol{\tc(v,c,r)},\ol{\ell})=\emptyset$ (closest pair
		in Sec.~\ref{ssec:sep1}), return \true
		\lline[5] For each pair of $(u,w)\in\cp(\ol{\tc(v,c,r)},\ol{\ell})$,
		\lline[10] if $d(u,w)\leq s$
		\lline[15] if $u\in\tc(v,c,r)$ and $w\in g$,
		\lline[20] return \false
		\lline[5] return \true
	}
	
	Case $f$ is a triangle:
	\Ldent\progb{
		\lline[0] Input: a traffic cone $\tc(v,c,r)$ and a triangle $T$,
			a real number $s>0$.
		\lline[0] Output: boolean ($\Sep(\tc(v,c,r),T)>s$).
		\lline[0] Method:
		\lline[5] If $\Sep(\tc(v,c,r),\partial T)\leq s$, return \true
		\lline[5] return ($\seg(v,c)\cap T\neq\emptyset$)
	}
	
\begin{figure}[htbp]
	\hspace{5pt}
	  \centering
	  \begin{minipage}[t]{0.25\textwidth}
	    \centering
	    \includegraphics[width=1\textwidth]{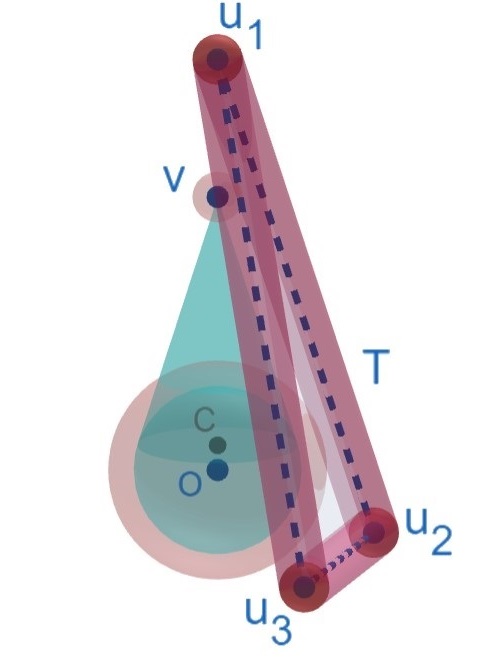} 
	    \caption*{(a)}
	  \end{minipage}%
	  \begin{minipage}[t]{0.5\textwidth}
	    \centering
	    \includegraphics[width=0.8\textwidth]{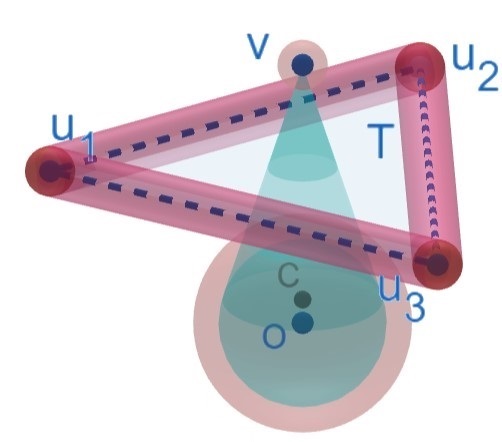}
	    \caption*{(b)}
	  \end{minipage}%

	\caption{{\small  Separation between triangle and ice-cream cone
		compared to the separation between their boundaries:\\
	(a) When the triangle does not intersect across the ice-cream cone,
		the boundary will always reach the minimum separation since the
		traffic cone is a developable surface.\\
	(b) If the triangle intersects across the ice-cream cone, it will
		intersect with the axis of the cone.}}
	\label{fig:Ice_Cream_Triangle}
	\end{figure}

	The return $\seg(v,c)\cap T\neq\emptyset$ in the last case may not be so
	obvious, but we can see it from the \refFig{Ice_Cream_Triangle}.
	
\ssectL[Cone]{Closest Pairs between Cone and Line}
	\zhaoqiX{I will rewrite this part according to the new descriptions
	of the ``closest pairs''.}

	In this section, let $\TC=\tc((0,0,0),(0,0,-1),r)$, i.e., the
	algebraic representation 
	for $\ol{\TC}$ is $x^2+y^2-r^2z^2=0$. We illustrate how to find
	$\cp(\ol{\TC},\ol{\ell})$ for traffic cone $\TC$ and line segment $\ell$.
	We assume that $\ol{\ell}$ is represented parametrically
	by $\bfq(t)=\bfv_0+\bfv t$, 
	for $t\in\RR$, where $\bfv=(a,b,c)$. Given a point $\bfp\in\ol{\TC}$, 
	the normal vector at that point is $\bfn_\bfp$.
		
	Before applying \refeq{system} to find local minima, we first check 
	if $\ol{\TC}\cap\ol{\ell}$ is non-empty. This is solving the two
	equations. If there are solutions, we regard them as additional
	closest pairs in $\cp(\ol{\TC},\ol{\ell})$ 
	in addition to the solutions of \refeq{system}. 
	
	Now we apply \refeq{system} to our problem. For our problem, 
	we only have $f_1(x,y,z)=x^2+y^2-r^2z^2$. Our formulation
	of the system \refeq{system} does not involve the polynomials
	$g_1$ or $g_2$, instead we have
		$\nabla g_1(\bfq)\times \nabla g_2(\bfq)$ given by
		the vector $\bfv$. 
	We do not have $\nabla f_1(\bfp)\times \nabla f_2(\bfp)$,
	but we have the $\bfn_\bfp$ which is equal to
	$\bfp-\bfq$ up to a constant multiple.
	The perpendicular conditions can be reduced to 
	$\bang{\bfn_\bfp,\bfv}=0$. This yields two equations to be solved
	for $\cp(\ol{\TC},\ol{\ell})$, where $(x,y,z)=\bfp$:
	
	\ignore{
		\bleml{cone_line}
		A point $w\in \ol{\ell}$ that makes $\Sep(w,\ol{\TC})$ a local minimum,
		\textbf{only if} the point $u\in\ol{\TC}$ such that
		$d(w,u)=\textrm{Sep}(w,\ol{\TC})$ satisfies $\bfn_u\perp\bfv$. The
		number of points that may satisfy this property is no more than $2$ as presented in \refFig{lm}.
		\eleml
		
		\bpf
		We assume that $\ol{\ell}=\{w(t)\in\RR^3|w(t)=v_0+\bfv t, t\in\RR\}$ 
		where $v_0\in\ol{\ell}$ and $\bfv$ is the direction of $\ol{\ell}$. 
		For each point $w(t)\in\ol{\ell}$, let's consider the point 
		$u(t)\in\ol{\TC}$ such that $d(w(t),u(t))=\Sep(w(t),\ol{\TC}$. 
		This point has be appeared in the plane $p$ containing 
		the axis of $\ol{\TC}$ and $w(t)$. The intersection between $p$ 
		with $\ol{\TC}$ are two intersecting lines, which we denote them by 
		$h_1$ and $h_2$ respectively. Then 
		$\Sep(w(t),\ol{\TC})=\min\{\Sep(w(t),h_1),\Sep(w(t),h_2)\}$. Each of the two
		$\Sep(w(t),h_i)$ is a differentiable function (function of $t$), and by 
		the Lagrangian's multiplier, the value reaches a
		local minimum or maximum when $\seg(u(t),w(t))\perp\bfv$.
		Note that, also by the Lagrangian's multiplier, $\bfn_u$ is the 
		direction of $\seg(u(t),w(t))$, and hence $\bfn_u\perp\bfv$. We
		denote the possible
		solutions to be $w_j$ for $j\in\NN$. For each $w_j$, the value of
		$\Sep(w_j,h_i)$ 
		has to be a local minimum of $\Sep(w(t),h_i)$, 
		since when $t\to\infty$ we have $\Sep(w(t),h_i)=+\infty$. Then the
		function $\min\{\Sep(w(t),h_1),\Sep(w(t),h_2)\}$ has at most two
		local minimums,
		which are $w_1$ and $w_2$ respectively corresponding to the local
		minimums of $\Sep(w(t),h_1)$ and $\Sep(w(t),h_2)$.
		\epf
		\begin{figure}
		\centering
	  	\begin{minipage}[t]{0.45\textwidth}
	  	\centering
		\includegraphics[width=0.75\textwidth]{figs/local_minimum.png}
		\caption{There are at most $2$ possible local minimums.}
		\label{fig:lm}
		\end{minipage}
		\hspace{0.06\textwidth}
	  	\begin{minipage}[t]{0.45\textwidth}
	  	\centering
		\includegraphics[width=1\textwidth]{figs/special_case.png}
		\caption{
		This is a case when the two $u_i$ are on the same branch, 
		while only one of the $u_i$ correspond to $w_i$ gives for
			$\Sep(w_i,\ol{\TC})$.}
		\end{minipage}
		%
		\end{figure}
		
		According to \refLem{cone_line}, when we have a point $u\in\ol{\TC}$
		whose normal vector is perpendicular to $\ol{\ell}$, it satisfies the equation
		\[0=\bfn_u\cdot\bfv=ax+by-r^2cz.\]
		
		For a simplification, let's define $\kappa=rc$. To find out possible
		$u\in\ol{\TC}$ 
		such that it is an example of $\cp(\ol{\TC},\ol{\ell})$,
		we solve the equations:
	}

		\begin{align}
			x^2+y^2=r^2z^2 \label{eq:cone_line_quad}\\
			ax+by=r^2cz \label{eq:cone_line_linear}
		\end{align}

	Let $\kappa \as rc$.  Multiplying \refeq{cone_line_quad} by
	$\kappa^2$
	and subtracting square of \refeq{cone_line_linear}, we get
	a quadratic equation for $x$ and $y$:
		\[(\kappa^2-a^2)x^2+(\kappa^2-b^2)y^2-2abxy=0,\]
	which is equivalently:
		\[\left(\begin{array}{cc}x & y\end{array}\right)
			\left(\begin{array}{cc}\kappa^2-a^2 & -ab\\ -ab &
					\kappa^2-b^2\end{array}\right)
			\left(\begin{array}{c}x \\
				y\end{array}\right)=0.\]
	Let
		\[A=\left(\begin{array}{cc}\kappa^2-a^2 & -ab\\
			-ab & \kappa^2-b^2\end{array}\right).\]
	Then
		\[\det(A)=(\kappa^2-a^2)(\kappa^2-b^2)-a^2b^2
				=\kappa^2(\kappa^2-a^2-b^2).\]
	
	When $\det(A)>0$, the equation has no real non-trivial ($x=y=0$ is
	trivial) solution. Hence $\textrm{cp}(\ol{\TC},\ol{\ell})=\emptyset$. In
	this case, $\ol{\TC}\cap\ol{\ell}\neq\emptyset$, which we have excluded in
	previous discussions.
	
	When $\det(A)\leq0$, the equation has real non-trivial solutions 
	$(\kappa^2-a^2)x+(ab\pm \sqrt{-\det(A)})y=0$, which are one or two planes. 
	We denote the planes by $P_1$ and $P_2$.
	
	To find the exact $\bfp$ and $\bfq$, one only need to notice that 
	$\bfq_1=P_1\cap\ol{\ell}$ and
		$\bfq_2=P_2\cap\ol{\ell}$ ($P_i\cap\ol{\ell}\neq\emptyset$ since 
	$\bfp\perp\ol{\ell}$ and $P_i$ contains the direction of $\bfp$).
	Moreover, let 
	$\we_i=P_i\cap\ol{\TC}$ which is a cross of two lines by the apex of $\ol{\TC}$, 
	there are $\bfp_{i1},\bfp_{i2}\in\we_i$ such that $(\bfp_{ij}-\bfq_i)\perp\we_i$ 
	for $i=1,2$ and $j=1,2$. This gives four pairs of $(\bfp_{ij},\bfq_i)$. Noticing 
	that some of the pairs gives a local maximum for $d(\bfp_{ij},\bfq_i)$, we should 
	get rid of those pairs, which may result in at most two pairs. The remaining possible 
	pairs give $\cp(\ol{\TC},\ol{\ell})$.
\sectL[adj]{Appendix: Subdivision for $\wh{SE}(3)$}


\ignore{
\newcommand{\SIGNS}{\ensuremath{\set{\pm 1,0}}}
\newcommand{\DIR}{\ensuremath{\SIGNS^7}}

A vector $\bfd=(b_1\dd b_d)\in\SIGNS^d$ is called
a \dt{direction} if $|\bfd|=1$
where $|\bfd|\as \sum_{i=1}^d |\beta_i|$.
Then $\DIR$ denote the set of directions in $\SIGNS^$,
called the set of \dt{directions} in $SE(3)$.
If $\bfd=(b_1\dd b_7)\in \DIR$,
let $\bfd^t\as(b_1,b_2,b_3)$ and $\bfd^r\as(b_4,b_5,b_6,b_7)$,
where $\bfd^t$ or $\bfd^r$ is called a direction
of $X^t$ or $X^r$.

}

\ssect{Box Adjacency Calculus} \label{sec:box-adj}
	We first introduce a notation for discussing
	the $j$-th component of a box $B=\prod_{i=1}^n I_i$
	where each $I_i$ is an interval.
	For $j=1\dd n$, let
		$\Proj_j(B) \as \prod_{i=1, i\ne j}^n I_i$
	denote the $n-1$ dimensional box obtained by omitting
	the $j$-th component.
	Then define the operator $\otimes_j$ using this identity:
	if $B=\prod_{i=1}^n I_i$, then
		$$B = \Proj_j(B) \otimes_j I_j$$
	For $j\ne k$, we extend the notation to $\otimes_{j,k}$
	using the identity
		$$B = \Proj_{j,k}(B) \otimes_{j,k} (I_j\times I_k)$$
	If $I=[a,b]$ and $I'=[a',b']$ are intervals, we write
			$I\dder{+1}I'$ if $b=a'$,
		and $I\dder{-1}I'$ if $a=b'$.
	Let $e_j$ be the elementary $j$-th vector in $\RR^n$
	(so $e_j$ is a $n$-vector of all $0$'s except for a $1$
	in the $j$-th position).
	We call $d\in \set{\pm e_1\dd \pm e_n}$ to be
	a \dt{semi-direction}.
	For $n\ge 2$, we say that $B$ is \dt{adjacent to $B'$ in the
	direction $\pm e_j$}, denoted
		$B\dder{\pm e_j} B'$,
	if 
		$B=\Proj_j(B)\otimes_j I_j$
		and
		$B'=\Proj_j(B')\otimes_j I'_j$
		and $I_j\dder{\pm 1} I'_j$ and
		($\Proj_j(B)\ib\Proj_j(B')$
			or
		$\Proj_j(B)\ip\Proj_j(B')$).

	\dt{The case of $\whSO$:}
	We extend the notion of adjacencies
	to the 3-dimensional boxes embedded in $\RR^4$.
	The boundary of $[-1,1]^4\ib\RR^4$ is subdivided into
	eight 3-dimensions boxes denoted $\pm C_i$ for $i=0\dd 3$, 
	where\footnote{
		We now index our components from
		$0\dd 3$ instead of $1\dd 4$.   Moreover,
		$(0,1,2,3)$ is also written $(w,x,y,z)$,
		as in $C_0=C_w$, etc.
	}
		$$C_i = [-1,1]^3 \otimes_i \oone$$
	where $\oone$ is an alternative symbol for $-1$.  Thus $-\oone = 1$.
	For instance, $C_2 = [-1,1]\times [-1,1]\times \oone \times[-1,1]$,
	and $-C_2= [-1,1]\times [-1,1]\times 1 \times[-1,1]$,
	However, $\whSO$ is viewed as $\cup_{i=0}^3 C_i$ because
	of the equivalence of points $\bfp,\bfq\in \partial[-1,1]^4$
	where $\bfp\equiv \bfq$ iff $\bfp=-\bfq$. Therefore,
	$C_i\equiv -C_i$.

	Suppose $B\ib C_i$ and $B'\ib C_j$ are boxes in $\whSO$.
	We want to define the relation
			$$B\dder{\pm e_k} B'.$$
		This relation is not defined if $k=i$.  Otherwise:
	\benum[({Case} 1)]
	\item $i=j$:
		Then $B= \Proj_i(B)\otimes_i \oone$
		and $B'=\Proj_i(B')\otimes_i \oone$.
		We define $B\dder{\pm e_k} B'$ if and only 
				if ``$\Proj_i(B)\dder{\pm e_k}\Proj_i(B')$.''
		The precise definition requires us to shift the index $k$
		after projection in case $k>i$: let
				$$k'= \clauses{k &\rmif\ k<i,\\
							 k-1 & \rmif\ k>i.}$$	
		So, $B\dder{\pm e_k} B'$ if and only if
				$\Proj_i(B)\dder{\pm e_{k'}}\Proj_i(B')$.
	\item $i\ne j$:  
		Then $B= \Proj_{i,j}(B)\otimes_{i,j} (\oone\times I_j)$
		and $B'=\Proj_{i,j}(B')\otimes_{i,j} (I'_i\times \oone)$.
		We say the relation 
				$B\dder{\pm e_k} B'$
		is undefined if $k\neq j$.  Otherwise, we have 2 possibilities:\\
			(i) $B\dder{-e_j} B'$ if $I_j=[a_j,b_j]=[\oone,b_j]$
				and 
				$\Proj_{i,j}(B)\dder{-e_j}\Proj_{i,j}(B')$.\\
			(ii) $B\dder{+e_j} B'$ if $I_j=[a_j,b_j]=[a_j,1]$
				and
				$\Proj_{i,j}(B)\dder{+e_j}  -(\Proj_{i,j}(B'))$.
	\eenum

\ssect{Maintaining Principal Neighbor Pointers}
	Recall that in Section~4 we introduced the principal neighbor pointers
	for boxes in $\RR^3$ and in $\whSO$.
	If $B\in \intbox\RR^3$, it has $6$ principal neighbor
	pointers as in \cite{aronov+3}.
	But if $B\ib C_i\ib \whSO$, then it has 8 principal neighbor
	pointers, denoted $B.d$ where $d\in \set{\pm e_0\dd\pm e_3}$.
	If $B=C_i$, all 8 pointers are non-null; otherwise
	two of them are null, namely $B.e_i= B.(-e_i) =\Null$.
	
	We assume the \dt{T/R Splitting scheme} in which we
	split a box $B=B^t\times B^r$ by either
	splitting $B^t$ or splitting $B^r$. 
	We need to update the principal neighbor pointers
	after such a split.  Since the split of $B^t$ is standard,
	we focus on $B^r$.    Initially, $B^r=\whSO$ and
	all its pointers are null.  After the first split,
	we have four boxes, $C_0\dd C_3$.   Their
	pointers are initialized as follows:
		$C_i.(\pm e_j) =C_j$ for $j \ne i$, and $C_i.(\pm e_i)=\Null$.
	When $B\ib C_i$ is split,
	each of its eight children $B_i$ ($i=1\dd 8$) has
	its principal neighbor pointers set up as follows:
	Two of them are null, inherited from its parent.
	Three of them point to siblings (as in the standard octree split).
	Three of them point to non-siblings as follows: if
	$B_i.d$ is pointing to a non-sibling, then 
		$B_i.d \ass B.d$ in case $B.d$ is a leaf.
	Otherwise, $B_i.d$ will point to the child $B'$
	of $B.d$ such that $B_i\dder{d} B'$.

	Now that we have set up the principal neighbor pointers
	of the children of $B$, we need to update the principal
	neighbor pointers
	of the other boxes: any box $B'$ that used to point to $B$
	may need to be updated to point to a child of $B$
	(call this the ``reverse pointer update'').
	First, we have to show how to 
	get to such $B'$s.  Suppose $B.d= B'$ and $depth(B')<depth(B)$
	then there is nothing to do in this direction.
	Otherwise, if $depth(B')=depth(B)$ and $B'$ has children,
	we must go to all descendants of $B'$ that pointed to $B$,
	and redirect their $-d$ pointers to point to the appropriate
	child of $B$.  Note that there are always 6  non-null pointers,
	which can cross from $C_i$ to some other $C_j$ $(j\ne i)$.
	In contrast, for the translational boxes,
	there are boundary boxes that can have fewer than 6 non-null
	pointers.

\ignore{
	When we split a box $B$ into 8 child boxes, each child box $c(B)$ of
	$B$ has 3 sibling neighbors. We can easily set up the pointers to
	such siblings by the information of the {\em child indicator} of $B$,
	which uses 3 digits, each indicating the direction $+e_j$ or $-e_j$
	for $j = 0\dd 3$ where the digit for $\pm e_i$ is empty for the
	octree representing $C_i$. Excluding the 3 directions that are
	adjacent to its siblings, the remaining 3 directions of $c(B)$
	correspond to 3 2-dimensional faces that are common as those of $B$.
	We call such 3 directions
        {\bf boundary directions}.  Let $d$ be a boundary direction,
	which is normal to the corresponding 2-dimensional
	face and points outward. We initially let $c(B)$ inherit the
	$d$-neighbor pointer from its parent $B$, which points to the principal
	$d$-neighbor of $B$, denoted by $B'$, for all such $d$'s.  If $B'$ is
	a leaf then we are done with $c(B)$. Otherwise, $B'$ must be of the
	{\em same} size as $B$, therefore the children of $B'$ are of the same
	size as $c(B)$, and we need to update the $d$-neighbor pointer of
	$c(B)$ so that it points to some child $c(B')$ of $B'$ (rather than to
	$B'$). The child indicator in $B$ for $c(B)$ and the child indicator
	in $B'$ for $c(B')$ are the same except for the digit of the $\pm d$
	directions: $c(B)$ is in the $d$ direction and $c(B')$ is in the $-d$
	direction.
	
	For all neighbors of $B$ in the $d$ direction that are at
	the level of $c(B')$ and deeper (i.e., with box size smaller or equal
	to that of $c(B')$), we need to update their $(-d)$-neighbor pointers
	from pointing to $B$ to pointing to one of the 4 $d$-children of
	$B$. To this end, for each of such 4 $d$-children of $B$, say $c(B)$,
	we first use the $d$-pointer of $c(B)$ to find its principal
	$d$-neighbor $c(B')$ and put $c(B')$ to a queue $Q$. Then in each
	iteration, we retrieve one box $\bar{B}$ from $Q$, update the
	$(-d)$-neighbor pointer of $\bar{B}$ to point to $c(B)$, and put the 4
	$(-d)$-children of $\bar{B}$ to $Q$. This process is repeated until $Q$
	is empty. Note that this process also provides an approach to access
	all the $d$-neighbor leaves of a (new) leaf box $c(B)$.
	Moreover, suppose $c(B)$ is on the subdivision boundary of
        $C_i$. Recall that at the top level the $\pm e_j$-pointers
        in $C_i$ point to $C_j$ for $j \neq i$, and that $c(B)$
        initially inherits the boundary-direction pointers from its
        parent $B$ (and hence ultimately from $C_i$). Therefore a
        pointer of $c(B)$ in the boundary direction leads to its
        principal neighbor in another octree representing $C_j$ for
        some $j \neq i$, using our same construction and accessing
        methods as described above.

	T/R Split:

    Now we discuss our overall data structure for the subdivision of
	$\wh{SE}(3) = \RR^3 \times \wh{SO}(3)$.
	Initially, we have one leaf box $B_0 = B^t_0 \times B^r_0$,
        where its translational box $B^t_0 \ib \RR^3$ is a box in
        $\RR^3$ and its rotational box $B^r_0$ is $B^r_0 = \wh{SO}(3)
        = \cup_{i=0}^3 C_i$, consisting of 4 boxes $C_i$ as
        discussed previously. Therefore at the top level, we have a
        box $B^t_0$ in the translational space and 4 boxes $C_i$ in
        the rotational space.  We then perform repeated subdivisions
        on leaves to create our subdivision data structure. When we
        split a leaf box $B = B^t \times B^r$, we either split $B^t$
        in the translations space (called {\bf T-split}), or split
        $B^r$ in the rotational space (called {\bf R-split}).  A
        T-split on $B$ splits $B^t$ into 8 child boxes, where $B^r$
        stays the same.  Before performing any R-splits, our boxes all
        have the full rotational range and we do not need to store the
        boxes of $C_i$ explicitly.  An R-split on $B$ keeps $B^t$
        the same; an initial R-split subdivides $\wh{SO}(3)$ into 4
        boxes $C_i$ for $i = 0,\cdots, 3$.  After that, an R-split
        on $B$ splits $B^r$ (a box in the octree of $C_i$ for some
        $i$) into 8 child boxes, using the scheme discussed
        previously.
	Typically we first perform a sequence of T-splits, then we interleave
	T-splits and R-splits.
}

	We remark that Nowakiewicz \cite{nowakiewicz:mst:10} also
	discusses subdividing the translational and rotational boxes
	separately, and cubic model. However,
        %
        the method does not classify boxes, and does
		not compute or use the adjacency information of boxes.
\ignore{
        the
	method eventually takes sample configurations (at the corners or
	centers) in subdivision boxes and is actually a sampling-based method.
	It does not classify the boxes and does not compute or use the
	adjacency information of the boxes.
        }

\ignore{
	When we split a box $B$ into 8 child boxes, each child box $c(B)$ of
	$B$ has 3 sibling neighbors. We can easily set up the pointers to
	such siblings by the information of the {\em child indicator} of $B$,
	which uses 3 digits, each indicating the direction $+e_j$ or $-e_j$
	for $j = 0, \cdots, 3$ where the digit for $\pm e_i$ is empty for the
	octree representing $C_i$. Excluding the 3 directions that are
	adjacent to its siblings, the remaining 3 directions of $c(B)$
	correspond to 3 2-dimensional faces that are common as those of $B$.
	We call such 3 directions
        {\bf boundary directions}.  Let $d$ be a boundary direction,
	which is normal to the corresponding 2-dimensional
	face and points outward. We initially let $c(B)$ inherit the
	$d$-neighbor pointer from its parent $B$, which points to the principal
	$d$-neighbor of $B$, denoted by $B'$, for all such $d$'s.  If $B'$ is
	a leaf then we are done with $c(B)$. Otherwise, $B'$ must be of the
	{\em same} size as $B$, therefore the children of $B'$ are of the same
	size as $c(B)$, and we need to update the $d$-neighbor pointer of
	$c(B)$ so that it points to some child $c(B')$ of $B'$ (rather than to
	$B'$). The child indicator in $B$ for $c(B)$ and the child indicator
	in $B'$ for $c(B')$ are the same except for the digit of the $\pm d$
	directions: $c(B)$ is in the $d$ direction and $c(B')$ is in the $-d$
	direction.
	
	For all neighbors of $B$ in the $d$ direction that are at
	the level of $c(B')$ and deeper (i.e., with box size smaller or equal
	to that of $c(B')$), we need to update their $(-d)$-neighbor pointers
	from pointing to $B$ to pointing to one of the 4 $d$-children of
	$B$. To this end, for each of such 4 $d$-children of $B$, say $c(B)$,
	we first use the $d$-pointer of $c(B)$ to find its principal
	$d$-neighbor $c(B')$ and put $c(B')$ to a queue $Q$. Then in each
	iteration, we retrieve one box $\bar{B}$ from $Q$, update the
	$(-d)$-neighbor pointer of $\bar{B}$ to point to $c(B)$, and put the 4
	$(-d)$-children of $\bar{B}$ to $Q$. This process is repeated until $Q$
	is empty. Note that this process also provides an approach to access
	all the $d$-neighbor leaves of a (new) leaf box $c(B)$.
	Moreover, suppose $c(B)$ is on the subdivision boundary of
        $C_i$. Recall that at the top level the $\pm e_j$-pointers
        in $C_i$ point to $C_j$ for $j \neq i$, and that $c(B)$
        initially inherits the boundary-direction pointers from its
        parent $B$ (and hence ultimately from $C_i$). Therefore a
        pointer of $c(B)$ in the boundary direction leads to its
        principal neighbor in another octree representing $C_j$ for
        some $j \neq i$, using our same construction and accessing
        methods as described above.
}%

\ignore{ 
		Let us introduce a new symbol, $\oone$ to represent $-1$.
		Let $\Sigma\as \set{1,\oone}$ and the elements in $\Sigma$
		are called \dt{bits}.  Moreover $1$ is the \dt{positive} bit,
		and $\oone$ the \dt{negative} bit.  Let $\set{1,\oone}^*$
		denote the set of strings over $\set{1,\oone}$.
		The \dt{empty string} is denoted $\ess$.
	
		We can associate a string $s\in\Sigma^*$ with a 
		a subinterval $I(s)\ib [-1,1]$ as follows:
		for any interval $[a,b]$, let $Left([a,b])\as [a, (a+b)/2]$
		and $Right([a,b])\as [(a+b)/2, b]$.  Then
			$$I(s)\as \clauses{ [-1,1] & \rmif\ s=\ess,\\
							Left(I(t)) & \rmif\ s=t\oone,\\
							Right(I(t)) & \rmif\ s=t 1.}$$
		E.g., $I(\oone)=[-1,0]$, $I(\oone 1)=[-\half,0]$.
		$I(\oone 1\oone)=[-\half,-\quarter]$.
		Let $1^\omega\as 111\cdots$ and $\oone^\omega\as \oone\oone\cdots$
		infinite strings of $1$'s and $\oone$'s (respectively).
		Then $I(1^\omega)=[1,1]=1$ and $I(\oone^\omega)=[-1,-1]=-1$.
		Let
		$\Sigma^\omega\as \Sigma^* \cup \set{1^\omega, \oone^\omega}$.
		The \dt{length} of $s\in \Sigma^\omega$ is denoted $|s|$.
		Moreover $|s|=\infty$ iff $s=1^\omega$ or $s=\oone^\omega$;
		otherwise, $|s|$ is a non-negative integer defined in the usual way.
	
		If $b\in\Sigma$
		is a bit, let $b'\in\Sigma$ be uniquely defined by
		the inequality $b'\neq b$.  In other words,
		$1'=\oone$, and $\oone'=1$.  We may call $(\cdot)'$ the
		\dt{flip} operation.
		We classify non-empty strings into one
		of three types: \dt{positive}, \dt{negative}
		and \dt{mixed}:  
		positive strings has only positive bits;
		and similarly for negative string;
		otherwise the string is mixed.
	
		We introduce three operations on strings over $\Sigma$.  Let
			$$\beta=b_1b_2\cdots b_k\in\Sigma^*$$
		be a string of length $k\ge 0$ where each $b_i\in\Sigma$.
		The three operations are $\beta\mapsto \beta'$ (prime),
		$\beta\mapsto \ol{\beta}$ (bar)
		and $\beta\mapsto \beta^-$ (adj).
		To illustrate these operations, let
		$\beta_1=1\oone \oone 111$
		$\beta_2=1\oone \oone 111\oone\oone$.
			\benum[(1)]
		\item (prime) $\beta'$ is defined to be $b_1\cdots b_{k-1} b_k'$
				(i.e., we flip the last bit $b_k$, leaving
				others unchanged).
				Thus $\beta_1'=1\oone \oone 11\oone$
				and $\beta_2'=1\oone \oone 111\oone 1$.
				If $k=0$, then $\beta=\ess$ and we define $\ess'$ to be $\ess$.
			\item (bar) $\ol{\beta}\as b_1'\cdots b_k'$ (i.e., we 
				flip each bit in $\beta$.
				Thus $\ol{\beta_1}= \oone 11 \oone \oone\oone$
				and $\ol{\beta_2}= \oone 11 \oone \oone\oone 11$.
				Again, we define $\ol{\ess}=\ess$.
			\item (adj) To define $\beta^-$, we define
				the \dt{suffix} $\suff(\beta)$ to be
				the shortest suffix that is mixed.  Note that if $\beta$
				is not mixed, then we define $\suff(\beta)=\beta$.
				The \dt{prefix} $\pre(\beta)$ is defined by this equation:
						$$\beta=\pre(\beta)\suff(\beta).$$
				For instance $\pre(\beta)=\ess$ iff $\beta$ is not mixed.
				Finally, we define
					$$\beta^- \as \pre(\beta)\ol{\suff(\beta)}.$$
				I.e., we replace suffix of $\beta$ by its bar.
				
				For instance, $\suff(\beta_1)=\oone 111$
				and $\suff(\beta_2)=1\oone\oone$.
				It follows that $\pre(\beta_1)=1\oone$
				and $\pre(\beta_2)=1\oone \oone 11$.
				Therefore
					$$\beta_1^- = 1\oone 1\oone\oone\oone, \quad
						\beta_2^-=1\oone \oone 11 \oone 1 1.$$
			\eenum
		
	
			We are now ready for the calculus of cubes.
			
			Given a cube $C_i$, we may subdivide it into subcubes
			by repeated splitting into 8 children.  
			These subcubes obtained by repeated subdivision will be called
			\dt{boxes}.  At level $k$,
			there are $8^k$ boxes which can be denoted using
			this notation: 
					$$C_i(\beta_1,\beta_2,\beta_3,\beta_4)$$
			where $\beta_j=\clauses{\oone^\omega & \rmif\ j=i,\\
									      s	\in\Sigma^k & \rmif\ j\neq i}$.
			For example, the box
			$$B_1\as
				C_2(1,\oone^\omega, \oone, 1)=[0,1]\times (-1) \times[-1,0]\times[0,1]$$
			is a child of $C_2=[-1,1]\times (-1) \times [-1,1]\times [-1,1]$
			at level 1.  The box
			$$B_2\as C_2(1\oone,\oone^\omega, \oone\oone, 11)=
					[0,\half]\times (-1) \times[-1,-\half]\times[\half,1]$$
			is a child of $B_1$ at level 2.
	
			Let $B$ be a box.
			Another box $B'$ is a \dt{neighbor} of $B$ if $B\cap B'$ is
			a 2-dimensional square.  We are interested in neighbors
			of $B$ that belong to the same level (and thus are congruent
			to $B$).  Each box $B$ has exactly 6 congruent neighbors:
			we can classify a congruent neighbor $B'$ into two types:
			\\(i) $B'$ is a \dt{sibling} if $B'$ and $B$ have the same parent.
			\\(ii) $B'$ is a \dt{cousin} if it is not a sibling.
	
			\blem
			Consider the box
					$$B=C_i(\beta_1,\beta_2,\beta_3,\beta_4).$$
			Then $B$ has three
			siblings, and they are contained in this list:
				\beqarrays
					C_i(\beta_1',\beta_2,\beta_3,\beta_4),\\
					C_i(\beta_1,\beta_2',\beta_3,\beta_4),\\
					C_i(\beta_1,\beta_2,\beta_3',\beta_4),\\
					C_i(\beta_1,\beta_2,\beta_3,\beta_4').
				\eeqarrays
			The three cousins of $B$ can be identified as follows:
			For simplicity, first assume that each non-infinite
			$\beta_i$ is mixed.  Then the three cousins of $B$ are
			contained in this list:
				\beqarrays
					C_i(\beta_1^-,\beta_2,\beta_3,\beta_4),\\
					C_i(\beta_1,\beta_2^-,\beta_3,\beta_4),\\
					C_i(\beta_1,\beta_2,\beta_3^-,\beta_4),\\
					C_i(\beta_1,\beta_2,\beta_3,\beta_4^-).
				\eeqarrays
			\elem
} 
\sect{Proof of Fundamental Theorem}\label{app:pf}


\ssect{Two Lemmas on Separation}
	
	Separation does not satisfy the triangular inequality because
	$\Sep(A,C)\le \Sep(A,B)+\Sep(B,C)$
	may be false.  
	We have a kind of ``triangular inequality'' if one of the summands
	is replaced by the Hausdorff distance:
	
	\blemT{Pseudo Triangle Inequality}{pseudoTriIneq}
		Let $(X,d)$ be a metric space. Then for any $A,B,C\subseteq X$,
				\[\Sep(A,C)\leq d_H(A,B)+\Sep(B,C).\]
	\elemT
	\bpf
		Let $a\in A, b\in B, c\in C$:
		\beqarrys
		\Sep(A,C) &=& \inf_{a,c} d_X(a,c)\\
				&\le& \inf_{b} \big( \inf_{a,c} d_X(a,b)+d_X(b,c)\big) 
						& \text{(triangular inequality)}\\
				&\le& \inf_{b} \big( \Sep(A,b) + \Sep(b,C)\big) \\
				&\le& \Sep(A,b^*) + \Sep(b^*,C)
						& \text{(choose $b^*$ so that
								$\Sep(b^*,C)=\Sep(B,C)$}\\
				&=& \Sep(A,b^*) + \Sep(B,C)\\
				&\le& d_H(A,B) + \Sep(B,C)
		\eeqarrys
	\epf
	
		In particular, for $\gamma,\gamma'\in X$, the pseudo triangular
		inequality implies
				\beql{clearance}
					|\Cl(\gamma)-\Cl(\gamma')|\le d_H(\gamma,\gamma').
					\eeql
	
	We have another lemma on separation that relates to the Minkowski-sum.

	\blem[Separation of Minkowski-Sum]
	Suppose that closed sets $A,B,C\subseteq\RR^m$. If $(A\oplus B)\cap
	C=\emptyset$, then for any $a\in A$, $\Sep(a\oplus B,C)\geq
	d_{\RR^m}(a,\partial A)$.
	\elem
	
	\begin{center}
	\includegraphics[width=8cm]{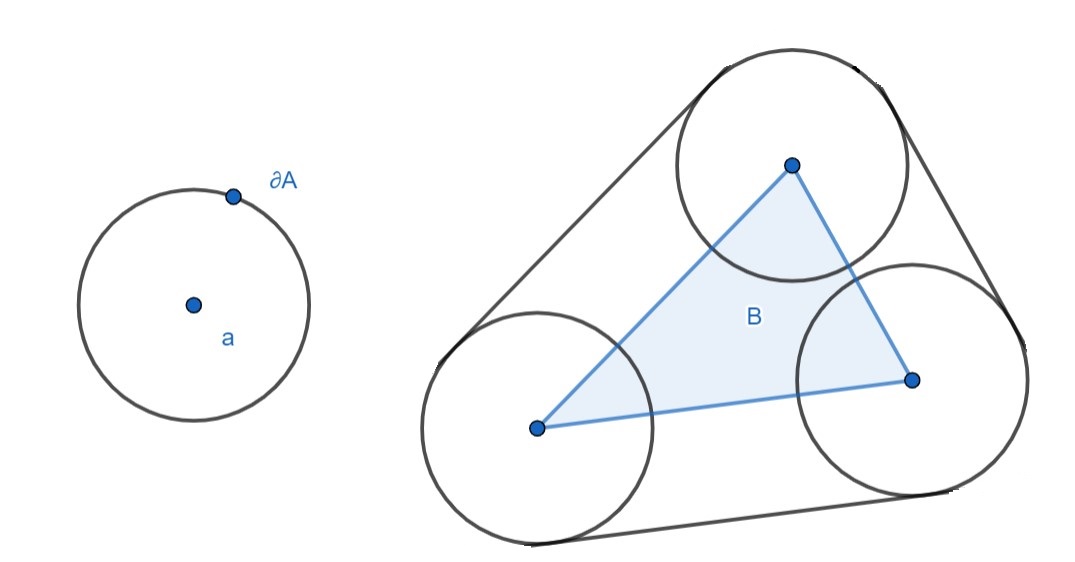}
	\end{center}
	
	\bpf
	We denote the complement of a set $S$ in $\RR^m$ by $\complement S$. Then since $(A\oplus B)\cap C=\emptyset$, $C\subseteq\complement(A\oplus B)$. Since $a\oplus B\subseteq A\oplus B$, we have
	\begin{align*}
	\Sep(a\oplus B,C) & \geq\Sep(a\oplus B,\complement(A\oplus B))\\
	& =\Sep(a\oplus B,\partial\complement(A\oplus B))\\
	& =\Sep(a\oplus B,\partial(A\oplus B)).
	\end{align*}
	Then since
	\begin{align*}
	\Sep(a\oplus B,\partial(A\oplus B)) & =\inf_{(a'+b')\in\partial(A\oplus B),b\in B}d_{\RR^m}(a+b,a'+b')\\
	& \geq\inf_{a'\in\partial A,b\in B,b'\in\partial B}d_{\RR^m}(a+b,a'+b')\\
	& \geq\inf_{a'\in\partial A,b\in B}d_{\RR^m}(a+b,a'+b)\\
	& =\inf_{a'\in\partial A}d_{\RR^m}(a,a')\\
	& =d_{\RR^m}(a,\partial A),
	\end{align*}
	we have
	\[\Sep(a\oplus B,C)\geq d_{\RR^m}(a,\partial A).\]
	\epf

\ssect{Proof of SSS framework}

	A subdivision is called a \dt{uniform $\veps$-subdivision}, if the
	widths of all the boxes in the subdivision are the same and it is
	less than $\veps$.

	\bleml{cover}
	Assume the Axioms \Azero, \Aone, \Atwo\ and \Athree.
	If there exists a path $\pi$ in $W$ with
	clearance $K\veps$ where $K=L_0C_0D_0\sigma$, then the SSS framework can
	find a path $\olP$.
	\eleml
	
	\bpf
	Suppose there exists a path $\pi$ with clearance
			$K\veps=L_0C_0D_0\sigma\veps$.
	We obtain a contradiction
	by assuming that the SSS framework could not find a path.
	In this case, the SSS framework will subdivide $\intbox
	W$ into minimum sizes. Hence, in this subdivision, all $\mixed$ tiles
	have width less than $\varepsilon$. Without loss of generality,
	let us assume the subdivision is a uniform $\veps$-subdivision.
	We consider all tiles in this subdivision that intersect with
	$\pi$. The union of these tiles forms a set
	$\textrm{Cover}(\pi)$. It is easy to see that this is a
	path-connected and compact set. Hence, by Poincar\'{e}'s duality,
	there is a sequence $P$ of adjacent boxes
	in the $\textrm{Cover}(\pi)$.  We
	prove that these boxes are $\free$, i.e., $P$ is a channel.
	Each tile $B\in P$ is $\free$ under the soft
	predicate, that is, $\forall q\in B$, $\Cl(q)>0$.
	
	For each tile $B\in P$, we consider the tile $\sigma B$.
	Since there is $t\in[0,1]$ such that $\pi(t)\in B\subseteq\sigma B$,
	for each $q\in\sigma B$, by axioms $\Atwo$ and $\Athree$, we have
	\begin{align*}
		\|\Cl(q)-\Cl(\pi(t))\| & \leq d_H(\mu(q),\mu(\pi(t)))
				& \indent & \textrm{(Pseudo Triangle Inequality)} \\
		& <L_0 d_X(\mu(q),\mu(\pi(t))) & \indent & \Atwo\\
		& <L_0C_0d_W(q,\pi(t)). & \indent & \Athree
	\end{align*}
	
	By axiom $\Aone$ and since the diameter of tile $\sigma B$ is no
	greater than $l(\sigma B)$, we have
	\begin{align*}
		d_W(q,\pi(t)) & \leq l(\sigma B) & \indent & \textrm{(definition)}\\
		& \leq D_0 w(\sigma B) & \indent & \Aone\\
		& <D_0\sigma\varepsilon. & \indent & \textrm{(width is
			}\varepsilon\textrm{)}
	\end{align*}
	
	Hence $\|\Cl(q)-\Cl(\pi(t))\|<L_0C_0D_0\sigma\varepsilon$ and then
	\[\Cl(q)\geq \Cl(\pi(t))-L_0C_0D_0\sigma\varepsilon>0.\]
	This gives $\forall q\in \sigma B$, $\Cl(q)>0$. So tile $B\in P$ is
	$\free$ under the soft predicate.
	\epf
	
	\Remark\  It can be seen from the proof that the coefficient
	$L_0C_0D_0\sigma$ is not related to the number of children each tile
	is split into. So the $D_0$ in this term can be taken as the
	aspect ratio of the tiles.
	
	\bleml{clearance}
		Assume the Axioms \Azero\  and \Afour. 
		If the SSS framework returns a path $\olP$,
		then there exists a path $\olP'$
		of essential clearance $\half\veps$.
	\eleml
	\bpf
	If $\olP$ is returned, then it implies that there is
	a channel $P'$ in the uniform $\veps$-subdivision.
	Let $\olP'$ be the concatenation of three polygonal paths,
			$$\olP'= (S_\alpha; \olP''; S_\beta)$$
	where $\olP''$ connects the midpoints of
	the adjacent boxes of $P'$, and $S_\alpha$ is the segment
	connecting $\alpha$ to the 
	midpoint of the initial box, and $S_\beta$ is the segment 
	connecting the midpoint of the last box to $\beta$.
	For the essential clearance of the path $\olP'$, we only need
	to ensure that the clearance of $\olP''$ is at least $\half\veps$.
	
	Let $B_1=B_1^t\times B_1^r$
	and $B_2=B_2^t\times B_2^r$ be the two adjacent tiles,
	their centers are $m_1=m_1^t\times m_1^r$ and $m_2=m_2^t\times m_2^r$
	respectively. The line segment connecting $m_1$ and $m_2$ is
	$\ol{m_1m_2}$.  Since $\dim(B_1\cap B_2)=k-1$, we have either
	$B_1^t=B_2^t$ or $B_1^r=B_2^r$. We discuss these two cases
	respectively.
	
	When $B_1^t=B_2^t=B^t$, and $m_1^t=m_2^t=m^t$, for any
	$q\in\ol{m_1m_2}$, we have $q=m^t\times b^r$. Since, $B_1$ and $B_2$
	are $\free$,
	\begin{align*}
	 B_1\textrm{ and }B_2\textrm{ are } \free
		& \Rightarrow Fp(B_1)\cap\Omega=\emptyset
			\textrm{ and }
		Fp(B_2)\cap\Omega=\emptyset
				\indent \text{(by \Azero)}\\
	 & \Rightarrow \left(Fp(B_1)\cup Fp(B_2)\right)
			\cap\Omega=\emptyset\\
	 & \Rightarrow \left(\left(B^t\oplus Fp(B_1^r)\right)
		\cup
		\left(B^t\oplus Fp(B_2^r)\right)\right)
		\cap
		\Omega=\emptyset \indent \text{(by \Afour)}\\
	 & \Rightarrow\left(B^t\oplus\left(Fp(B_1^r)
			\cup Fp(B_2^r)\right)\right)\cap\Omega=\emptyset
	\end{align*}
	\begin{center}
	\includegraphics[width=4cm]{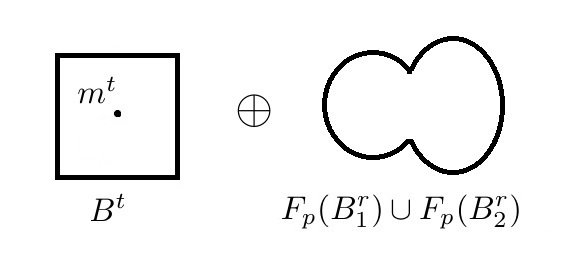}
	\end{center}
	Then since $m^t\in B^t$,
		$Fp(\ol{m_1^rm_2^r})\subseteq\left(Fp(B_1^r)\cup
			Fp(B_2^r)\right)$,
		and so by separation of Minkowski-sum,
	\begin{align*}
		\Cl(\ol{m_1m_2}) & =\Sep(m^t\oplus Fp(\ol{m_1^rm_2^r}),\Omega)\\
			& \geq\Sep(m^t\oplus\left(Fp(B_1^r)
				\cup Fp(B_2^r)\right),\Omega)\\
			& \geq d_{\RR^m}(m^t,\partial B^t)\\
			& \geq\frac{1}{2}\varepsilon.
	\end{align*}
	
	When $B_1^r=B_2^r=B^r$, and $m_1^r=m_2^r=m^r$, for any
	$q\in\ol{m_1m_2}$, we have $q=q^t\times m^r$. Since, $B_1$ and $B_2$
	are $\free$,
	\begin{align*}
	 B_1\textrm{ and }B_2\textrm{ are }\free & \Rightarrow
		Fp(B_1)\cap\Omega=\emptyset\textrm{ and
		}Fp(B_2)\cap\Omega=\emptyset \indent \Azero\\
	 & \Rightarrow \left(Fp(B_1)\cup Fp(B_2)\right)\cap\Omega=\emptyset\\
	 & \Rightarrow \left(\left(B_1^t\oplus
		Fp(B^r)\right)\cup\left(B_2^t\oplus
		Fp(B^r)\right)\right)\cap\Omega=\emptyset \indent \Afour\\
	 & \Rightarrow \left(\left(B_1^t\cup B_2^t\right)\oplus
		Fp(B^r)\right)\cap\Omega=\emptyset.
	\end{align*}
	\begin{center}
	\includegraphics[width=4cm]{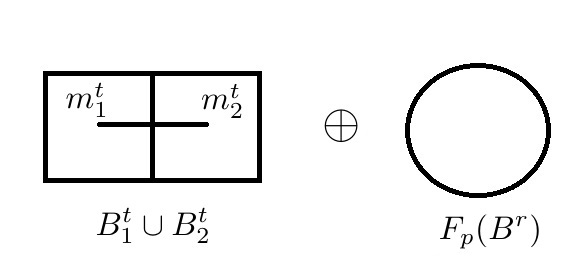}
	\end{center}
	Then since for any $q\in\ol{m_1m_2}$, we have $q\in B_1^t\cup B_2^t$,
	and so by Lemma on Separation of Minkowski-sum,
	\begin{align*}
		\Cl(\ol{m_1m_2}) & =\inf_{q\in\ol{m_1m_2}}\Sep(q^t\oplus
			Fp(B^r),\Omega)\\
		& \geq\inf_{q\in\ol{m_1m_2}}d_{\RR^m}(q^t,\partial(B_1^t\cup B_2^t))\\
		& =\Sep(\ol{m_1^tm_2^t},\partial(B_1^t\cup B_2^t))\\
		& \geq\frac{1}{2}\varepsilon.
	\end{align*}
	Hence, for any consecutive centers $m_1$ and $m_2$, the clearance of
	direct connection is always no less than $\half\veps$.
	Therefore, the path $\olP''$ has clearance $\half\veps$.
	\epf
	
	REMARK: our construction of $\olP'$ uses a rectilinear
	path $\olP''$, while our SSS algorithm uses a non-rectilinear
	path.  This non-rectilinear path only has a clearance of
	$\tfrac{1}{2\sqrt{n}}\veps$.  Note that Axiom \Afour\
	is extremely strong and ensures a clearance independent
	of constants such as $L_0, D_0,\sigma$.

	The definition of resolution exactness has two requirements,
	(Path) and (NoPath).  
	\refLem{cover} satisfies the requirement of (Path), and
	\refLem{clearance} satisfies the requirement of (NoPath).
	This implies the resolution-exactness of our SSS planner:
	
	\bthm
	Assuming the Axioms \Azero, \Aone, \Atwo, \Athree\  and \Afour,
	the SSS algorithm is resolution exact with
	constant $K=\max\{L_0C_0D_0\sigma,2\}$.
	\ethm

\ssect{Analysis of the SSS Algorithm for the Delta robot}
	We now analyze the basic properties
	of our SSS path planner for the Delta robot.

	\bthm
		The Lipschitz constant in Axiom \Atwo\ is
			$L_0=1$.
	\ethm
	\bpf
	Given any two configurations, we could compare their Hausdorff
	distance by moving the relative center of a triangle to another
	without changing its direction in $r\in SO(3)$, i.e. for each point
	$T\in\triangle \pA \pO \pB$, let $t=|T \pO|$, and since our
	$\lambda=1=\sup t$, we have the inequalities:
		\begin{align*}
		d_H((\bfx,r),(\bfx',r'))
		& \leq d_H((\bfx,r),(\bfx,r'))+d_H((\bfx,r'),(\bfx',r'))\\
		& <t\theta+d_{\RR^3}(\bfx,\bfx')\\
		& =t d_{SO(3)}(r,r')+d_{\RR^3}(\bfx,\bfx')\\
		& \leq \lambda d_{SO(3)}(r,r')+d_{\RR^3}(\bfx,\bfx')\\
		& =d_X((\bfx,r),(\bfx',r')).
		\end{align*}
	Hence, the minimum Lipschitz constant $L_0\leq1$. In fact, 
	we could take $L_0=1$, since we have
	taken the physical length of the edge $\pA \pO$ as $1$ unit in $\RR^3$.
	\epf
	

	\bthm[Theorem 2 in main paper]
	The approximate footprint of the Delta robot
			is $\sigma$-effective with $\sigma=(2+\sqrt{3})<3.8$.
	\ethm
	
	\bpf
	Notice that the radius of $\wtFp_\pA(B^r)$, which is the approximate footprint 
	of $\pA$ for a rotation box is less than
	$\theta$ (the rotation angle), which is restricted by
	$\sqrt{4}w(B)/2=w(B)$ (box dimension is $4$, atlas constant is $2$),
	so $\wtFp(B^r)\subseteq Ball(0,w(B))\oplus Fp(B^r)$. Hence,
	we have
	\begin{align*}
	\wtFp(B^t\times B^r) & =Ball(B^t)\oplus \wtFp(B^r)\\
	& \subseteq Ball(m(B^t),\sqrt{3}w(B^t)/2)
		\oplus Ball(0,w(B))\oplus Fp(B^r)\\
	& =Ball(m(B^t),(2+\sqrt{3})w(B^t)/2)\oplus Fp(B^r)\\
	& \subseteq Fp((2+\sqrt{3})B^t)\oplus Fp(B^r)\\
	& \subseteq Fp((2+\sqrt{3})B).
	\end{align*}
	So this soft predicate is $(2+\sqrt{3})$-effective.
	\epf

	\bthm[Theorem 3 in main paper]
	Our SSS planner for the Delta Robot is
	resolution exact with constant
		$K=4\sqrt{6}+6\sqrt{2}< 18.3$.
	\ethmT
	\bpf
	The SSS framework for the Delta robot
	is resolution exact with constant $K=4\sqrt{6}+6\sqrt{2}$
	since $L_0=1, C_0=2, D_0=\sqrt{6}, \sigma=2+\sqrt{3}$.
	\epf

\end{document}